\documentclass[11pt]{article}

\usepackage[a4paper,margin=1in]{geometry}
\usepackage{times}

\usepackage{amsmath,amsfonts,bm}

\def\1{\bm{1}}

\DeclareMathAlphabet{\mathsfit}{\encodingdefault}{\sfdefault}{m}{sl}
\SetMathAlphabet{\mathsfit}{bold}{\encodingdefault}{\sfdefault}{bx}{n}

\def\gN{{\mathcal{N}}}

\newcommand{\pdata}{P_{\rm{data}}}

\newcommand{\R}{\mathbb{R}}

\DeclareMathOperator{\Var}{Var}
\DeclareMathOperator{\diag}{diag}

\DeclareMathOperator{\Tr}{Tr}
\newcommand{\bbE}{\mathbb{E}}

\usepackage{xcolor}
\usepackage[T1]{fontenc}
\usepackage[utf8]{inputenc}
\usepackage{lmodern}
\usepackage{amsmath,amssymb,mathtools,bm,empheq}
\usepackage{graphicx}
\usepackage{microtype}
\usepackage[hidelinks]{hyperref}
\usepackage{tcolorbox}
\usepackage{array}
\usepackage{tikz}
\usepackage{enumitem}
\usepackage{natbib}
\usetikzlibrary{positioning,arrows.meta,calc}
\usepackage{url}
\usepackage{subcaption}
\usepackage{wrapfig}

\newcommand{\spiralplot}[1]{%
    \includegraphics[width=0.17\textwidth]{figures/collected_thick_spiral_samples_new/#1}%
}
\newcommand{\spirallabel}[1]{%
    \raisebox{0.066\textwidth}[0pt][0pt]{\mbox{#1}}%
}

\usepackage{booktabs}
\usepackage{multirow}

\definecolor{keyeqblue}{RGB}{242,248,253}
\definecolor{keyeqframe}{RGB}{112,154,190}
\newcommand*\keyeqbox[1]{%
    \fcolorbox{keyeqframe}{keyeqblue}{\hspace{0.5em}#1\hspace{0.5em}}%
}
\definecolor{keyeqpurpleframe}{RGB}{128, 70, 160}
\definecolor{keyeqpurple}{RGB}{245, 235, 250}
\newcommand*\purpleeqbox[1]{%
    \fcolorbox{keyeqpurpleframe}{keyeqpurple}{\hspace{0.5em}#1\hspace{0.5em}}%
}

\newcommand{\maybeinclude}[2][]{%
\IfFileExists{#2}{\includegraphics[#1]{#2}}{%
\fbox{\parbox[c][4cm][c]{0.9\linewidth}{\centering Missing figure:\\ \texttt{\detokenize{#2}}}}}%
}

\usepackage[nameinlink]{cleveref}

\definecolor{tabblue}{HTML}{1f77b4}
\definecolor{taborange}{HTML}{ff7f0e}
\definecolor{tabgreen}{HTML}{2ca02c}
\definecolor{tabred}{HTML}{d62728}

\usepackage{amsthm}
\theoremstyle{definition} 
\newtheorem{theorem}{Theorem}
\newtheorem{remark}[theorem]{Remark} 
\Crefname{remark}{Remark}{Remarks}
\Crefname{equation}{Eq.}{Eqs.}
\Crefname{appendix}{App.}{App.}

\hypersetup{
	colorlinks=true,       
	linkcolor=blue,        
	citecolor=violet!80!black,     
	filecolor=magenta,    
	urlcolor=blue
}

\usepackage[normalem]{ulem}

\title{Beyond the Manifold Hypothesis: Hybrid Spectral Parameterizations for Flow \mbox{Matching}}

\author{
Ségolène Martin$^{1}$ \quad
Anne Gagneux$^{2}$ \quad
Quentin Bertrand$^{3}$ \\
Rémi Emonet$^{3,4}$ \quad
Mathurin Massias$^{1}$
\\[0.8em]
\begin{tabular}{c}
\small
$^{1}$ Inria, ENS de Lyon, CNRS, Université Claude Bernard Lyon 1, \\
\small
LIP UMR 5668, 69342 Lyon Cedex 07, France
\\[0.3em]
\small
$^{2}$ ENS de Lyon, CNRS, Université Claude Bernard Lyon 1, Inria, \\
\small
LIP UMR 5668, 69342 Lyon Cedex 07, France
\\[0.3em]
\small
$^{3}$ Université Jean Monnet Saint-Étienne, CNRS, Institut d'Optique Graduate School, \\
\small
Inria, Laboratoire Hubert Curien UMR 5516, F-42023 Saint-Étienne, France
\\[0.3em]
\small
$^{4}$ Institut Universitaire de France
\end{tabular}
}

\begin{document}
\maketitle

\begin{abstract}
Flow matching and diffusion can be trained to predict different quantities, most commonly the data $x_1$, the source noise $x_0$, or the velocity~$v$. Although theoretically equivalent, these can lead to substantially different performances.
We identify two main drivers for these differences: the source--data signal-to-noise ratio, and the information bottleneck induced by the neural architecture. We show that, beyond intrinsic data dimension, the factor affecting the optimal parametrization the most is a certain signal-to-noise ratio in each data covariance direction.
From this analysis, we introduce new \emph{spectral hybrid} parameterizations that adapt across time and data covariance directions; we show that these are optimal for Gaussian data.
We also show that architecture-induced compression changes which parameterization is easier to learn, with $v$-prediction being more sensitive to discarded directions than $x_1$-prediction.
Experiments across architectures and source scales show that our spectral parameterizations are robust across regimes, can substantially accelerate optimization, while incurring essentially no additional training cost compared with standard parameterizations.
\end{abstract}


\looseness=-1
The modern success of diffusion models \citep{sohlDickstein2015diffusion,Ho2020,Song2021} heavily relies on a number of specific design choices whose impact are often not yet fully understood.
Key design choices include specific
neural architectures, such as U-Nets \citep{ronneberger2015unet} or ViTs \citep{dosovitskiy2021vit,peebles2023dit},
loss weighting schemes \citep{karras2022elucidating},
optimizers \citep{kingma2014adam},
and the choice of network parameterization, \emph{i.e.}, which quantity the network is trained to predict.

Network parameterization has recently attracted renewed interest.
The seminal diffusion works \citep{song2019generative,Ho2020} originally parameterized the network to predict the noise added to the data.
With the rise of flow matching \citep{lipman2023flow,liu2023rectifiedflow,albergo2023stochasticinterpolant}, 
predicting the velocity emerged as a strong contender.
More recently, motivated by the \emph{manifold hypothesis} \citep[Chap. 1.2]{chapelle2006semi}, which posits that natural images exhibit a low-dimensional structure,
\citet{li2025back} or \citet{geng2026improvedmeanflows} have advocated for predicting the clean data.

Images exhibit structured 
regularities that have long been exploited through image-specific architectures \citep{LeCunBBH98,KrizhevskySH12}, as well as tailored training objectives.
Modern approaches for incorporating image-specific information into generative models rely on two-stage procedures \citep{van2017neural,esser2021taming,tian2024visual}, in which a latent representation is first learned using image-specific 
perceptual losses \citep{zhang2018unreasonable} that have been empirically shown to better preserve high frequencies.
Incorporating the image statistics directly into one-stage training procedures has also been actively explored, for instance through anisotropic noising schemes and frequency-dependent noise schedules \citep{rissanen2022generative,hoogeboom2022blurring,falck2025fourier}.


\paragraph{Contributions.}
In this work, we propose a principled approach to
network parameterization by elucidating how it connects to the spectral structure of the data. More precisely:

\begin{itemize}[leftmargin=*,topsep=-2pt]
    \item First, we challenge the common intuition that the \emph{manifold hypothesis} favors $x_1$-prediction:
    we show that  rescaling the source distribution, while leaving the data manifold unchanged, can reverse the performance ordering between $x_0$-, $v$-, and $x_1$-prediction (\Cref{fig:spiral_dimension_snr}).
    This highlights the source--data SNR as a key factor for parameterization performance.
    \item Second, based on these findings, we introduce two \emph{spectral hybrid
    parameterizations} that adapt between $x_0$ and $x_1$ across time and data covariance directions.
    In a Gaussian-to-Gaussian model, we show that they are optimal.
    \item 
    Third, we investigate the interactions between the parameterization and the architecture, highlighting the role of information compression.
    We show that velocity prediction is particularly sensitive to discarded information, while the spectral hybrids are very robust.
    Experiments on image datasets show that the proposed
    spectral hybrid parameterizations are competitive with the best fixed parameterization across source scales and architectures, with larger gains under strong architectural bottlenecks,
    and improve training convergence.
\end{itemize}

\section{Reminders}\label{sec:reminders}


Flow matching and diffusion transform a simple source distribution\footnote{In this paper, we adopt the flow matching convention: $x_0$ is the noise, $x_1$ is the clean data.} 
$P_0$ into a target data distribution $\pdata$, by mapping a sample $x_0 \sim P_0$ to the solution at time 1 of an Ordinary Differential Equation (ODE) governed by a velocity field $V$
\begin{equation}\label{eq:ode}
    \begin{cases}
        x(0) = x_0 \\
        \partial_t x(t) = V(x(t), t)   \quad \forall t \in [0, 1]
    \end{cases}
\end{equation}
In the most basic version of flow matching, for independent $X_0 \sim P_0$ and $X_1 \sim \pdata$, $V$ is learned as a neural network $N_\theta$ whose parameters $\theta$ solve
\begin{equation}
    \min_\theta \bbE_{\substack{X_0, X_1, t \sim \mathrm{unif}([0, 1])}} [\Vert N_\theta(X_t, t) - (X_1 - X_0) \Vert^2]   \quad \text{where } X_t := (1-t) X_0 + t X_1.
\end{equation}
If the neural network can approximate any measurable function and optimization is performed exactly, the optimal network is $\mathbb E[X_1 - X_0 \mid X_t= \cdot]$, for which the solutions of \eqref{eq:ode} at time 1 are guaranteed to be distributed according to $\pdata$ under mild regularity conditions.

Instead of learning the velocity ($v$-prediction), one can parametrize the network $N_\theta$ to predict, from $X_t$, either the data $X_1$  ($x_1$-prediction),
the source $X_0$ ($x_0$-prediction).
Still assuming infinite model capacity, the corresponding networks
recover the conditional means
$\mathbb E[X_1\mid X_t=x]$, $\mathbb E[X_0\mid X_t=x]$, and
$\mathbb E[X_1 - X_0 \mid X_t=x]$.
As detailed in \Cref{app:equivalence_param_loss}, these three quantities are algebraically related: knowledge of one implies knowledge of the other two.
In practice however, each parametrization can lead to very different performances.

In order to study their respective performances, we compare parameterizations through their induced estimator of $X_1$ given $x_t$,
which we refer to as the denoiser.
Let $N_\theta^p$ denote the raw network
output for a parametrization $p\in\{x_1,v,x_0\}$. 
The corresponding denoisers are
\begin{align}
    D_\theta^{x_1}(x,t)
    &=
    N_\theta^{x_1}(x,t),
    &
    D_\theta^{v}(x,t)
    &=
    x+(1-t)N_\theta^{v}(x,t),
    &
    D_\theta^{x_0}(x,t)
    &=
    \tfrac{x-(1-t)N_\theta^{x_0}(x,t)}{t}.
    \label{eq:param_classes}
\end{align}
In the infinite-capacity limit, all three recover the same optimal
denoiser
$D^\star(x,t):=\mathbb E[X_1\mid X_t=x]$. 
To compare them fairly, we train all models against the same target, $X_1$:
\begin{equation}
    \mathcal L^p (\theta)
    =
    \mathbb E_{t,X_0,X_1}
    \left[
        \tfrac{1}{(1-t)^2}
        \left\|
            D_\theta^p(X_t,t)-X_1
        \right\|^2
    \right],
    \qquad
    t\sim\mathcal U([0,1]).
    \label{eq:denoising_loss}
\end{equation}
\looseness=-1
We use the weighting $(1-t)^{-2}$ for all parameterizations in \eqref{eq:denoising_loss}, as it was found to achieve
the best performances across parameterizations
\citep{li2025back,gagneux2026training}\footnote{In the vocabulary of \citet{li2025back}, this means we always use $v$-loss; see~\Cref{app:equivalence_param_loss}.}. This also ensures that our comparisons isolate
parameterization rather than time weighting.

\begin{remark}\label{rk:velocity_error}
The parameterizations also differ in how prediction errors are propagated to the velocity used for sampling \eqref{eq:ode}.
Notably, under $x_0$-prediction, the corresponding velocity $V_\theta^{x_0} = (X_t-N_\theta^{x_0}(X_t,t)) / {t}$ becomes singular when $t \to 0$\footnote{The amplification of prediction errors also depends on the
interpolation schedule. Standard diffusion schedules behave differently from the path considered here; see
\Cref{app:diffusion_parameterizations}.}:
sampling with this parameterization 
yields poor generative performances \citep{gagneux2026training}.
This motivates the following evaluation protocol. We compare the
parameterizations both through their denoising MSE as a function of $t$ and
through their generative performance (FID).
Since sampling with $x_0$-pred is unstable, we only include $x_1$ and $v$-prediction for FID comparisons.
\end{remark}

\section{Beyond the manifold hypothesis: role of source--data SNR}
\label{sec:snr}

\begin{figure*}[t]
    \centering

    \renewcommand{\spiralplot}[1]{%
        \includegraphics[width=0.36\linewidth]{figures/collected_thick_spiral_samples_new/#1}%
    }
    \renewcommand{\spirallabel}[1]{%
        \raisebox{0.14\linewidth}[0pt][0pt]{\mbox{#1}}%
    }
    \begin{minipage}[t]{0.325\textwidth}
        \centering
        \footnotesize
        \textbf{Fixed source scale}
        \\[1mm] 
        $\sigma_0^2 = 1$
        \tcbox[
            colback=gray!8,
            colframe=gray!60,
            boxrule=0.8pt,
            arc=3mm,
            boxsep=0pt,
            left=0.5mm,
            right=0mm,
            top=0.5mm,
            bottom=0mm
        ]{%
        \begin{tabular}{@{}c@{\hspace{-1mm}}c@{}c@{}}
            & $x_1$-pred & $v$-pred \\[-1mm]
            \spirallabel{$D=16$} &
            \spiralplot{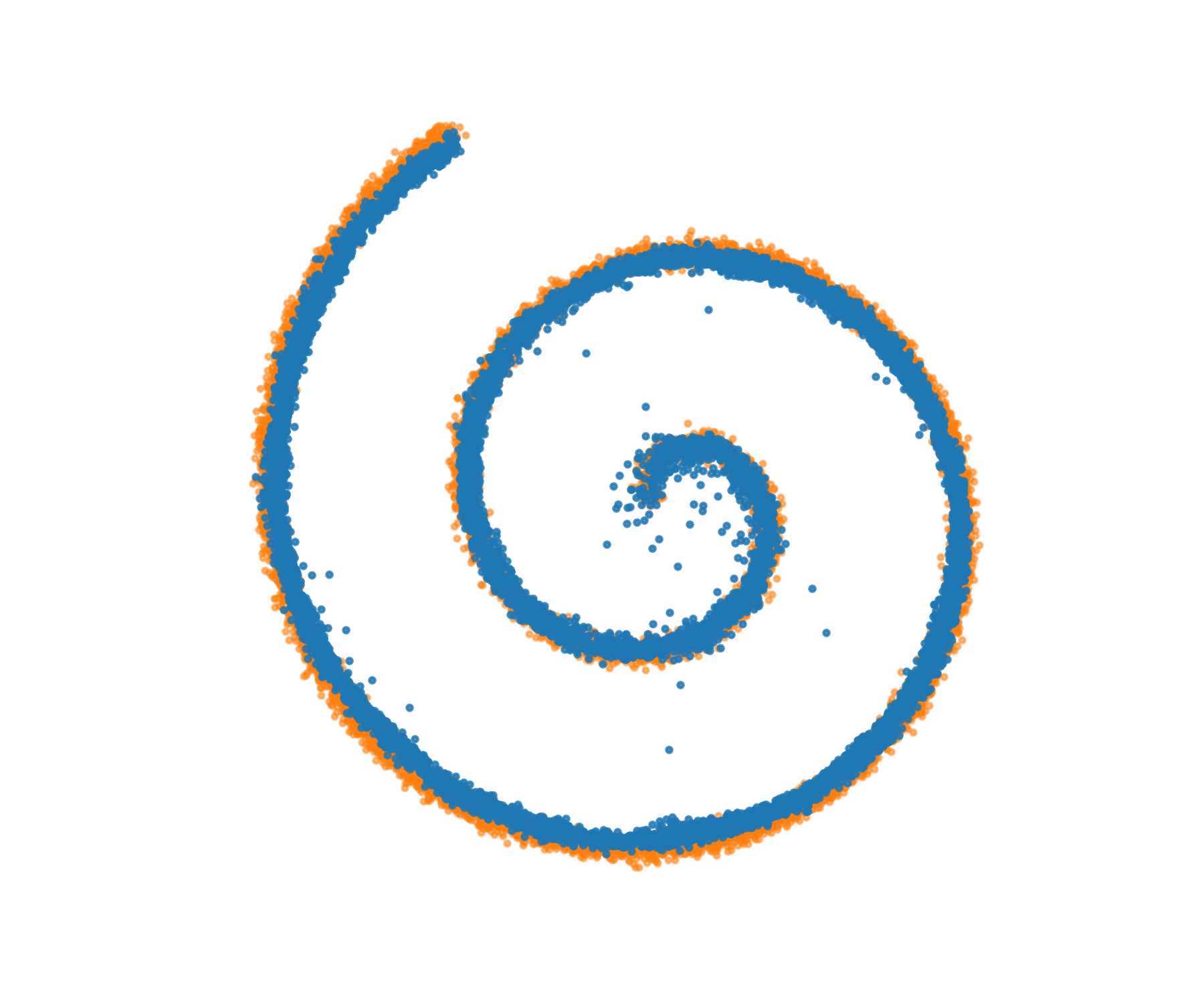} &
            \spiralplot{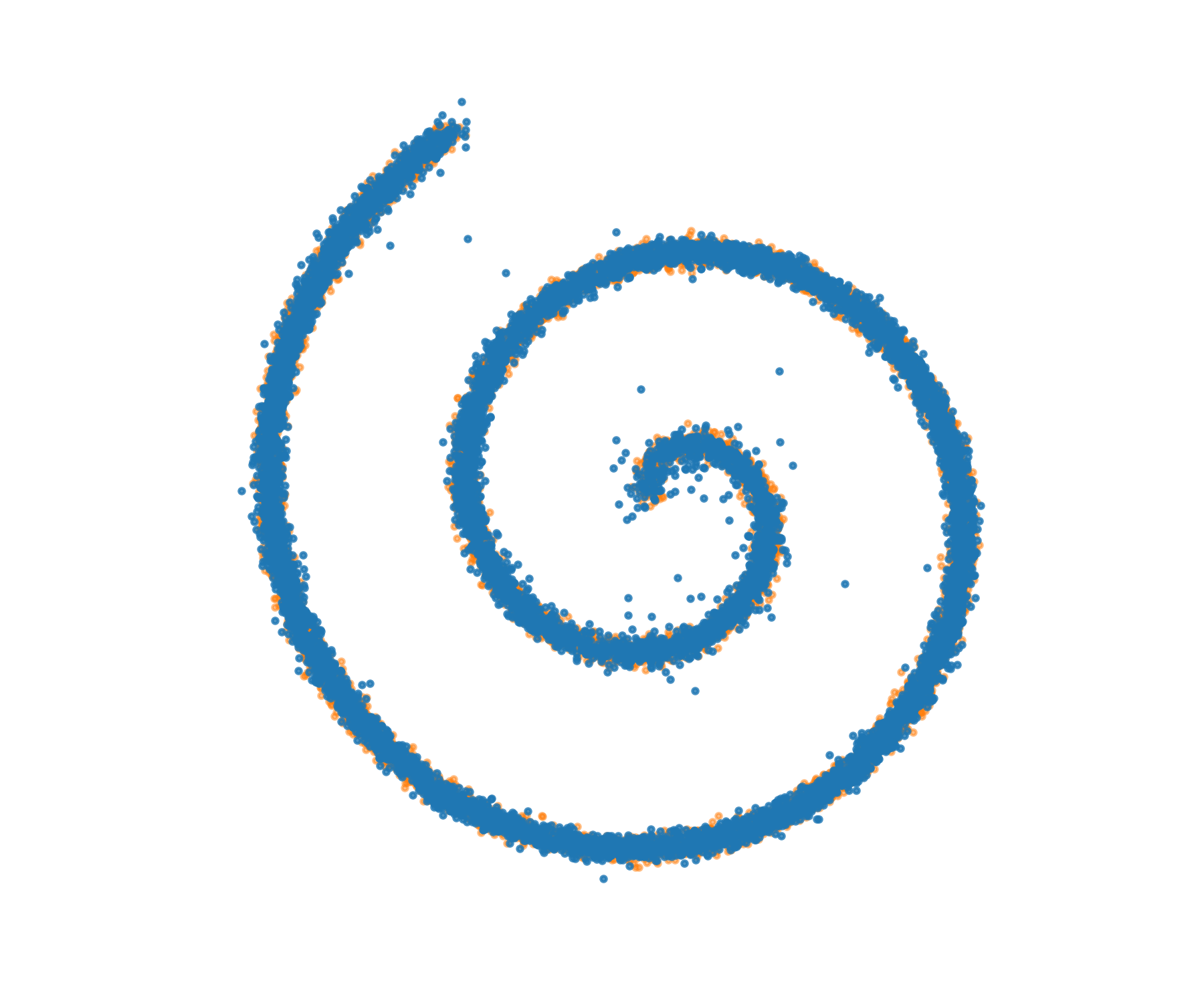} \\[-2mm]
            \spirallabel{$D=128$} &
            \spiralplot{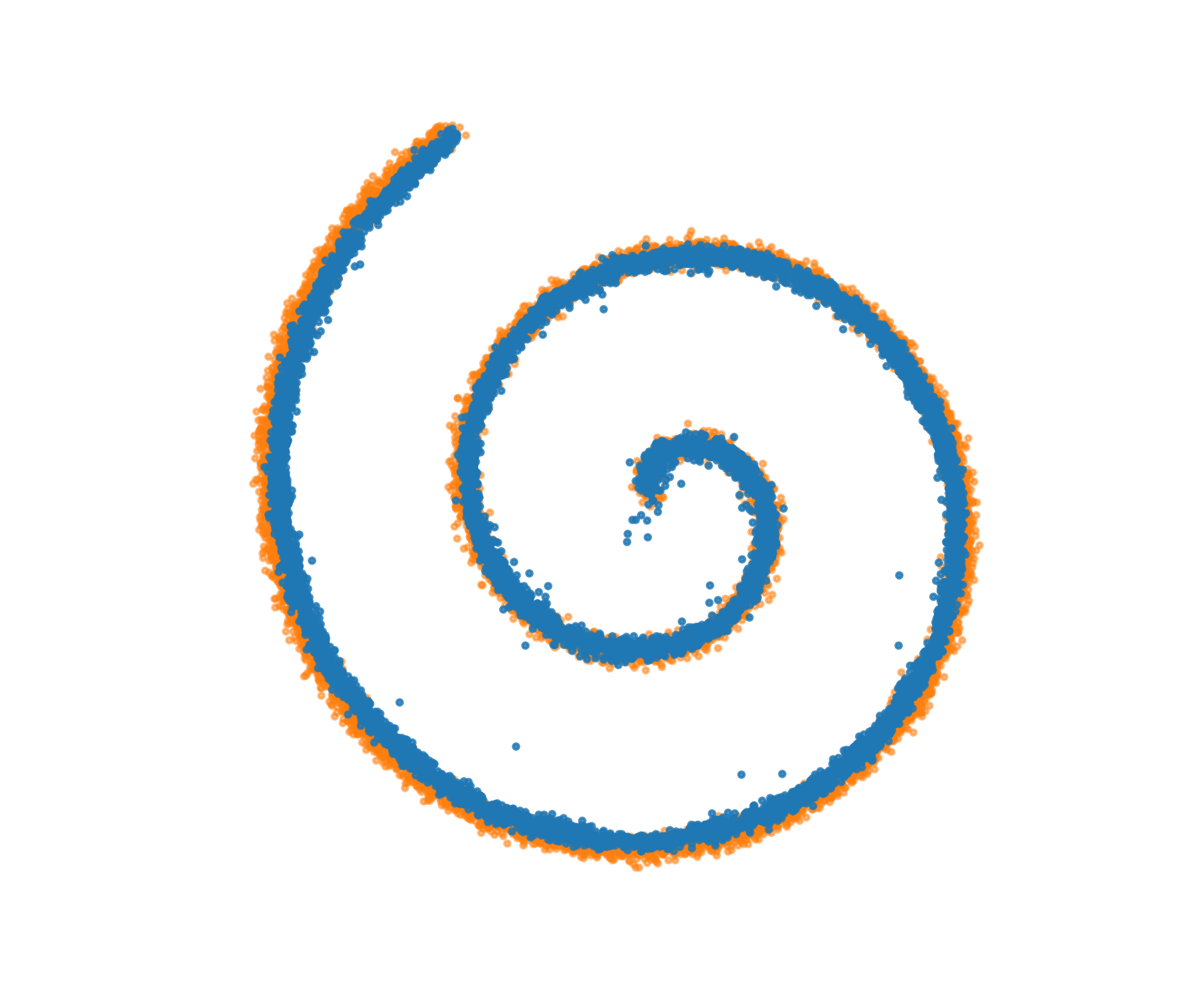} &
            \spiralplot{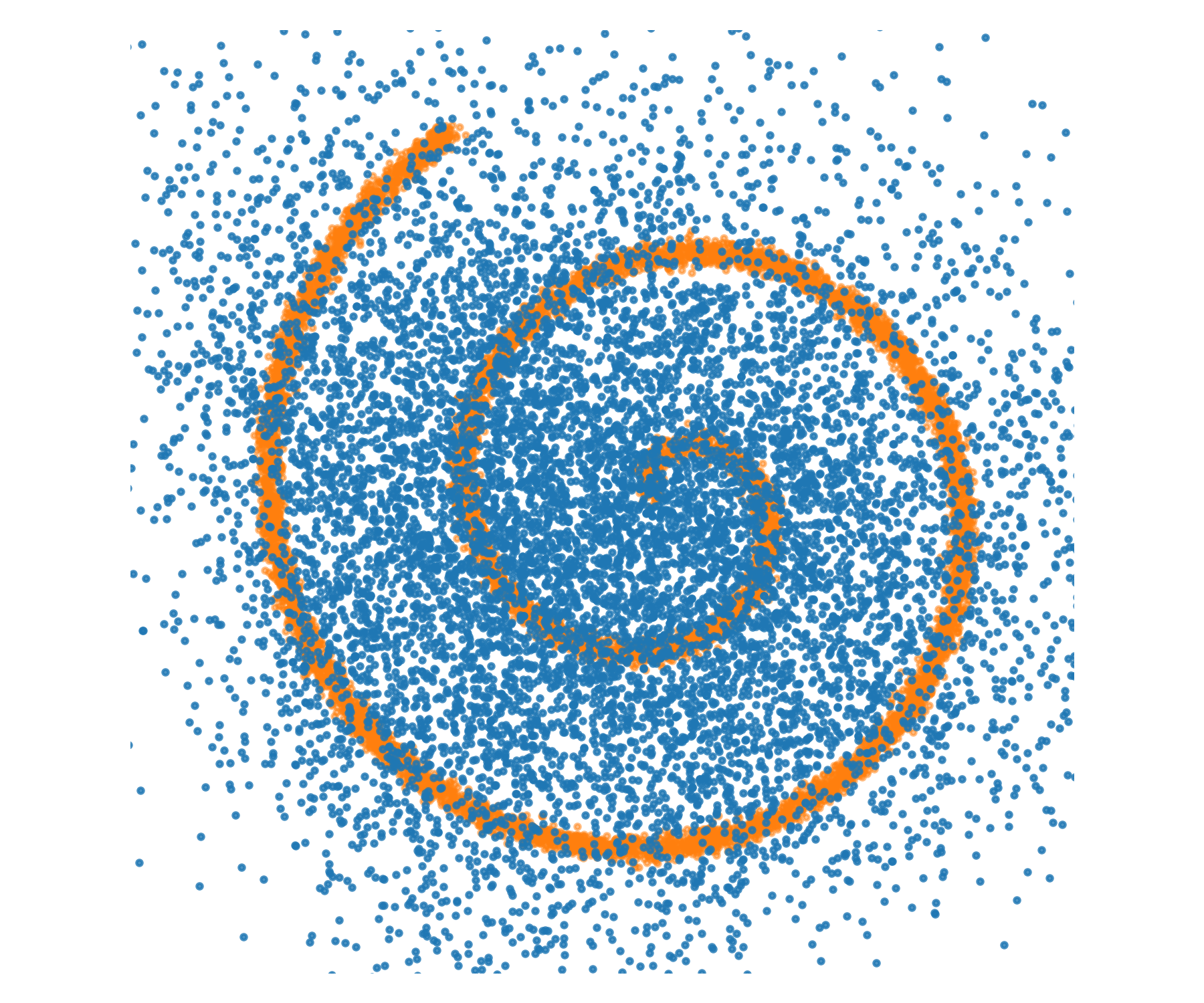} \\[-2mm]
            \spirallabel{$D=256$} &
            \spiralplot{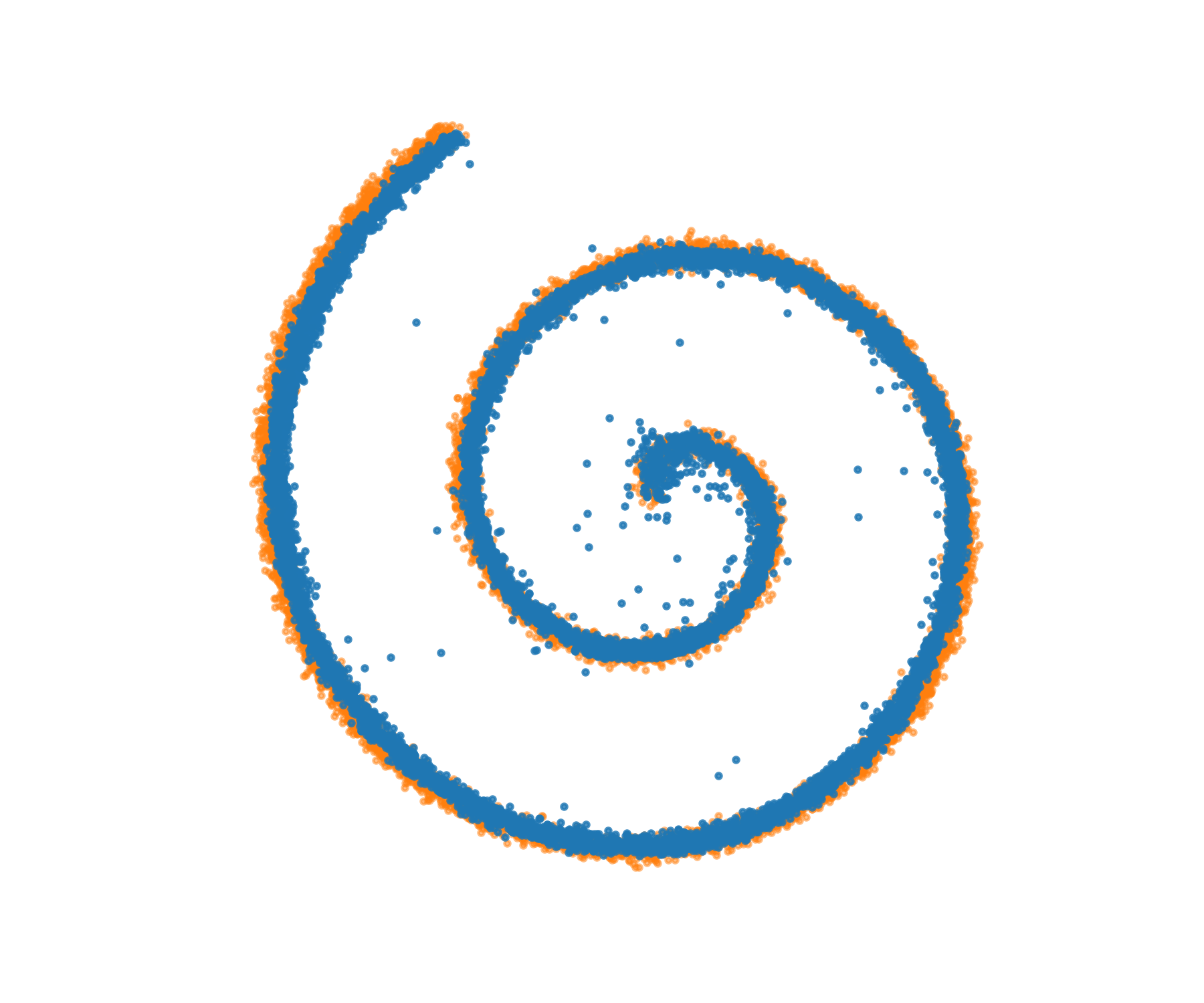} &
            \spiralplot{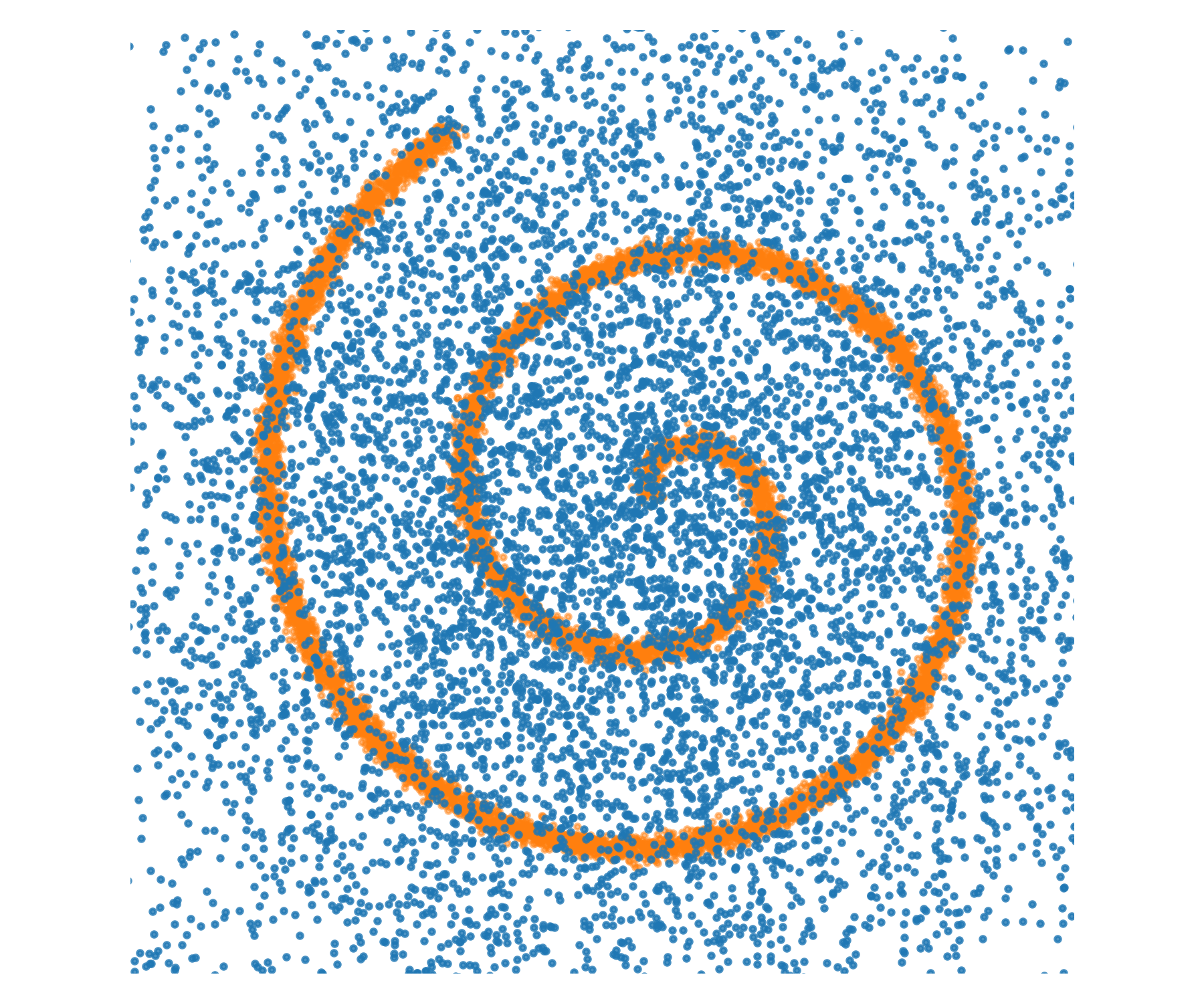}
        \end{tabular}
        }
    \end{minipage}\hfill
    \begin{minipage}[t]{0.325\textwidth}
        \centering
        \footnotesize
        \textbf{Fixed SNR} $\rho = 1$\\[1mm] $\sigma_0^2 = {\Tr\Sigma_1}/{D}$
        \tcbox[
            colback=gray!8,
            colframe=gray!60,
            boxrule=0.8pt,
            arc=3mm,
            boxsep=0pt,
            left=0.5mm,
            right=0mm,
            top=0.5mm,
            bottom=0mm
        ]{%
        \begin{tabular}{@{}c@{\hspace{-1mm}}c@{}c@{}}
            & $x_1$-pred & $v$-pred \\[-1mm]
            \spirallabel{$D=16$} &
            \spiralplot{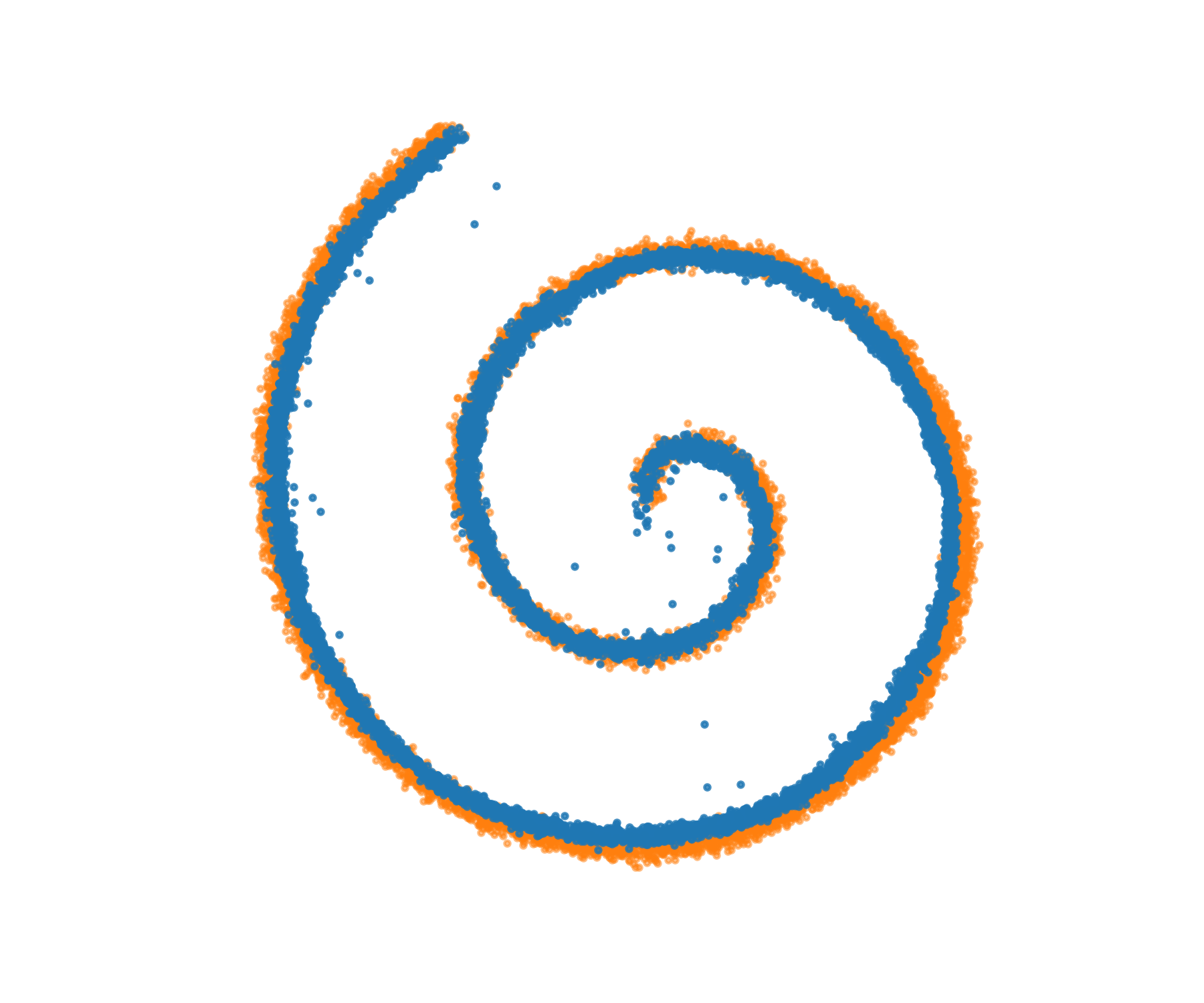} &
            \spiralplot{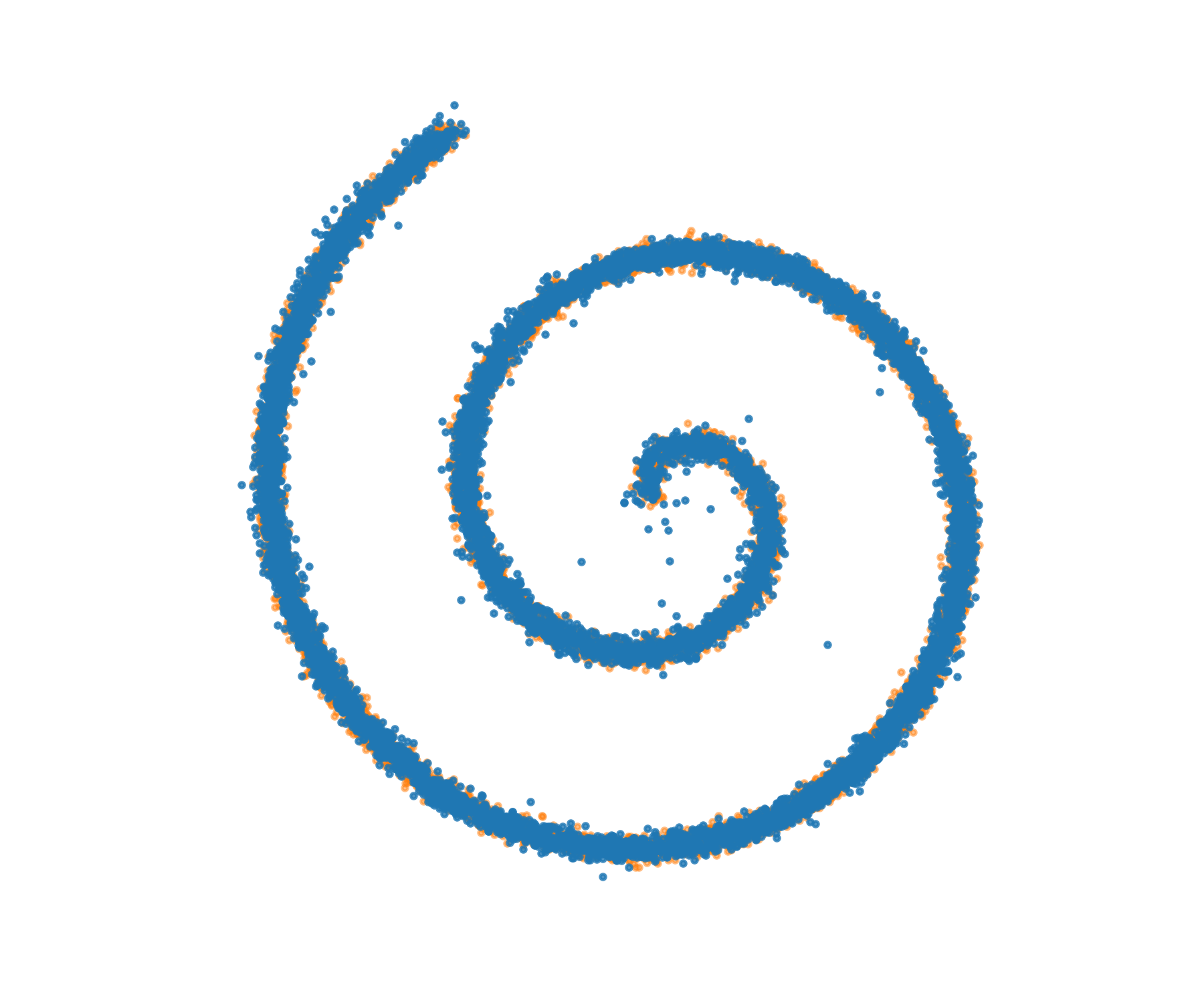} \\[-2mm]
            \spirallabel{$D=128$} &
            \spiralplot{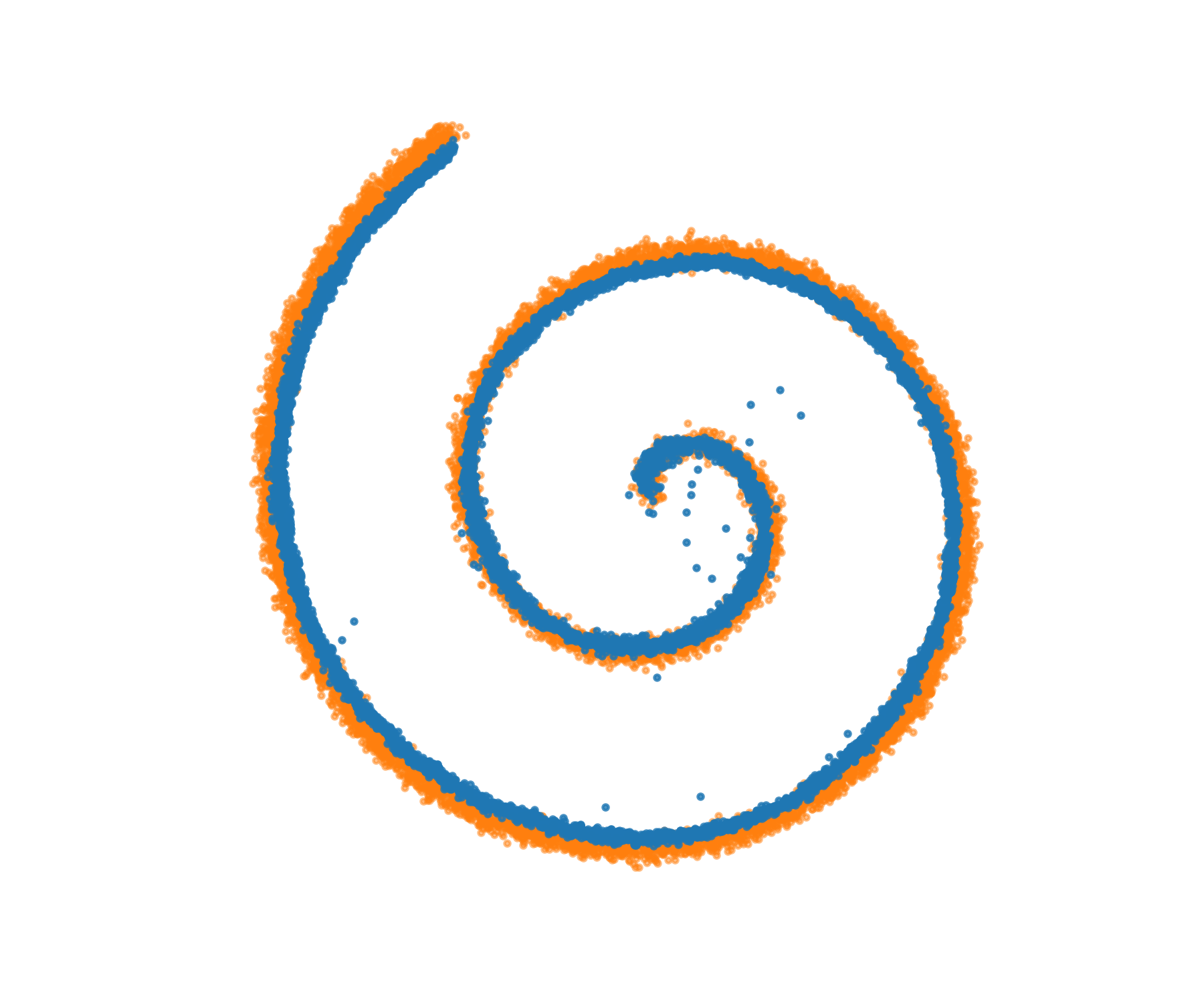} &
            \spiralplot{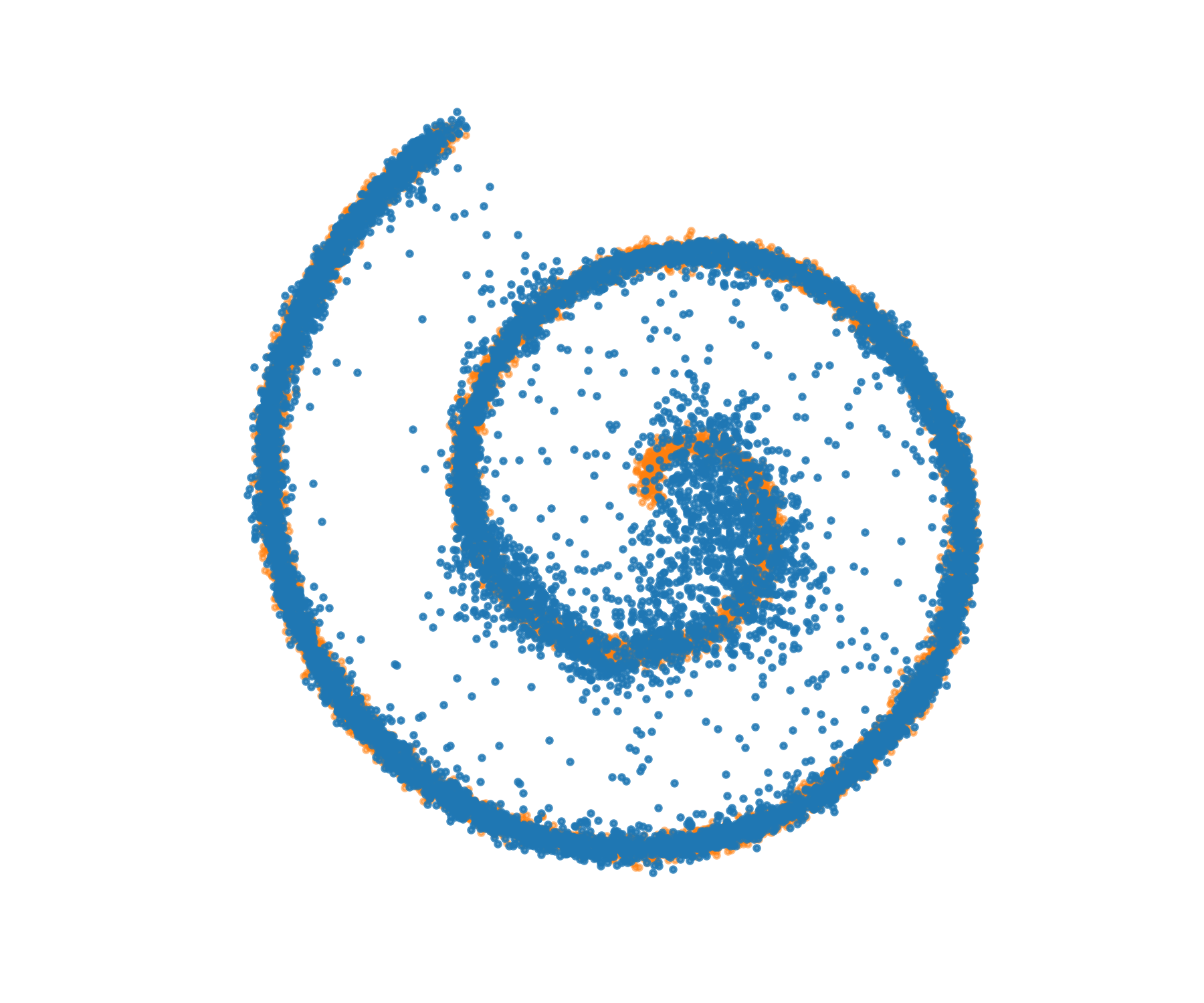} \\[-2mm]
            \spirallabel{$D=256$} &
            \spiralplot{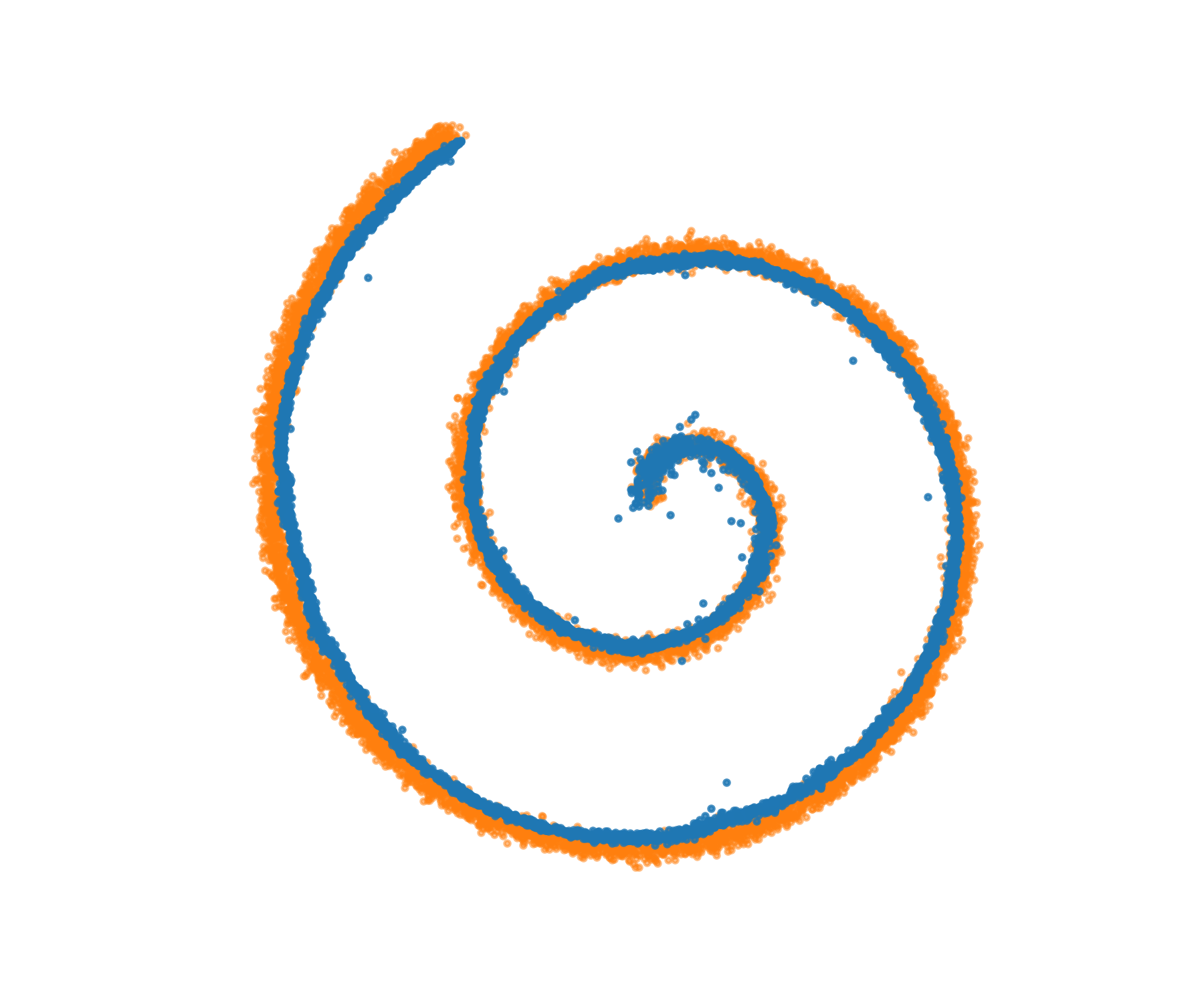} &
            \spiralplot{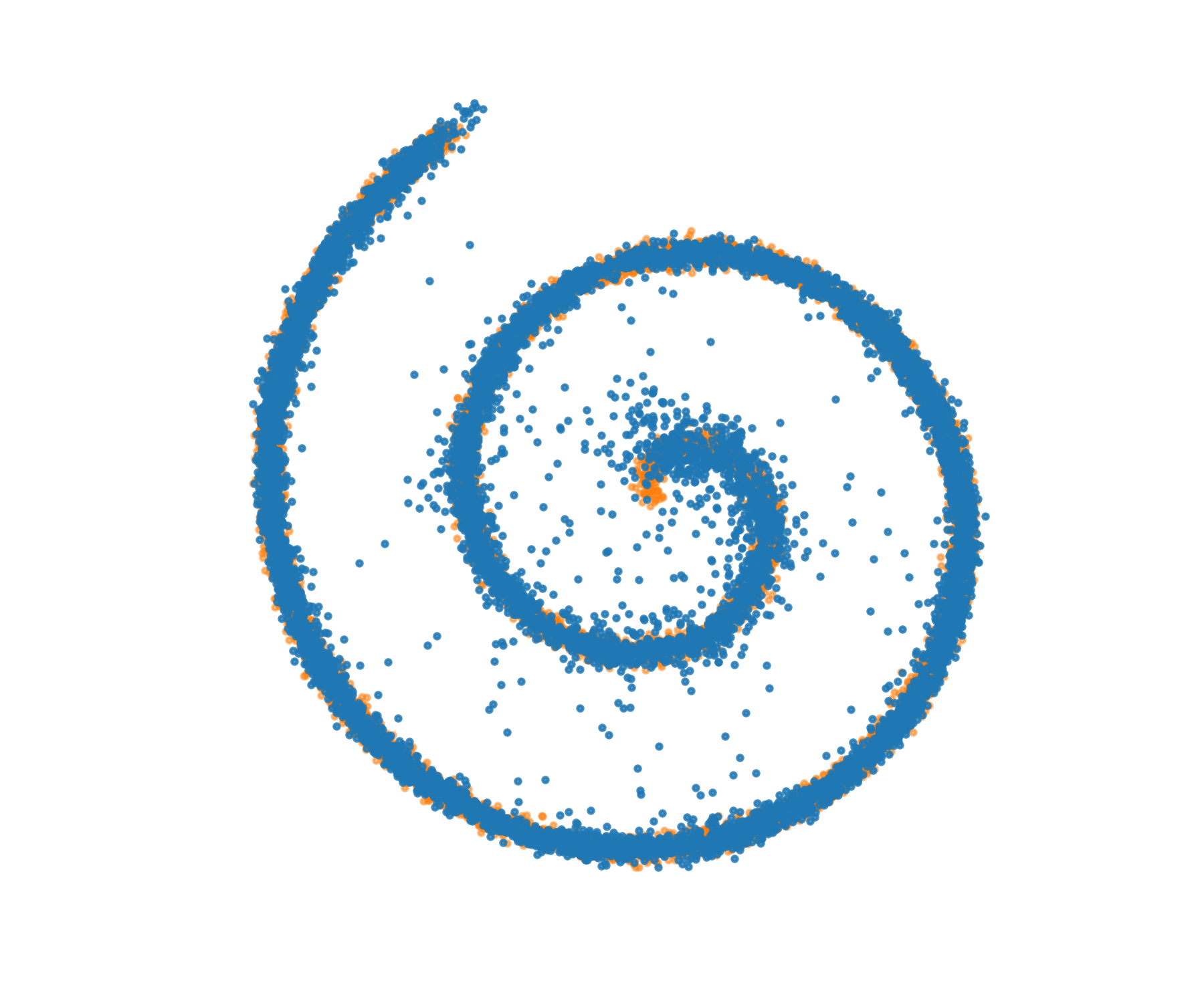}
        \end{tabular}
        }
    \end{minipage}\hfill
    \begin{minipage}[t]{0.325\textwidth}
        \centering
        \footnotesize
        \textbf{Fixed ambient dim $D=256$}\\[1mm]  $\sigma_0^2 = {\Tr\Sigma_1}/{(D \rho)}$
        \tcbox[
            colback=gray!8,
            colframe=gray!60,
            boxrule=0.8pt,
            arc=3mm,
            boxsep=0pt,
            left=0.5mm,
            right=0mm,
            top=0.5mm,
            bottom=0mm
        ]{%
        \begin{tabular}{@{}c@{\hspace{-1mm}}c@{}c@{}}
            & $x_1$-pred & $v$-pred \\[-1mm]
            \spirallabel{$\rho=0.01$} &
            \spiralplot{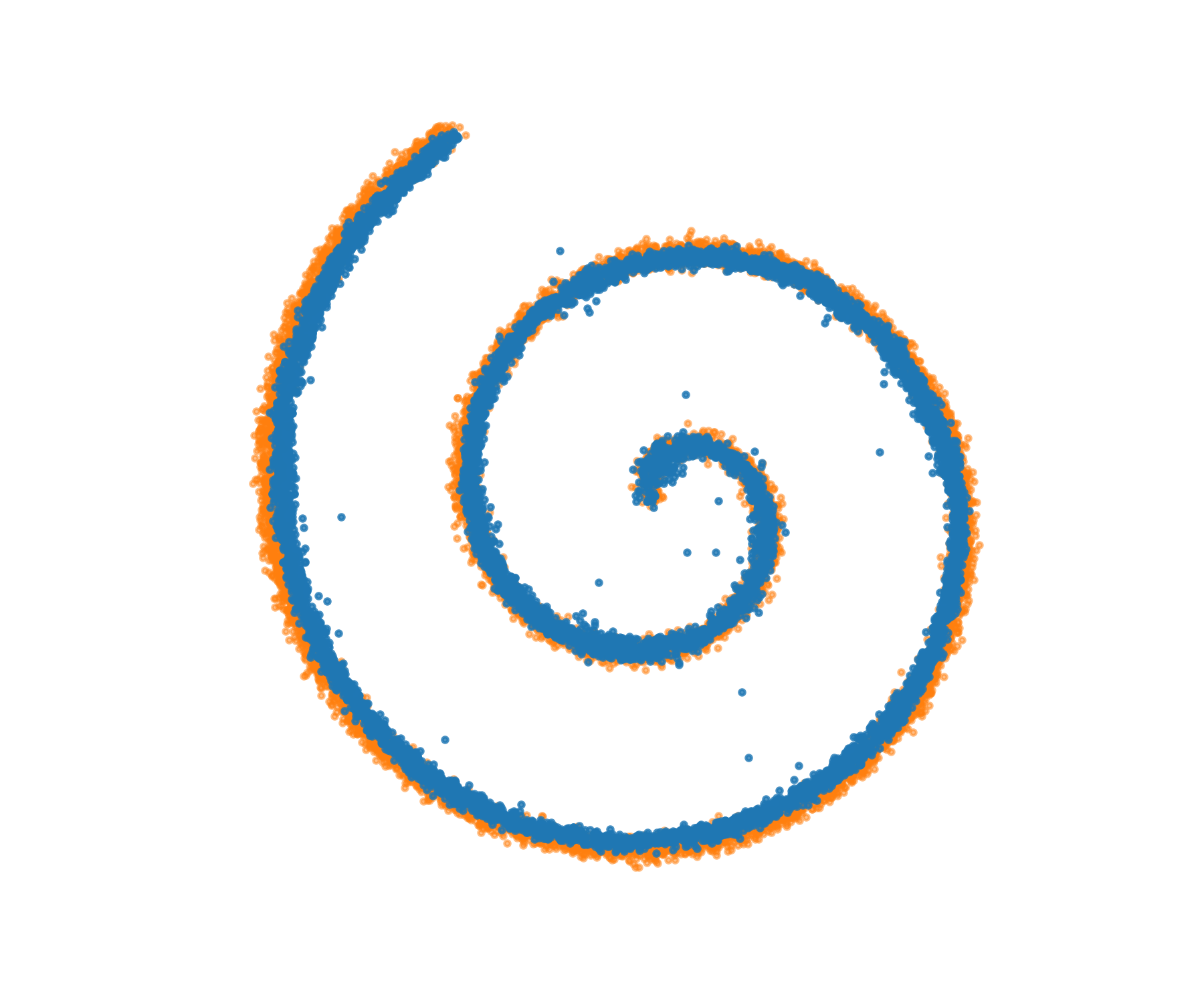} &
            \spiralplot{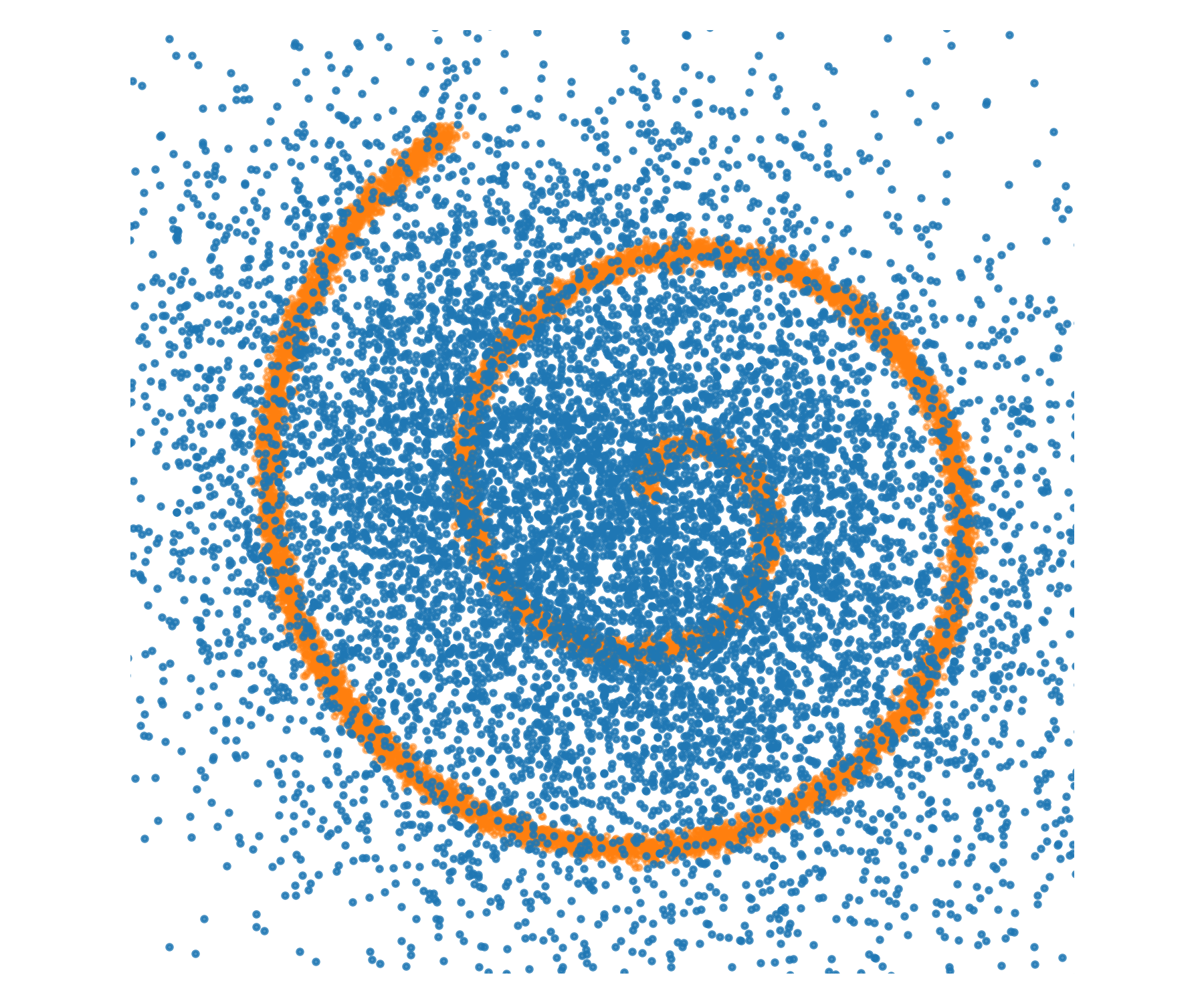} \\[-2mm]
            \spirallabel{$\rho=1$} &
            \spiralplot{thick_spiral_D256_snr1.0_sigma0.0601644010322685_x1_pred.png} &
            \spiralplot{thick_spiral_D256_snr1.0_sigma0.0601644010322685_v_pred.png} \\[-2mm]
            \spirallabel{$\rho=100$} &
            \spiralplot{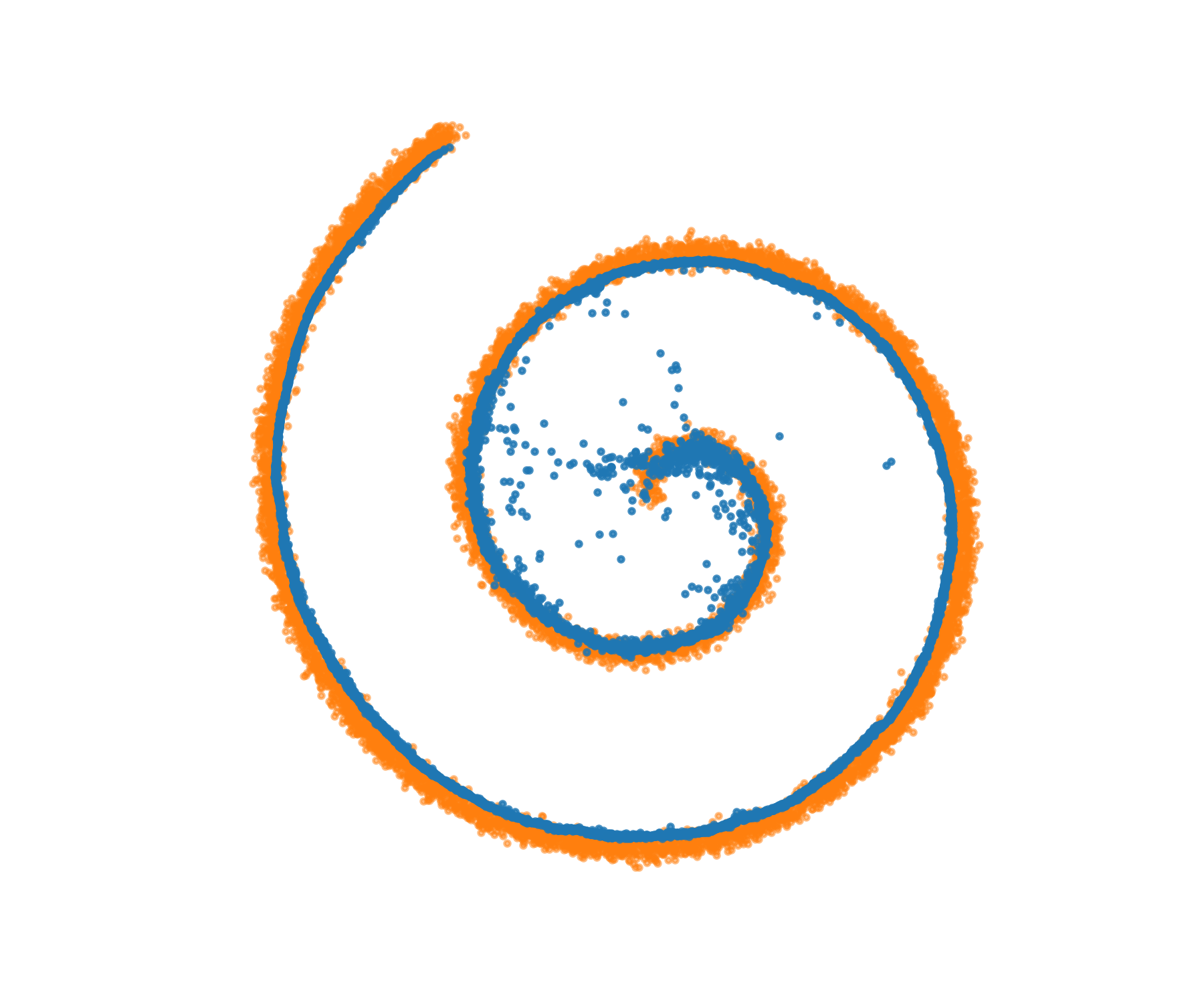} &
            \spiralplot{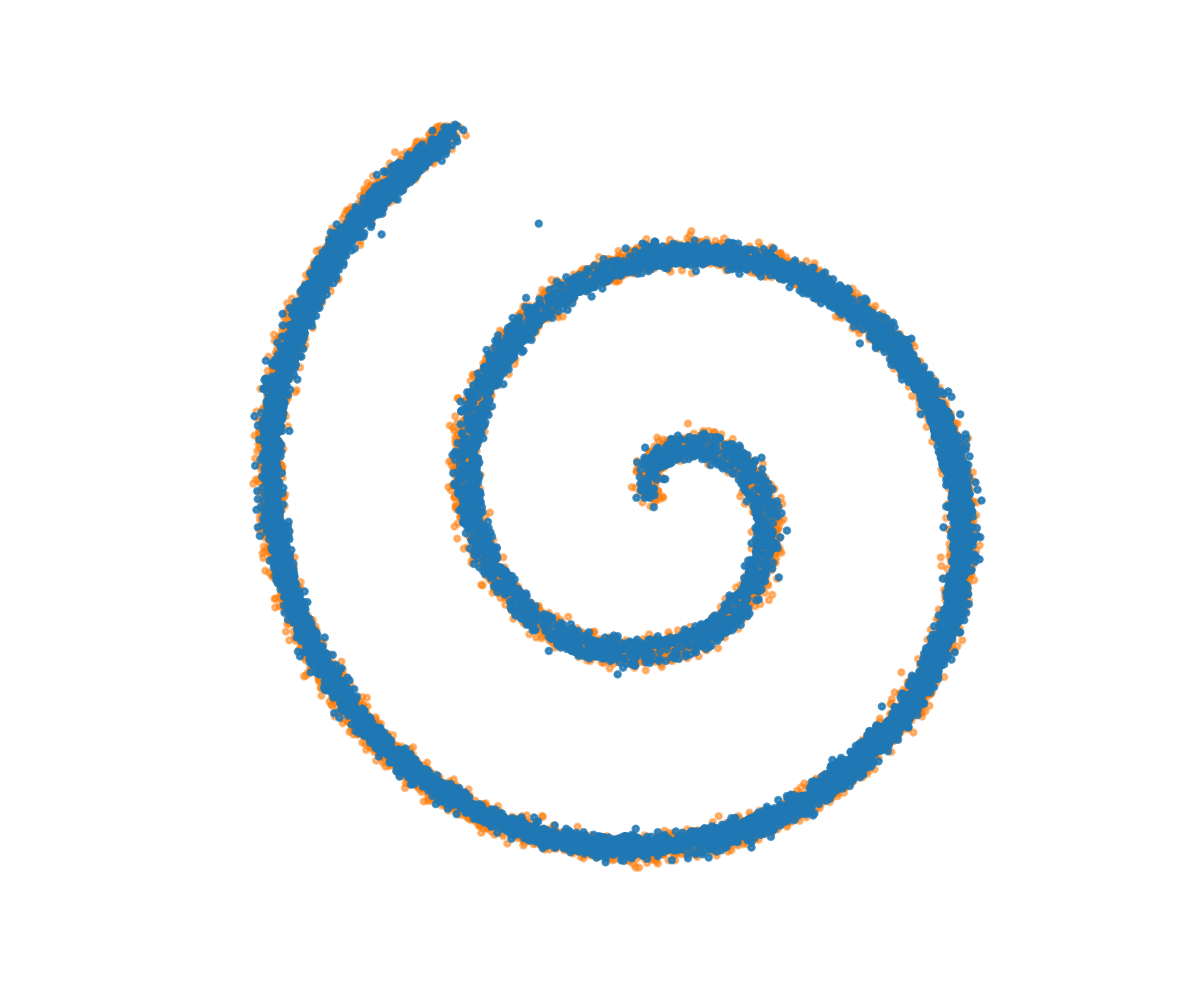}
        \end{tabular}
        }
    \end{minipage}

    \caption{
        Disentangling ambient dimensionality and source--data SNR. True data are shown in
        \textcolor{taborange}{orange} and generated samples in
        \textcolor{tabblue}{blue}. For all columns, $x_0 \sim \gN(0, \sigma_0^2 I_D)$. 
        \emph{Left:} fixed 
        $\sigma_0=1$; as the dimension $D$ increases, the source--data
        SNR $\rho$ decreases. 
        \emph{Center:} the source-data SNR $\rho=1$ is kept fixed (by adapting $\sigma_0$) as $D$ increases. 
        \emph{Right:} at a fixed $D=256$, the source--data SNR varies as
        $\rho\in\{0.01,1,100\}$, by using $\sigma_0\in\{0.6,0.06,0.006\}$.
        The conclusions of \citet{li2025back} (left panel) no longer hold when the SNR is kept fixed (middle panel). \textbf{The right panel identifies the SNR as a critical factor}.
    }
    \label{fig:spiral_dimension_snr}
\vspace*{-4mm}
\end{figure*}

A common explanation for the success of data prediction is the
\emph{manifold hypothesis} \citep{chapelle2006semi,de2022convergence}:  data is assumed to concentrate on a low-dimensional manifold of a much larger ambient space.
This hypothesis has yielded the intuition that clean data $X_1$ may be simpler than predicting the
source noise $X_0$ or the velocity $X_1-X_0$, which both contain
full-dimensional noise components \citep{li2025back,jin2026revisiting}. 
While this argument captures an important property of natural data, we show in this section that \textbf{intrinsic
dimensionality alone does not determine the best prediction target}.

Throughout this section, we assume $X_1\in L^2$ and denote its covariance by
$\Sigma_1:=\operatorname{Cov}(X_1)$, while the source is $X_0\sim\mathcal N(0,\sigma_0^2I_D)$.
Following the usual definition of SNR
as signal power divided by noise power, we define the source--data SNR
as
\begin{equation}
    \rho
    :=
    \frac{\Var X_1}{\Var X_0}
    =
    \frac{\Tr(\Sigma_1)}{D\sigma_0^2}.
    \label{eq:global_snr}
\end{equation}
The SNR $\rho$ can be changed by varying $\sigma_0$ without modifying
the data scale and geometry.

\paragraph{Disentangling dimensionality and SNR.}

We revisit the experiment of \citet[Fig.~2]{li2025back}, in which a two-dimensional spiral is
embedded into an space of dimension $D$. They report that increasing
$D$ strongly favors $x_1$-prediction over $v$-prediction, and interpret
this behavior through the increasing gap between intrinsic and ambient
dimension.
Because the
spiral is embedded isometrically, $\Tr(\Sigma_1)$ remains independent of $D$,
while Li and He use a standard Gaussian source, corresponding to
$\sigma_0=1$ ($\Var X_0 = D$). 
Therefore, increasing the ambient dimension has the hidden side effect of decreasing the source--data SNR \eqref{eq:global_snr}.

To disentangle these effects, in \Cref{fig:spiral_dimension_snr} (left), 
we reproduce the setting of \citet{li2025back} (fixed $\sigma_0$). 
In \Cref{fig:spiral_dimension_snr} (middle) we vary $D$ at fixed $\rho=1$, using
$\sigma_0^2=\Tr(\Sigma_1)/D$, showing that \emph{varying the dimension $D$ at fixed SNR has little effect on the parameterization performance}.
Finally, in \Cref{fig:spiral_dimension_snr} (right) we fix $D=256$ while varying $\sigma_0$, and observe that \emph{changing the SNR with fixed dimension has a strong impact on
the relative parameterization performance}: the source-data SNR is a critical factor in determining the optimal prediction.  

\section{From SNR-dependent prediction to spectral hybrid models}
\label{sec:hybrid}

Having highlighted that the global SNR is a single statistic that critically affects parametrization performance, we now refine this perspective by providing a time- and direction-dependent analysis.
For the theoretical analysis in this section, we use the Gaussian model
\begin{equation}
    X_1\sim\mathcal N(0,\Sigma_1),
    \qquad
    X_0\sim\mathcal N(0,\sigma_0^2 I),
\end{equation}
with $X_0$ and $X_1$ independent. For clarity, we assume throughout the paper that the data distribution is
centered and that the source is isotropic.\footnote{The extension to a non-zero data
mean $m$ is given in \Cref{app:non_centered}, and amounts to applying the same
analysis to the centered variables $X_1-m$ and $X_t-tm$}.


\subsection{A binary switch from directional SNR}
\label{sec:binary_switch}

In real anisotropic data, the energies vary
substantially across the covariance spectrum. 
To model this, we consider the spectral decomposition of the data covariance:
$\Sigma_1 = U\operatorname{diag}(\lambda_1,\ldots,\lambda_D)U^\top$, 
where $U$ has columns $u_j$.
Along direction $u_j$, the source and data
contributions to the observation
$ X_t = (1-t)X_0 + t  X_1$ have variances
$(1-t)^2\sigma_0^2$ and $t^2\lambda_j$ respectively. We therefore define the
\emph{directional} SNR at time $t$ as
\begin{equation}
    \rho_j(t)
    :=
    \frac{t^2\lambda_j}{(1-t)^2\sigma_0^2}.
    \label{eq:directional_snr}
\end{equation}
Based on~\Cref{sec:snr}, we expect the time at which the two contributions have equal energy (i.e. $\rho_j(t)$ equals $1$) to play an important role on the optimal prediction choice. 
This time is 
\begin{empheq}[box=\keyeqbox]{equation}
    t_j^{\mathrm{switch}}
    :=
    \frac{\sigma_0}{\sigma_0+\sqrt{\lambda_j}}.
    \label{eq:snr_switch}
\end{empheq}

Similarly to the global SNR case, we expect that $\rho_{j}(t) \leq 1$ should favor $x_1$-pred and $\rho_{j}(t) \geq 1$ should favor $x_0$-pred in the $j$-th direction.

\paragraph{Validation on a Gaussian model.}
We first test whether the switching time~\eqref{eq:snr_switch} predicts the
time when the performances of separately trained $x_1$- and $x_0$-prediction models cross. 
We consider centered Gaussian data in dimension $D=1024$, with
covariance eigenvalues $\lambda_j\propto j^{-2}$, normalized such that
$\sum_{j=1}^D \lambda_j=D$. For $U$, we use a DCT eigenbasis, yielding
frequency-ordered directions with strongly anisotropic variances, as commonly
observed in image data~\citep{van1996modelling}. The source is
$X_0\sim\mathcal N(0,\sigma_0^2 I_D)$. 

\begin{figure}[t]
    \centering



 \begin{subfigure}{0.45\linewidth}
    \centering
    \includegraphics[width=\linewidth]{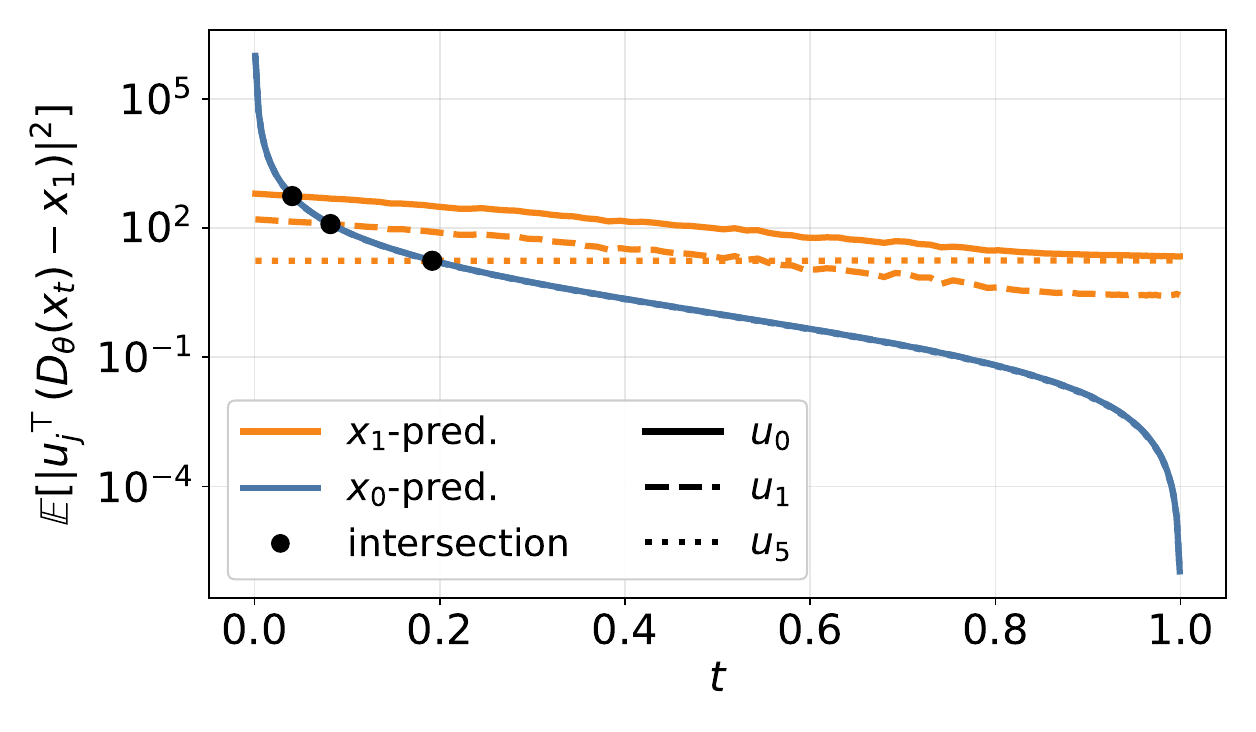
    }
    \caption{Directional MSE as a function of time}
    \label{fig:gaussian_directional_mse}
\end{subfigure}
 \begin{subfigure}{0.45\linewidth}
    \centering
    \includegraphics[width=\linewidth]{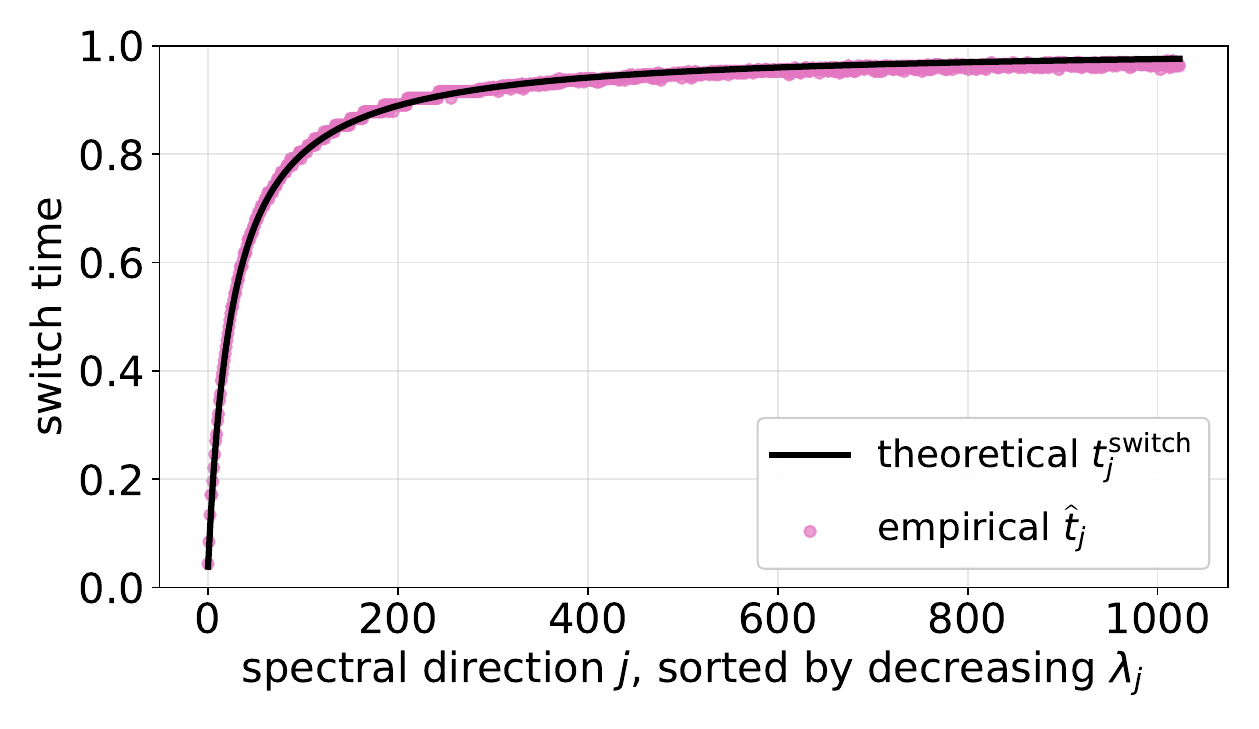
    }
    \caption{$t_{\mathrm{switch}}$: practice confirms theory.}
    \label{fig:gaussian_switch_times}
\end{subfigure}

    \caption{
       Validation of the theoretical switching time on the anisotropic Gaussian
        model, for $\rho=1$ ($\sigma_0=1$). \emph{Left:} directional MSEs of $x_0$- and $x_1$-prediction for three selected directions $u_j$ (for $x_0$-pred, the curves are superposed). The crossing (black dot) occurs earlier in high-variance directions and later in low-variance directions. 
        \emph{Right:} For each spectral direction $j$, we compare the
        theoretical switching time $t_j^{\mathrm{switch}}$ with the empirical
        transition time 
        (black dot in left figure).
        The theory agrees strongly with empirical observation. 
    }
    \label{fig:toy_spectral_switch_extremes}
\vspace*{-3mm}
\end{figure}

We train two models with the same loss~\eqref{eq:denoising_loss}: one parameterized to predict $x_0$ and
the other to predict $x_1$ (implementation details in
\Cref{app:gaussian_switch}). To compare them fairly, we use for
both the induced denoiser $D_\theta$~\eqref{eq:param_classes} and consider the \emph{directional MSE}, i.e. the average of the residual error projected
onto directions of the spectral basis, 
$\bbE \left|u_j^\top\bigl(D_\theta(X_t,t)-X_1\bigr)\right|^2$,
which we track as a function of $t$. 
Typical curves are shown in \Cref{fig:gaussian_directional_mse}; as expected from the above discussion, for each direction $j$, $x_1$-prediction outperforms $x_0$-prediction for small times (low $\rho_j$), while performance switches for larger times (high $\rho_j$).

Next, we define the \emph{empirical transition time} in direction $j$ as the
smallest time for which the error of the $x_0$-model falls below that of the
$x_1$-model. 
\Cref{fig:gaussian_switch_times} shows the alignment
between this time and $t_j^{\mathrm{switch}}$ identified in
\Cref{eq:snr_switch}: the relative performance of $x_0$- and
$x_1$-prediction changes at a time that is almost perfectly predicted by the directional SNR.


\paragraph{A first hybrid spectral model: the binary switch strategy.} This motivates us to propose a first new training method, namely to switch from $x_1$- to $x_0$-prediction depending on time, but also on eigendirection, by outputting 
\begin{empheq}[box=\purpleeqbox]{equation}
\begin{array}{c@{\qquad}l}
\text{\bfseries $y_t^{\mathrm{bin}}$-prediction:}
&
\begin{aligned}
Y_t^{\mathrm{bin}}
&:=
M_tX_1+(I-M_t)X_0,
\label{eq:binary_target} \\
\text{where} \, 
M_t
&:=
U\operatorname{diag}\!\left(
    \mathbf 1_{\{t<t_1^{\mathrm{switch}}\}},\ldots,
    \mathbf 1_{\{t<t_D^{\mathrm{switch}}\}}
\right)U^\top.
\end{aligned}
\end{array}
\end{empheq}
Let $N_\theta^{\mathrm{bin}}(x,t)$ be trained to predict
$Y_t^{\mathrm{bin}}$, i.e. $X_1$ on the subspace selected by $M_t$ and
$X_0$ on its complement. Converting the latter to $X_1$ using
$X_1=(X_t-(1-t)X_0)/t$ gives
\begin{equation}
    D_\theta^{\mathrm{bin}}(x,t)
    =
    M_tN_\theta^{\mathrm{bin}}(x,t)
    +(I-M_t)
    \frac{x-(1-t)N_\theta^{\mathrm{bin}}(x,t)}{t},
    \label{eq:binary_denoiser}
\end{equation}
that we train with the same denoising objective as in
\eqref{eq:denoising_loss}. Experiments evaluating this new prediction
parameterization are reported in \Cref{sec:experiments},  where we show
that this simple spectral hybrid is remarkably effective in practice.

\subsection{What does the binary switch parametrization minimize?}
\label{sec:binary_regression}

The switching time identified above can also be derived
as being optimal from a regression perspective in the Gaussian setting.
Recalling that $u_j$ is the $j$-th column of $U$, define
the coordinates
$ X_{1,j}=u_j^\top X_1$,
$X_{0,j}=u_j^\top X_0$, and
$X_{t,j}=u_j^\top X_t$.
From \Cref{eq:param_classes}, for all parametrizations $p \in \{x_0, x_1, v\}$, we write the directional denoiser\footnote{Such writing is also used in \citet[Eqs. 7 and 8]{karras2022elucidating}, but not direction-wise.} as
\begin{equation}
   \forall x \in \R, \quad D_{\theta,j}^p( x,t)
    =
    \underbrace{a_j^p(t) x}_{\text{analytical part}}
    +
    \underbrace{b_j^p(t)N_{\theta,j}^p( x,t)}_{\text{learnable correction}}.
    \label{eq:generic_param}
\end{equation}
For the three standard parameterizations,
$(a_j^{x_1},b_j^{x_1})=(0,1)$,
$(a_j^v,b_j^v)=(1,1-t)$, and
$(a_j^{x_0},b_j^{x_0})=(1/t,-(1-t)/t)$.
We interpret the coefficient $a_j^p(t)$ as the analytical part imposed by the
parameterization, while the network learns the remaining correction.
Let
$D_j^\star( x,t)
:=
\mathbb E[X_{1,j}\mid  X_{t,j}= x]$
denote the optimal squared-error denoiser. We define
\begin{empheq}[box=\keyeqbox]{equation}
    \mathcal R_{t,j}^p
    :=
    \mathbb E
    \left[
        \left(
            a_j^p(t) X_{t,j}
            -D_j^\star( X_{t,j},t)
        \right)^2
    \right].
    \label{eq:residual_complexity}
\end{empheq}
\textbf{The residual $\mathcal R_{t,j}^p$ measures how far the analytical
part is from the optimal denoiser, hence how much remains to be corrected}
by $N_\theta^p$ in direction $j$ at time $t$.

In the Gaussian model, recall that
$\lambda_j=\Var(\bar X_{1,j})$ and define $
    \sigma_{t,j}^2
    =
    (1-t)^2\sigma_0^2+t^2\lambda_j = \Var(X_{t,j})$.
In that setting, the optimal denoiser is linear and equals
$
    D_j^\star(\bar x,t)
    = {t\lambda_j}  / {\sigma_{t,j}^2} 
  $ (see \Cref{app:details_mmse_gaussian_case})
and therefore
$\mathcal R_{t,j}^p
=
\sigma_{t,j}^2
\left(
    a_j^p(t)-t\lambda_j/\sigma_{t,j}^2
\right)^2$.
Explicitly, we get
\begin{align}
    \mathcal R_{t,j}^{x_0}
    &=
    \frac{(1-t)^4\sigma_0^4}{t^2\sigma_{t,j}^2},
    & \mathcal R_{t,j}^{x_1}
    &=
    \frac{t^2\lambda_j^2}{\sigma_{t,j}^2},
    &
    \mathcal R_{t,j}^{v}
    &=
    \frac{(1-t)^2
    \bigl((1-t)\sigma_0^2-t\lambda_j\bigr)^2}
    {\sigma_{t,j}^2}.
    \label{eq:gaussian_residual_energies}
\end{align}
This allows comparing $x_1$- and $x_0$-prediction: 
$\mathcal R_{t,j}^{x_1}\leq\mathcal R_{t,j}^{x_0}$ if and only if
$t^2\lambda_j\leq(1-t)^2\sigma_0^2$, namely $t \leq t_j^\mathrm{switch}$. 
Hence, the residual criterion recovers
the same switching time and ordering as the directional SNR criterion in
\eqref{eq:snr_switch}: 
$x_1$-prediction is \emph{easier} for $t\leq t_j^\mathrm{switch}$ if one 
measures task hardness by the magnitude of what remains to be 
learned by the network (i.e., $\mathcal{R}_{t, j}$)\footnote{The comparison 
including $v$-prediction is given in \Cref{app:binary_regression_details}.}. 
The parametrization $D^\mathrm{bin}$ introduced in \Cref{eq:binary_denoiser} precisely exploits this finding.


\subsection{From a binary switch to continuous spectral parametrization}
\label{sec:continuous_preconditioning}

So far, we have considered only a ``binary'' model, restricting the choice between $x_1$- and $x_0$-prediction. 
We now relax this discrete choice,
by considering any possible scalar value for the coefficient $a_j^p(t)$, beyond $a_j^{x_0}(t) = 1/t$ and $a_j^{x_1}(t) = 0$.
Minimizing the residual criterion $\mathcal R_{t,j}$ in \eqref{eq:residual_complexity}
over $a_j(t)$ yields
$    a_j^\star(t) ={t\lambda_j}/{\sigma_{t,j}^2}$, 
to which we associate the linear operator
\begin{equation}
    A^\star(t)
    :=
    t\Sigma_1
    \left(
        (1-t)^2\sigma_0^2 I+t^2\Sigma_1
    \right)^{-1} 
    = U \diag\left(a_1^\star(t), \ldots, a_d^\star(t) \right) U^\top .
    \label{eq:continuous_linear_operator}
\end{equation}
Our intuition for this choice of analytical part is that it makes the network's task the easiest, as it minimizes the magnitude of what remains to be learned by the network to approach the optimal denoiser $D^\star$.
Minimizing $\mathcal R_{t,j}$ determines the analytical coefficient
$a_j^\star(t)$, but not the scale of the learned correction, since any nonzero
factor can be absorbed into the residual network. We choose the same
coefficient $b(t)=1-t$ in every direction, which gives the desired limiting
behavior: at $t=0$, $A^\star(0)=0$ and $b(0)=1$, recovering
$x_1$-prediction, while as $t\to1$, $A^\star(t)\to I$ and $b(t)\to0$,
like $x_0$-prediction. 
This yields our second proposed model, the continuous denoiser
\begin{equation}
    D_\theta^{\mathrm{cont}}(x,t)
    =
    A^\star(t)x
    +
    (1-t)N_\theta^{\mathrm{cont}}(x,t) .
    \label{eq:continuous_model}
\end{equation}
This new denoiser corresponds to the prediction target defined as
\begin{empheq}[box=\purpleeqbox]{align}
\begin{aligned}
\text{\bfseries $y_t^{\mathrm{cont}}$-prediction:}
\qquad
Y_t^{\mathrm{cont}}
&= \frac{1}{1-t}
\bigl(I-tA^\star(t)\bigr)X_1
- A^\star(t)X_0 .
\end{aligned}
\end{empheq}



The analytical term
$A^\star(t)(x)$ is the Wiener estimator of $X_1$ from
$X_t$, i.e. the optimal linear estimator under squared error.
For Gaussian data, it is also the optimal estimator. 
Hence, if the data were Gaussian, the network wouldn't have anything to do; for real data, it must learn the nonlinear correction to this covariance-based prediction.

\section{Architecture as an information bottleneck}
\label{sec:architecture}

The previous sections study prediction parameterization independently of
architectural information loss, yet previous work indicates that architecture and optimal parametrization are connected \citep{gagneux2026training}.
To characterize this connection, 
we consider architectural bottlenecks that discard
input directions.
This allows us to determine which directions each parameterization tends to preserve
and how it behaves when some are unavailable, thus explaining some previously observed failures and successes in the training of flow matching.

\paragraph{Compression changes target preference.}

\begin{wrapfigure}{r}{0.50\textwidth}
    \centering
    \vspace{-1em}
    \includegraphics[width=\linewidth]
    {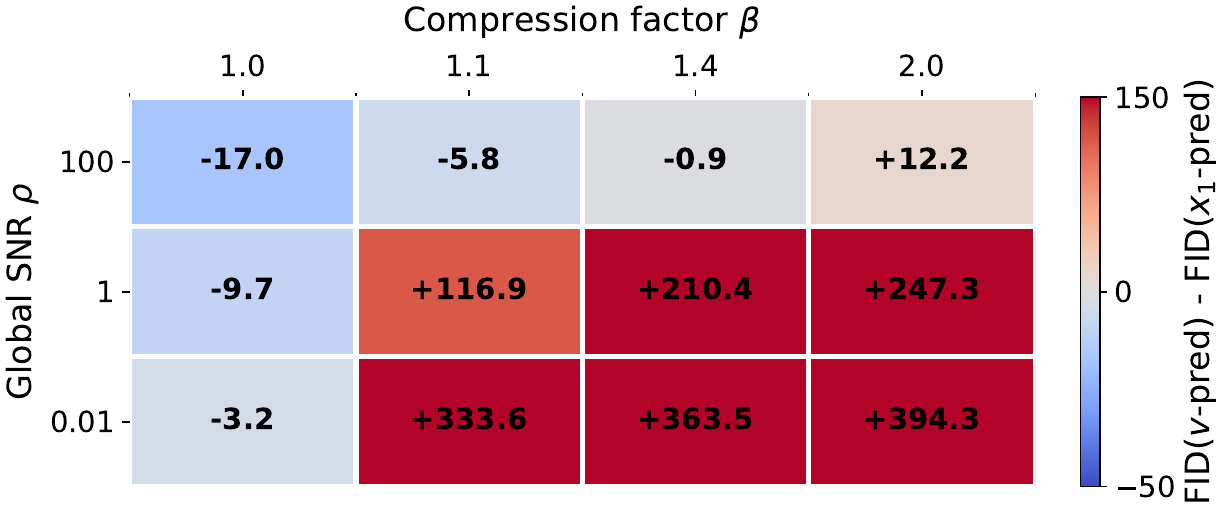}
    \caption{
    Joint effect of architectural compression and global SNR on
    CIFAR-10 generation.
    Increasing compression favors $x_1$-prediction,
    whereas increasing $\rho$ favors velocity prediction.
    Generated samples are shown in
    \Cref{fig:cifar10_images_compression}.
    \label{fig:cifar_fid_compression_snr}}
    \vspace{-1em}
    \vspace{-3mm}
\end{wrapfigure}

We first investigate empirically whether architectural bottlenecks change the
relative performance of prediction parameterizations.
Following~\citet{li2025back}, we introduce a linear bottleneck of dimension
$r$ inside each ViT patch embedding. 
A patch $X_t^{(k)}\in\mathbb R^{Cp^2}$ obtained by flattening $p \times p$ pixels with $C$ channels is mapped as
$H_t^{(k)} = B A X_t^{(k)}$, with $ A\in\mathbb R^{r\times Cp^2}$ and $B\in\mathbb R^{h\times r}$; 
the Transformer token dimension $h$ is kept fixed and only $r$ is
varied. 
We quantify the amount of compression by
$\beta := \frac{Cp^2}{r}$
with $\beta > 1$ imposing an explicit rank bottleneck.

This construction provides a controlled way of discarding information from each
patch while keeping the token dimension fixed.
Without a bottleneck, the compression ratio becomes $ (Cp^2) / h$: thus
increasing the patch size $p$ or decreasing the token dimension $h$ also increases compression.
Yet, varying either $p$ or $h$ changes the architecture while varying the bottleneck dimension $r$ does not,
enabling us to study compression \emph{in isolation}.
This setting is also motivated by the benefits of bottlenecks for generation quality observed in \citet{li2025back}.


Using the FID~\citep{heusel2017},
\Cref{fig:cifar_fid_compression_snr} reports the CIFAR-10 comparison through
the difference $\operatorname{FID}(v)-\operatorname{FID}(x_1)$: positive
values favor $x_1$-prediction and negative values favor $v$-prediction. 
Two clear effects emerge. First, at fixed $\rho$, increasing compression
systematically shifts the comparison toward $x_1$-prediction, consistently
with the observations of~\citet{li2025back}. Second, at fixed compression,
decreasing $\rho$ produces the same shift toward $x_1$-prediction, in
agreement with the SNR dependence identified in the previous section.
Architectural and training details are given in
\Cref{app:compression_snr_details}.

\paragraph{Which spectral directions does each parametrization preserve?}

We now connect this notion of compression to the spectral directional analysis conducted in \Cref{sec:hybrid}.
To understand why different prediction parameterizations react differently to
compression, we consider a controlled Gaussian experiment based on the
covariance of CIFAR-10 patches
and learn a network including a rank-$r$ linear bottleneck for each
prediction parameterization. 

\begin{figure*}[t]
    \centering

    \begin{minipage}[t]{0.48\textwidth}
        \centering
        \includegraphics[width=0.85\linewidth]{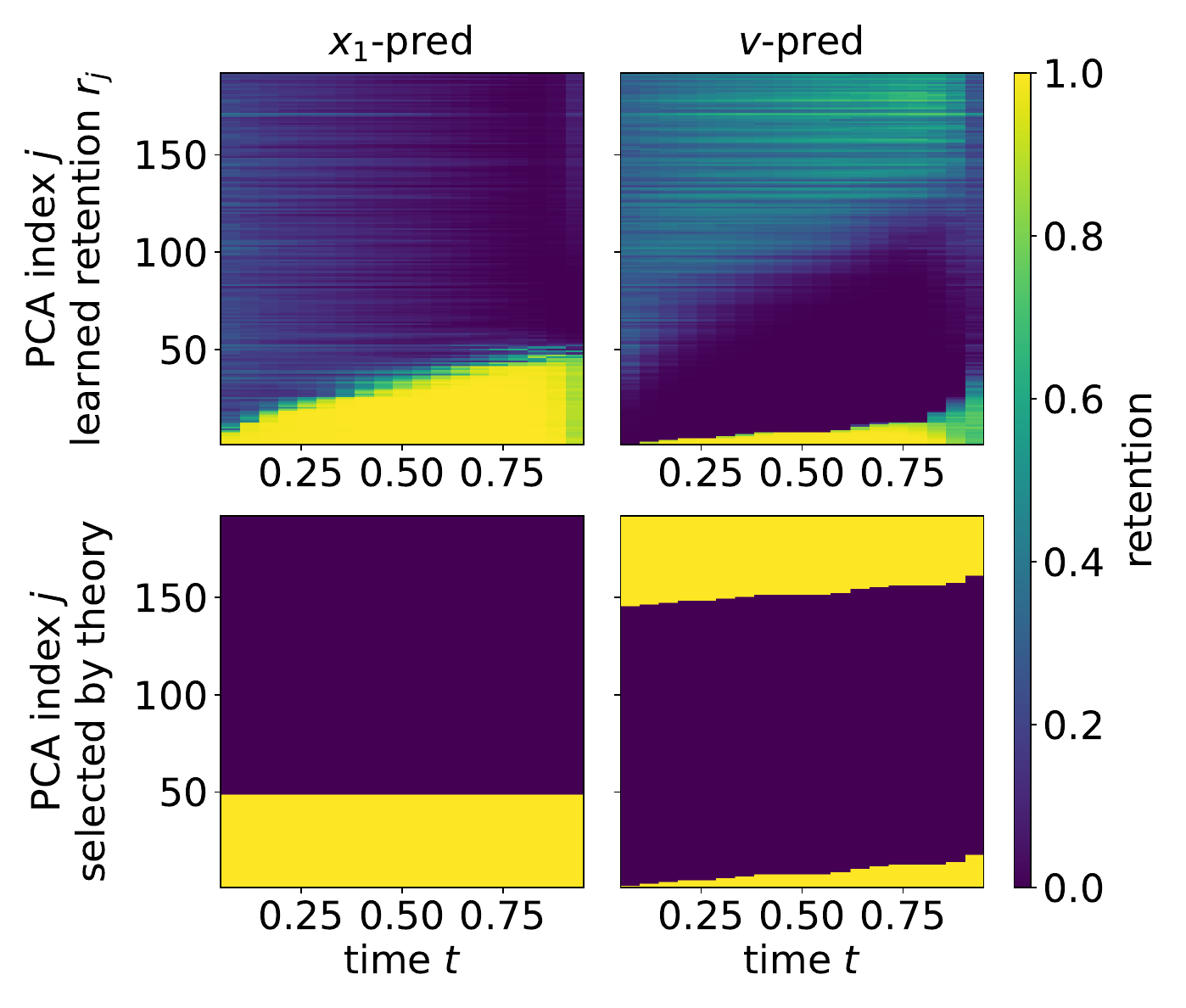}
        \captionof{figure}{
        Theoretical and learned spectral directions under a rank-$r$ bottleneck.
 \emph{Top}: retention measured from the
learned bottleneck,
$r_j^p(t)=\|P_{W_t^p}u_j\|^2$, where $P_{W_t}$ projects onto the subspace preserved by $W_t$. $r_j^p(t)\simeq1$ indicates that the
PCA direction $u_j$ is preserved. \emph{Bottom}: The $r$ PCA directions with largest residual criterion $\mathcal R_{t,j}^p$.
        }
        \label{fig:bottleneck_allocation}
    \end{minipage}
    \hfill
    \begin{minipage}[t]{0.48\textwidth}
        \centering
        \includegraphics[width=\linewidth]
        {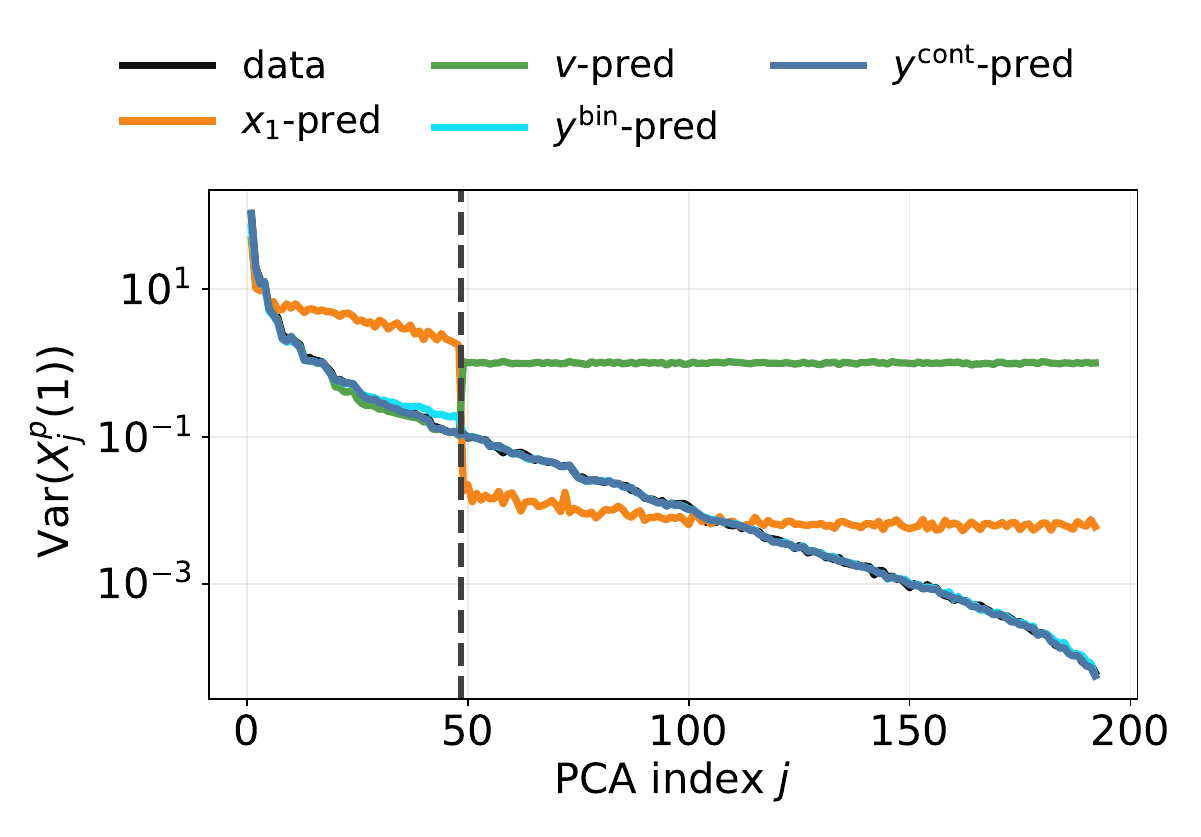}
        \captionof{figure}{
        Coordinatewise variance of generated samples using the same fixed
        rank-$48$ top-PCA bottleneck for all parameterizations. Beyond the
        cutoff, $x_1$-prediction suppresses variability and velocity
        prediction retains the source variance, whereas the spectral hybrids
        closely recover the target variance spectrum, in agreement with
        \eqref{eq:missing_direction_solution}.
        }
        \label{fig:patch_bottleneck_generation}
    \end{minipage}
\vspace*{-3mm}
\end{figure*}

In \Cref{fig:bottleneck_allocation} (row 1), we observe that the learned bottleneck depends on the target and aligns well with the PCA decomposition. \textbf{For
$x_1$-prediction, the learned bottleneck preserves the $r$ leading
PCA directions. For velocity prediction, the retained directions vary 
more strongly with time}, indicating that different parts of the spectrum are
needed at different stages of the trajectory. 
For figure with all predictions see \Cref{app:spectral_allocation_details}.

This behavior follows directly from the residual criterion
$\mathcal R_{t,j}^p$ introduced in~\eqref{eq:residual_complexity}.
Fix $t\in(0,1)$ and consider a rank-$r$ bottleneck retaining the directions
indexed by $S\subset\{1,\ldots,D\}$, with $|S|=r$. The denoising MSE
decomposes as
\begin{align*}
    \mathbb E\!\left[
        \left\|D_\theta^p(X_t,t)-X_1\right\|^2
    \right]
    &=
    \mathcal L^\star_t
    +
    \underbrace{
    \sum_{j\in S}
    \mathbb E\!\left[
        \left(
        D_{\theta,j}^p( X_{t,j},t)
        -
        D_j^\star(X_{t,j},t)
        \right)^2
    \right]}_{\text{error on retained dir.}}
    +
    \underbrace{
    \sum_{j\notin S}\mathcal R_{t,j}^p
    }_{\text{error on discarded dir.}},
    \label{eq:bottleneck_mse_decomposition}
\end{align*}
where
$\mathcal L^\star_t
:=
\mathbb E[\|D^\star(X_t,t)-X_1\|^2]$
is the irreducible denoising error and is independent of $S$.
For an ideal network, the error on retained directions vanishes.
Therefore, minimizing the fixed-time denoising MSE over $S$ amounts to
minimizing $\sum_{j\notin S}\mathcal R_{t,j}^p$, so the optimal rank-$r$
bottleneck retains the $r$ directions with largest
$\mathcal R_{t,j}^p$. Comparing the two rows of \Cref{fig:bottleneck_allocation} shows that the learned
bottlenecks closely follow this predicted retained direction.


\paragraph{What happens to discarded directions?}
The previous subsection studies which directions a bottleneck tends to retain.
We now consider the complementary question: \emph{what happens when a
direction is discarded by the bottleneck?} \\
We return to the Gaussian model and consider a PCA direction $u_j$ that is
absent from a PCA-aligned bottleneck. The network can then predict
the correction in direction $u_j$ only from the retained coordinates.
However, under the Gaussian model, PCA coordinates are independent,
so the retained coordinates contain no information about the target
in direction $u_j$. 
The network's optimal squared-error prediction in direction $j$ is therefore zero.

Let $X^p(t)$ denote the inference-time trajectory generated by the trained
model under parameterization $p$, initialized from $X^p(0)=X_0$, and let
$ X_j^p(t)$ denote its coordinate along $u_j$. Projecting the
sampling dynamics onto the discarded direction and using the zero network
prediction gives, at $t=1$ (see \Cref{app:discarded_direction_details}),
$ X_j^{x_1}(1)=0$,
$X_j^{v}(1)=X_{0,j}$,
$X_j^{\mathrm{bin}}(1)
=  X_j^{\mathrm{cont}}(1) 
=
\frac{\sqrt{\lambda_j}}{\sigma_0}X_{0,j}$.
Consequently, since $X_{0,j}\sim\mathcal N(0,\sigma_0^2)$,
\begin{equation}
    \Var\!\left( X_j^{x_1}(1)\right)=0,
    \qquad
    \Var\!\left( X_j^{v}(1)\right)=\sigma_0^2,
    \qquad
    \Var\!\left(X_j^{\mathrm{bin}}(1)\right)=
    \Var\!\left( X_j^{\mathrm{cont}}(1)\right)=\lambda_j.
    \label{eq:missing_direction_solution}
\end{equation}
Thus, $x_1$-prediction suppresses variability along a discarded direction,
collapsing its generated coefficient to 0 (i.e. the data mean), whereas
velocity prediction preserves the source variance $\sigma_0^2$.
\textbf{Remarkably, both spectral hybrids recover exactly the target variance
$\lambda_j$, although the direction is never available to the learned
residual.} \Cref{fig:patch_bottleneck_generation} shows remarkable alignment between this theory and practice, even on non-Gaussian CIFAR-10 patches. 
Experimental details and
additional controls are given in
\Cref{app:discarded_direction_details}.

\begin{remark}
The analysis above is a first step toward understanding compression in Vision
Transformers. Here, the image is treated as a single patch, so the bottleneck
and the spectral hybrid act in the same covariance basis. In a standard ViT,
the bottleneck acts in local patch space whereas the hybrid uses global image
statistics, so the exact variance-matching argument no longer applies
directly. Nevertheless, the analysis highlights mechanisms that may also
contribute to compression effects in practical ViTs.
\end{remark}

\section{Experiments}
\label{sec:experiments}
\begin{figure*}[t]
    \centering
    \begin{subfigure}[t]{0.49\textwidth}
        \centering
    \includegraphics[width=1.0\linewidth]{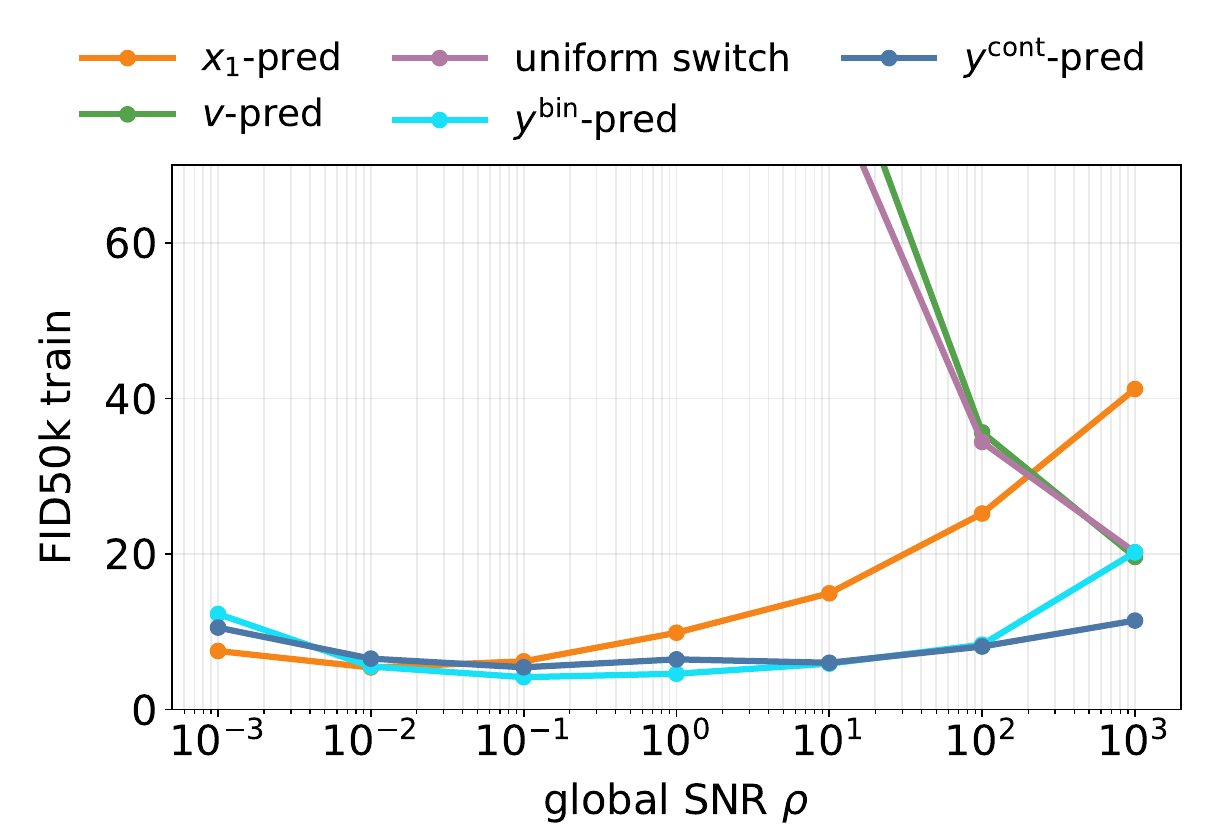}
    \caption{Final FID versus SNR (CelebA-64)}
    \label{fig:celeba_robustness}
    \end{subfigure}
    \begin{subfigure}[t]{0.49\textwidth}
        \centering
    \includegraphics[width=1.0\linewidth]{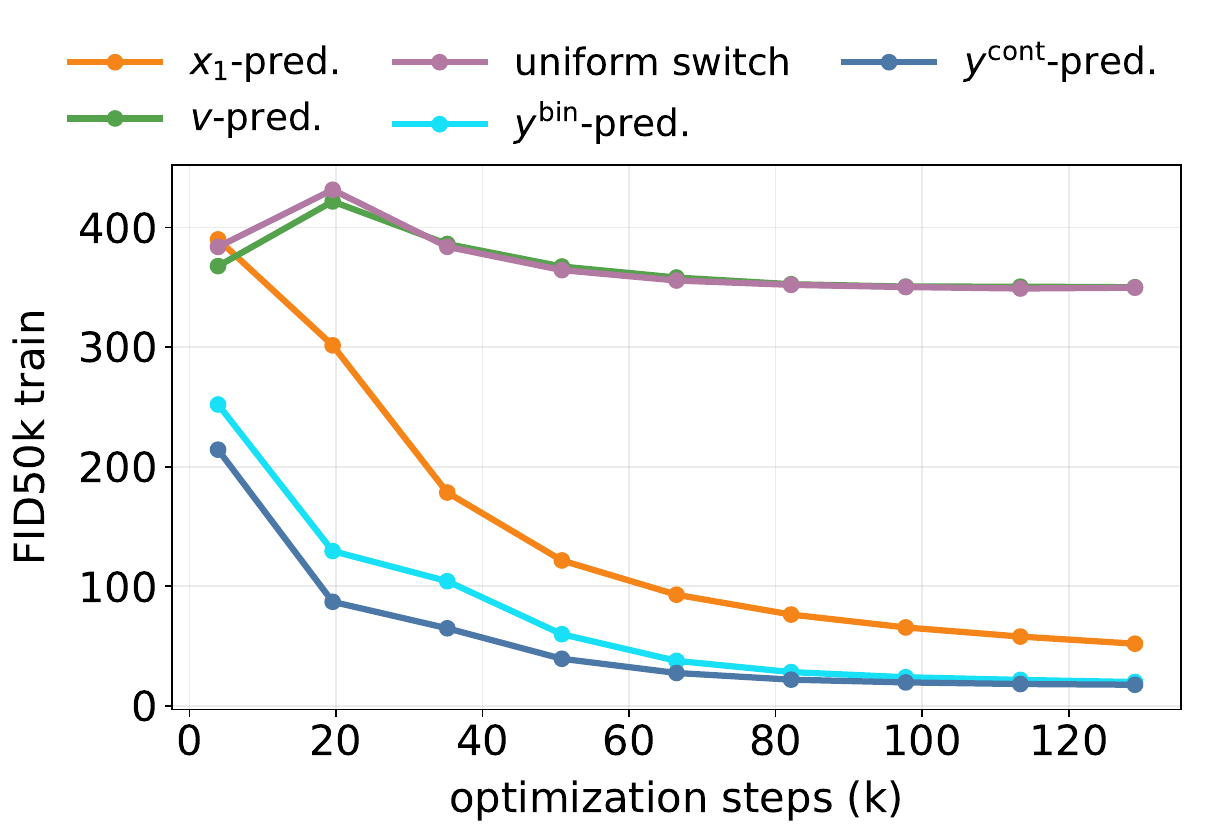}
    \caption{FID throughout training (Bedrooms-64)}
    \label{fig:bedroom_convergence}
    \end{subfigure}
    \caption{
    Generative performance (FID) 
for a ViT with patch size $4$ and compression factor $2$. \emph{Left:} On CelebA-$64$, we compare $x_1$-pred, $v$-pred, a fixed $x_1/x_0$ switch at $t=1/2$ (uniform switch), and the binary and continuous spectral parameterizations. The two hybrid parameterizations $Y^\text{bin}$ and $Y^\text{cont}$ are robust to SNR
variations and architectural compression, and in the worst case nearly match the best fixed prediction target. \emph{Right}:  FID throughout training on Bedroom-$64$. Our spectral hybrid
parameterizations 
converge substantially faster.
    }
\end{figure*}

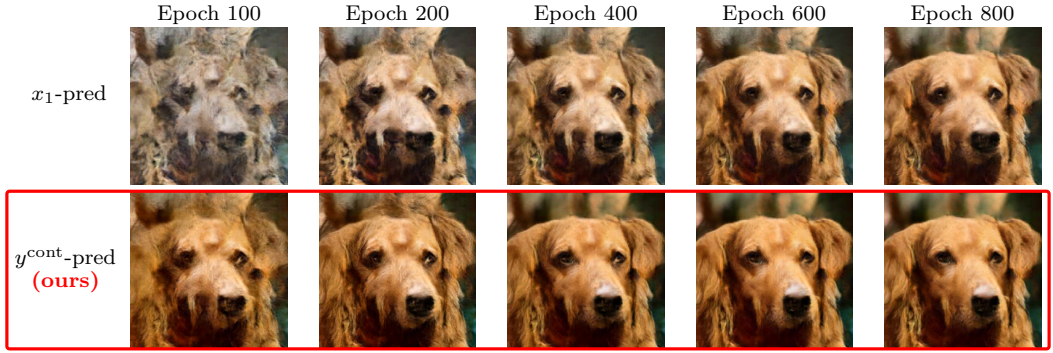
\begin{figure*}[t]
    \centering
    \scriptsize
    \begin{tabular}{@{}>{\centering\arraybackslash}m{0.09\textwidth}@{\hspace{1mm}}ccccc@{}}
        & Epoch 100 & Epoch 200 & Epoch 400 & Epoch 600 & Epoch 800 \\

        \raisebox{0.07\textwidth}[0pt][0pt]{$x_1$-pred} &
        \includegraphics[width=0.13\textwidth]{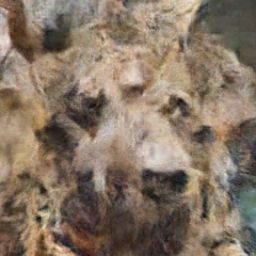} &
        \includegraphics[width=0.13\textwidth]{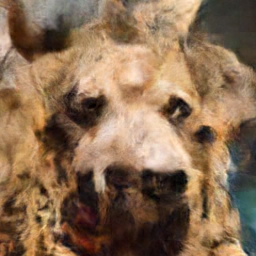} &
        \includegraphics[width=0.13\textwidth]{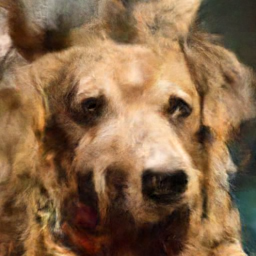} &
        \includegraphics[width=0.13\textwidth]{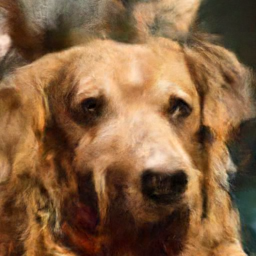} &
        \includegraphics[width=0.13\textwidth]{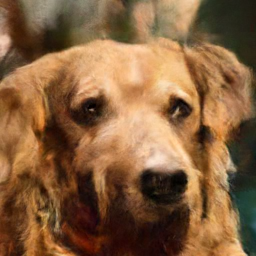} \\

        \raisebox{0.065\textwidth}[0pt][0pt]{%
            \tikz[remember picture,baseline=(afhqoursfour.center)]
            \node[inner sep=0pt,align=center] (afhqoursfour)
            {$y^\text{cont}$-pred\\[-0.3mm]\textcolor{red}{\textbf{(ours)}}};
        } &
        \tikz[remember picture,baseline=(afhqoursfourstart.base)]
            \node[inner sep=0pt] (afhqoursfourstart)
            {\includegraphics[width=0.13\textwidth]{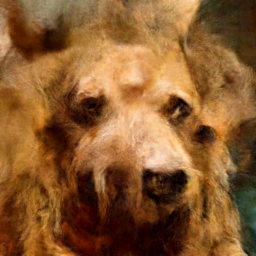}}; &
        \includegraphics[width=0.13\textwidth]{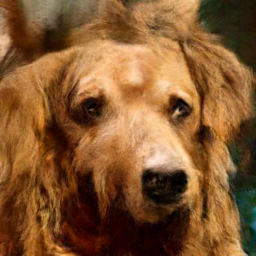} &
        \includegraphics[width=0.13\textwidth]{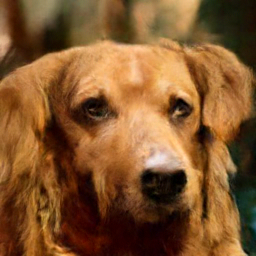} &
        \includegraphics[width=0.13\textwidth]{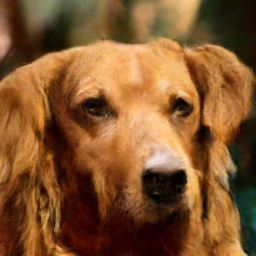} &
        \tikz[remember picture,baseline=(afhqoursfourend.base)]
            \node[inner sep=0pt] (afhqoursfourend)
            {\includegraphics[width=0.13\textwidth]{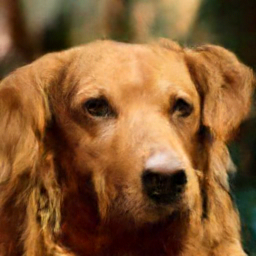}}; \\
    \end{tabular}

    \begin{tikzpicture}[remember picture,overlay]
        \draw[red,line width=1.2pt,rounded corners=1pt]
            ($(afhqoursfour.west |- afhqoursfourstart.north)+(-1mm,0mm)$) rectangle
            ($(afhqoursfourend.south east)+(1mm,0mm)$);
    \end{tikzpicture}

    \caption{
    AFHQ-$256$ samples generated from the same fixed source sample throughout
    training, using a ViT-$16$ without bottleneck. The best fixed prediction target here is $x_1$-pred.
    }
    \label{fig:afhq_epochs}
\vspace*{-4mm}
\end{figure*}

\looseness=-1
We now evaluate the proposed spectral parameterizations on standard image-generation settings.
\Cref{fig:celeba_robustness} illustrates the impact of the global SNR on the model's final performance on CelebA \citep{celeba}, and \Cref{fig:bedroom_convergence,fig:afhq_epochs} show faster training on standard image datasets Bedroom-64 \citep{yu15lsun} and AFHQ-256 \citep{choi2020starganv2}.
We use alternatively U-Net architectures \citep{ronneberger2015unet} and ViTs \citep{dosovitskiy2021vit} from the implementation of \citet{li2025back}, and use the Adam optimizer \citep{kingma2014adam}.

\paragraph{Robustness to source scale and architecture}

On CelebA-$64$, we vary only the source scale to obtain
$\rho\in\{10^{-3},10^{-2},10^{-1},1,10,10^2,10^3\}$ and compare the fixed
prediction targets with the binary and continuous spectral models. We also
include a parametrization with an $x_1/x_0$ switch at $t=1/2$ \emph{uniformly} in every spectral direction. This baseline tests whether the
gain of the proposed models comes from avoiding the ill-conditioned
$x_0$-prediction regime near $t=0$, or from their spectral adaptation, which
selects a different parameterization across directions according to their
variance. 
%
As can be seen in~\Cref{fig:celeba_robustness}; 
the spectral models remain close to the best fixed target across the SNR
range, whereas the uniform switch does not provide the same robustness. This
shows that adapting the parameterization across spectral directions is crucial.

\paragraph{Optimization speed-ups}

We next evaluate FID throughout training on a subset of Bedroom-$64$ ($100,000$ training images) at $\sigma_0=1$, for
the same U-Net and ViT architectures.
As shown in~\Cref{fig:bedroom_convergence} (full results with UNet in \Cref{fig:bedroom_convergence_full}), the hybrid predictions
$y^{\mathrm{cont}}$ and $y^{\mathrm{bin}}$ consistently converge faster than the best fixed target.
For the U-Net, it reaches the low-FID regime noticeably earlier than
$v$-prediction. For the compressed ViT, the gain is even larger: $y^{\mathrm{cont}}$ reaches
FID below $60$ after about $35$k optimization steps, whereas $x_1$-prediction needs three times as many steps to reach a
similar level.
These results are consistent with the residual construction: the analytical
spectral term reduces what remains to be learned by the network, which can
translate into faster optimization.

We finally complement this study with a higher-resolution experiment on
AFHQ-$256$, using a ViT-$16$ without bottleneck and token size $768$. 
In this setting, $x_1$-prediction is the strongest fixed target, and we compare
it directly with the continuous spectral model, implemented in a DCT basis. The results, reported in \Cref{fig:afhq_epochs}, show that the hybrid parametrization allows to recover better images and finer details, as suggested by our theory in~\Cref{sec:architecture}.

\section{Conclusion}

We showed that prediction target preference depends on both SNR and architecture,
motivating new spectral parameterizations that adapt across time and directions.
These are robust across regimes and can substantially accelerate optimization,
with no additional training cost beyond computing the data PCA once beforehand.
While PCA computation becomes difficult in large scale settings, our AFHQ experiments show that
a fixed DCT basis can provide an effective alternative.
Several questions remain open. In particular, the relative advantage of the
binary and continuous models may depend on how accurately a Gaussian
second-order approximation describes the data, as suggested by the spiral
experiments in~\Cref{fig:spiral_dimension_snr_hybrids}. 
We also deliberately decoupled parameterization from loss weighting, although the two choices may interact. 

\appendix
\newpage
\crefalias{section}{appendix}
\crefalias{subsection}{appendix}

\clearpage
\bibliographystyle{apalike}
\bibliography{biblio}

\subsection*{Fundings and acknowledgments}
This work was granted access to the HPC resources of IDRIS
 under the allocation 2026-AD011017676, 2026-AD011017183, and 2026-AD010616781 made by GENCI.

We gratefully acknowledge the support of the Centre Blaise Pascal's IT test
platform at ENS de Lyon (Lyon, France) for providing machine learning
computing facilities. The platform operates the SIDUS solution developed by
Emmanuel Quemener~\citep{quemener2014sidus}.

\subsection*{AI use statement}

In this work, we used generative AI tools for research ideation and execution,
including brainstorming on prediction parameterizations, source--data SNR,
spectral switching, and architectural compression; for checking and improving
mathematical derivations, including directional switching rules and residual
formulations; and for discussing the design and interpretation of controlled
experiments, including the spiral, PCA-bottleneck, and spectral-hybrid
experiments. We also used generative AI to prototype code for some synthetic
diagnostic experiments and to assist with drafting and restructuring parts of
the manuscript. Additionally, we used generative AI for literature discovery
and comparison with related work, and for polishing the writing. No other
required-disclosure uses are applicable to this work.

We reviewed all AI-assisted work. In particular, we compiled and checked the
final bibliography manually, verified the mathematical claims and derivations,
and checked the implementation and experimental results. We take responsibility
for the final content of this work.

\section{Background on parametrization and loss}
\label{app:param_loss_background}

\subsection{Equivalence between parametrizations and losses}
\label{app:equivalence_param_loss}
A fundamental observation is that for $X_t = (1-t) X_0 + t X_1$, the ideal velocity $\bbE[X_1 - X_0 \mid X_t=x]$, the ideal denoiser $\bbE[X_1 \mid X_t = x]$ and the ideal noise estimator $\bbE[X_0 \mid X_t=x]$ are all algebraically related to each other.
For example, the ideal velocity is linked to the ideal denoiser by:
\begin{align}\label{eq:den_to_vel}
    \bbE[X_1 - X_0  \mid X_t = x] = \bbE[\tfrac{X_1 - X_t}{1 - t} \mid X_t = x] = \frac{\bbE[X_1 \mid X_t = x] - x}{1 - t}
\end{align}
Therefore, from a network $N_\theta^{x_1}$ outputting the clean image $X_1$, one can obtain a prediction for the velocity as $\frac{N_\theta^{x_1}(x, t) - x}{1 -t}$.
As a consequence, this network can be trained by regressing against the velocity, by using as a loss:
\begin{align}
    \bbE_{x_0,x_1, t} [\Vert \tfrac{N_\theta^{x_1}(x, t) - x}{1 - t} - (x_1 - x_0) \Vert^2]
\end{align}
Interestingly, this loss rewrites $\bbE_{x_0,x_1, t} [\tfrac{1}{(1-t)^2} \Vert N_\theta^{x_1}(x, t) - x_1  \Vert^2] $: regressing against the velocity is actually regressing against the clean image under a time dependent weighting of $(1-t)^{-2}$.
The same argument applies to any pair of prediction and regression target: each reduces to a weighted regression onto the network's output, with a weight specific to the pair.

The 3$\times$3 possible combination of loss and prediction are summarized in \Cref{tab:pred_loss} (included for completeness, it  also be found e.g. in \citet[Table 1]{li2025back}).

\textbf{Derivation of the denoiser for each parametrization in} \Cref{eq:param_classes}):
Using $X_t = (1-t) X_0 + t X_1$, one can easily check that:
\begin{align}
    \bbE[X_1 \mid X_t = x] &= x + (1-t) \bbE[X_1 - X_0 \mid X_t = x] \\
    &= \frac{x - (1-t) \bbE[X_0 \mid X_t = x]}{t}
\end{align}
Hence, the natural denoiser for $v$-pred is $\mathrm{Id} + (1-t) N_\theta^v$, while for $x_0$-pred it is $\frac{\mathrm{Id} - (1-t) N_\theta}{t}$


\begin{table}[t]
\centering
\caption{Combinations of prediction and loss. The network $N_\theta^p$ outputs $p \in \{x_0, v, x_1\}$ (row); the loss regresses against another (column), possibly after algebraic conversion. Every cell reduces to a weighted regression onto the network's own output,
$\bbE_{x_0,x_1,t}\big[w(t)\,\Vert N^p_\theta(x_t,t) - p\Vert^2\big]$}
\label{tab:pred_loss}
\begin{tabular}{@{}ll ccc@{}}
\toprule
& & $x_1$-loss & $v$-loss & $x_0$-loss \\
\midrule
& $x_1$-pred & $1$                            & $\tfrac{1}{(1-t)^2}$ & $\tfrac{t^2}{(1-t)^2}$ \\[1.2ex]
& $v$-pred   & $(1-t)^2$                      & $1$                  & $t^2$ \\[1.2ex]
& $x_0$-pred & $\tfrac{(1-t)^2}{t^2}$         & $\tfrac{1}{t^2}$     & $1$ \\
\bottomrule
\end{tabular}
\end{table}

\subsection{Error amplification for diffusion paths}
\label{app:diffusion_parameterizations}

The amplification discussed in \Cref{rk:velocity_error} depends on the
interpolation path. Consider the general Gaussian interpolation
\begin{equation}
    X_t=\alpha_t X_1+\sigma_t X_0,
    \qquad
    X_0\sim\mathcal N(0,I),
\end{equation}
where, as in the main text, $t=0$ corresponds to the source distribution and
$t=1$ to the data distribution. Differentiating gives
\begin{equation}
    \dot X_t=\dot\alpha_t X_1+\dot\sigma_t X_0.
\end{equation}
The optimal velocity writes
\begin{equation}
   V^\star(x,t) = \mathbb{E}[\dot X_t \mid X_t = x] =  \frac{\dot\alpha_t}{\alpha_t}X_t + \left(\dot\sigma_t - \sigma_t\frac{\dot\alpha_t}{\alpha_t}\right)\mathbb{E}[X_0 \mid X_t = x] .
\end{equation}
For $x_0$-prediction, we therefore have
\begin{equation}
    V_\theta^{x_0}(X_t,t) = \frac{\dot\alpha_t}{\alpha_t}X_t + \left( \dot\sigma_t  - \sigma_t\frac{\dot\alpha_t}{\alpha_t}\right) N_\theta^{x_0}(X_t,t).
\end{equation}
Similarly, for $x_1$-prediction,
\begin{equation}
    V_\theta^{x_1}(X_t,t)
    =
    \frac{\dot\sigma_t}{\sigma_t}X_t
    +
    \left(
        \dot\alpha_t
        -
        \alpha_t\frac{\dot\sigma_t}{\sigma_t}
    \right)
    N_\theta^{x_1}(X_t,t).
\end{equation}
Hence errors due to training are multiplied respectively by
\begin{equation}
    c_{x_0}(t)
    =
    \dot\sigma_t
    -
    \sigma_t\frac{\dot\alpha_t}{\alpha_t},
    \qquad
    c_{x_1}(t)
    =
    \dot\alpha_t
    -
    \alpha_t\frac{\dot\sigma_t}{\sigma_t}.
\end{equation}
Consider a variance-preserving diffusion. Let $s=1-t$
denote the usual forward diffusion time and
\begin{equation}
    \bar\alpha_s
    =
    \exp\left(
        -\frac12\int_0^s \beta_r\,dr
    \right),
    \qquad
    \bar\sigma_s^2=1-\bar\alpha_s^2.
\end{equation}
where $\beta$ denotes a positive, continuous, and bounded noise-rate schedule.
Define $
    \alpha_t=\bar\alpha_{1-t}$ and $
    \sigma_t=\bar\sigma_{1-t}$.
A direct differentiation gives
\begin{equation}
    \dot\alpha_t
    =
    \frac{\beta_{1-t}}{2}\alpha_t,
    \qquad
    \dot\sigma_t
    =
    -\frac{\beta_{1-t}\alpha_t^2}{2\sigma_t},
\end{equation}
and therefore (using that $\sigma_t^2 + \alpha_t^2=1$)
\begin{equation}
    |c_{x_0}(t)|
    =
    \frac{\beta_{1-t}}{2\sigma_t},
    \qquad
    |c_{x_1}(t)|
    =
    \frac{\beta_{1-t}\alpha_t}{2\sigma_t^2}.
\end{equation}

Near $t=0$, $\sigma_t\simeq1$, and thus $x_0$-prediction does not exhibit
the $1/t$ singularity of the interpolation used in flow matching, provided that
$\beta_{1-t}$ remains bounded (e.g., the standard linear
$\beta$ schedule).. By contrast, near $t=1$, if $\beta_0>0$,
\begin{equation}
\int_0^s \beta_r\,dr
=
\beta_0 s+o(s).
\end{equation}
Hence \begin{align}
\bar\alpha_s^2
&=
\exp\left(-\beta_0 s+o(s)\right) \nonumber \\
&=
1-\beta_0 s+o(s),
\end{align}
which implies that
\begin{equation}
    \sigma_t^2
    \simeq
    \beta_0(1-t).
\end{equation}
Therefore
\begin{equation}
    |c_{x_1}(t)|
    \sim
    \frac{1}{2(1-t)}.
\end{equation}
Thus $x_1$-prediction remains ill-conditioned close to $t=1$, similarly to
the $(1-t)x_0 + tx_1$ interpolation.


\section{Additional related work}
\label{app:related_work}

\paragraph{Prediction parameterization and weighting.}
DDPMs popularized noise prediction~\citep{Ho2020}, while
\citet{salimans2022progressive} introduced $v$-prediction as a better
conditioned parameterization for progressive distillation and few-step
sampling. Velocity prediction is also the standard regression target in
flow matching~\citep{lipman2023flow}. EDM~\citep{karras2022elucidating} reorganized diffusion models into a
modular design space, separating choices related to network
preconditioning, training, and sampling.

A separate line of work has studied how the contribution of different noise
levels should be weighted during training. \citet{choi2022perception} show
that different SNR ranges correspond to qualitatively different denoising
tasks and propose to emphasize intermediate noise levels that are most
relevant to perceptual content. 
\citet{kingma2023understanding} show that many diffusion training objectives
can be written as weighted integrals of a denoising loss over noise levels.
Different objectives therefore mainly differ in how much weight they assign
to each SNR regime.

More recently, \citet{gagneux2026training} found that the weighting
$1/(1-t)^2$, corresponding to the standard flow-matching velocity loss written in
the denoising representation, provides a robust choice across several
architectures and prediction targets. In our experiments, we therefore use
the same denoising representation and time weighting for all
parameterizations, so that their comparison isolates the effect of the
regression target rather than differences induced by loss weighting.

\paragraph{Manifold and intrinsic-dimension explanations.}
\citet{li2025back} motivate direct data prediction through the
manifold hypothesis: natural images are concentrated near a low-dimensional
set, whereas source noise and velocity contain full-dimensional components.
\citet{jin2026revisiting} formalize a related idea in a linear model and
derive a dimension-dependent optimum over a continuous family of prediction
targets. Specifically, they consider
$U_k=kX_1-(1-k)X_0$, which encompasses noise-, $v$-, and data-prediction
up to scaling and sign. Their theoretical analysis considers the optimal
linear predictor of $U_k$ from $X_t$ and selects $k$ by minimizing the
resulting regression loss. In their main result, the data are assumed to lie in a $d$-dimensional subspace of $\mathbb R^D$ and to have unit variance along each direction
of this subspace and zero variance in the orthogonal directions. Under these assumptions, they obtain
$k^\star=D/(D+d)$ and interpret its dependence on $d/D$ in terms of
intrinsic versus ambient dimension.

\paragraph{Global SNR and source distribution} To the best of our knowledge, previous works have not considered SNR at the level of individual covariance directions, and we believe this perspective is new. Several works, however, have highlighted the importance of the global SNR $\rho$ for training flow-matching and diffusion models.

FasterDiT \citep{yao2024faster} highlights the importance of the signal-to-noise ratio during training, showing that rescaling the data variance can substantially affect generation quality. Their analysis focuses primarily on the distribution of SNR values induced by the interpolation schedule rather than on the prediction target. They nevertheless observe that $v$-prediction in flow matching is more robust to scale changes than noise prediction. In contrast, we show that varying the source--data SNR alone can reverse the performance ordering between $x_0$-, $x_1$-, and $v$-prediction, identifying it as a key factor governing the relative effectiveness of prediction parameterizations.

\cite{lee2026is,kim2026better} study the role of the source distribution in flow matching. \cite{lee2026is} identify source--data norm mismatch as a factor increasing the learning cost and propose rescaling the Gaussian source to match the norm of the data distribution, which is closely related to controlling the relative source--data scale captured by our SNR. Going further, \cite{kim2026better} treat the source distribution itself as a design variable and learn a condition-dependent source, emphasizing the importance of source--data directional alignment. Together, these works show that the conventional Gaussian source is not always the best choice. Their focus, however, is on the design of the source distribution under a fixed flow-matching prediction objective, rather than on how the source--data configuration affects the relative performance of different prediction parameterizations.

The source--data SNR also appears in the appendix of
\citet{jin2026revisiting}, where their optimal target parameter satisfies
\begin{equation}
k^\star
=
\frac{D}{D+\Tr(\Sigma_1)}
=
\frac{1}{1+\rho},
\qquad
\rho=\frac{\Tr(\Sigma_1)}{D}.
\end{equation}
This generalizes their dimension-dependent result and makes the dependence on the global SNR explicit. However, this dependence is partly entangled with the $k$-dependent energy of the regression target itself, since $\mathbb E\Vert U_k\Vert^2$ is minimized at the same value of $k$. In contrast, our comparisons map all parameterizations to a common denoising space and use the same time weighting, so that differences cannot be attributed simply to changes in target magnitude.

\paragraph{Covariance and spectral explanations.}
Several recent works emphasize that prediction parameterizations behave differently across covariance directions. \citet{fu2026jlt} provide a local Gaussian analysis showing that, in low-data-variance directions, the statistics of $x_1$- and $v$-prediction differ substantially because the velocity target retains a source-noise component. Importantly, this does not imply that $x_1$-prediction is intrinsically easier simply because its raw target variance is smaller. In our analysis, all parameterizations are first mapped to a common denoising representation, and we compare the residuals that each parameterization requires the network to learn. This reveals that the relative difficulty is governed by the directional source--data SNR rather than by the raw variance of the regression target alone.

Related works exploit spectral representations or frequency-dependent
training to better match the anisotropic structure of image data.
\citet{finn2026score} derive an analytically tractable approximation of the
score in a wavelet basis, relating denoising behavior to data moments and
architectural inductive biases. DCTdiff~\citep{ning2025dctdiff} performs
diffusion directly in the DCT domain and interprets generation as a
coarse-to-fine spectral process, while \citet{esteves2026spectrally} use the
image power spectrum to design noise schedules.
Similarly, \citet{guo2026pixel} progressively reveal finer image information
through time-dependent transformed targets, and
\citet{degeorge2026balancing} rebalances the flow-matching objective across
frequencies. Together, these works show that spectral components can have
very different denoising and optimization behavior, although they do not
study how this heterogeneity changes the relative performance of prediction
parameterizations.

A Gaussian approximation closely related to the one underlying our continuous hybrid parameterization was studied by \citet{wang2024unreasonable}. They show that such an approximation captures a substantial part of the score dynamics at moderate-to-high noise levels and use the resulting closed-form dynamics to accelerate sampling. In contrast, in our case, the Gaussian analysis is instead used to derive a covariance-dependent prediction parameterization that is built directly into training.

\paragraph{Hybrid prediction strategies.}
\citet{benny2022dynamic} jointly predict clean data and noise and learn a
time-dependent combination of the corresponding denoising updates, motivated
by their complementary behavior at different stages of the diffusion
trajectory. The combination is global across image directions, whereas our
hybrid rule can vary both with time and across covariance directions. More recently, \citet{han2026selfconsistent} train a single model to
predict $x_0$, $x_1$, and $v$ simultaneously in flow matching, while
enforcing consistency between the three predictions. At sampling time, they use the velocity reconstructed from the $x_0$- and $x_1$-predictions during one part of the trajectory, and switch to the directly predicted velocity after a fixed time.
Our hybrid parameterizations instead derive this complementarity from source--data statistics.

\paragraph{Architecture and representation bottlenecks.}
Prediction target also interacts with architectural inductive bias.
\citet{gagneux2026training} observed that replacing a U-Net by a
patch-based Transformer can reverse the ordering between $x_1$- and
$v$-prediction. This is consistent with broader analyses showing that neural
denoisers are constrained by locality, equivariance, receptive fields, and
representation bottlenecks~\citep{kamb2024analytic}. \Cref{sec:architecture}
focuses specifically on the information removed by a rank-constrained first
representation.



\section{Additional Gaussian analysis for the spectral switch}
\label{app:gaussian_switch}

\subsection{Directional MSE curves and switching time}
\label{app:gaussian_mse_curves}

For this experiment, we use the synthetic anisotropic
Gaussian setup described in the main text. The data endpoint is sampled as
$x_1 \sim \mathcal N(0,\Sigma_1)$, with eigenvalues
$\lambda_j \propto j^{-2}$ normalized so that
$\sum_{j=1}^D \lambda_j = D$, and with eigenvectors given by a 2D-DCT basis.
We use dimension $D=1024$ and train
MLP denoisers with four hidden layers of width
$512$ for $20{,}000$ Adam steps,
batch size $512$, and learning rate $10^{-3}$.

\begin{figure}[t]
    \centering



 \begin{subfigure}{0.45\linewidth}
    \centering
    \includegraphics[width=\linewidth]{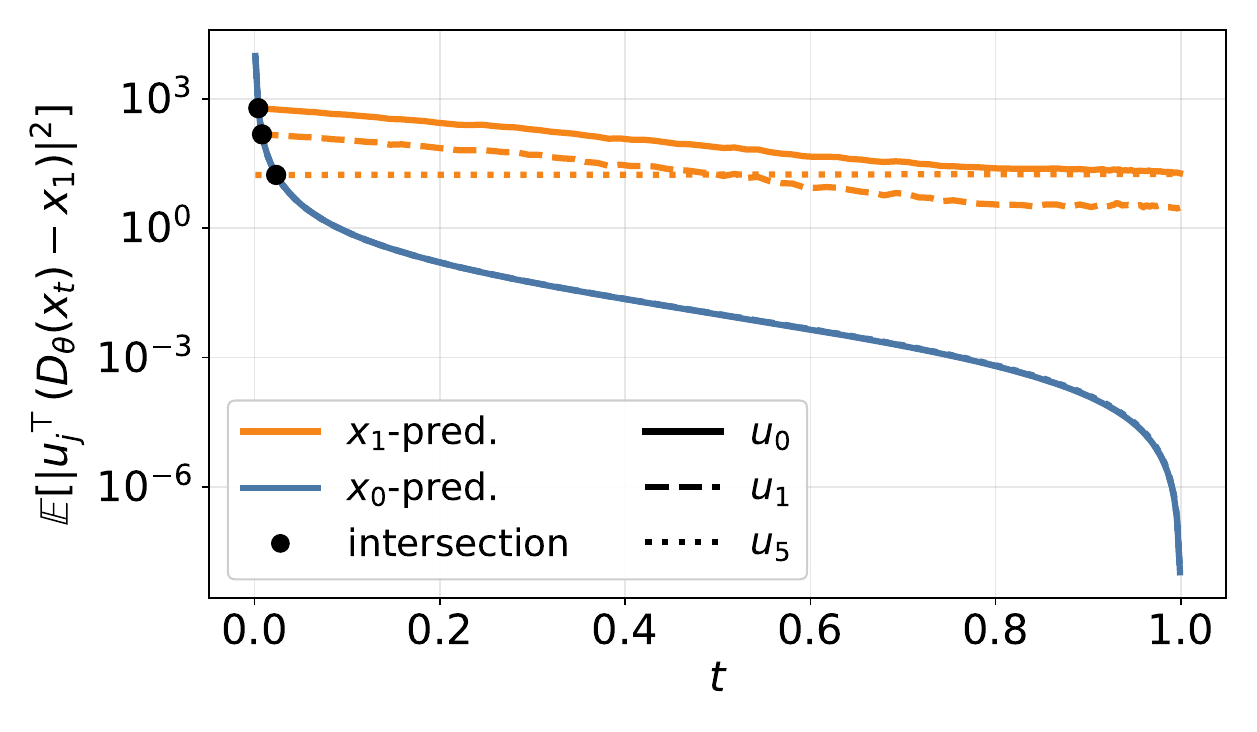}
    \caption{Directional MSE as a function of time}
    \label{fig:gaussian_directional_mse_2}
\end{subfigure}
 \begin{subfigure}{0.45\linewidth}
    \centering
    \includegraphics[width=\linewidth]{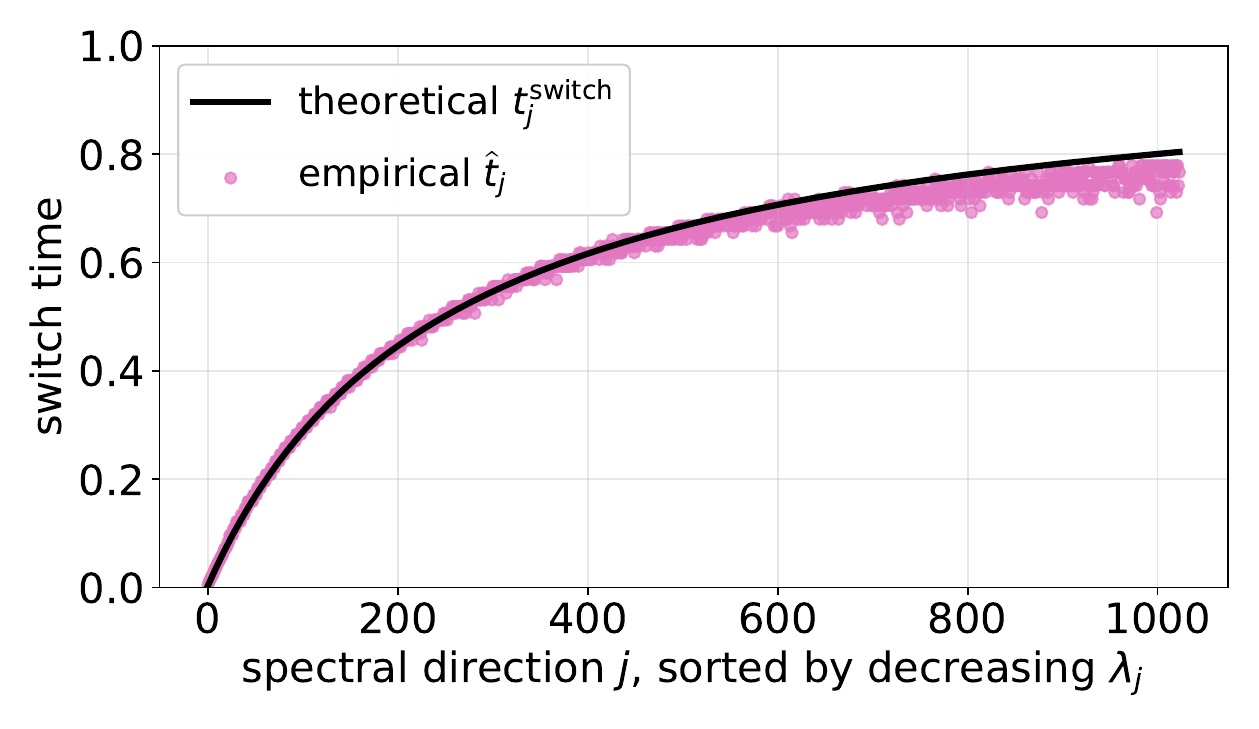}
    \caption{$t_{\mathrm{switch}}$: practice confirms theory.}
    \label{fig:gaussian_switch_times_2}
\end{subfigure}

    \caption{
       Validation of the theoretical switching time on the anisotropic Gaussian
        model, for $\rho=100$ (i.e., $\sigma_0^2=0.01$). \emph{Left:} directional MSEs of $x_0$- and $x_1$-prediction for three selected directions $u_j$. The crossing (black dot) occurs earlier in high-variance directions and later in low-variance directions. 
        \emph{Right:} For each spectral direction $j$, we compare the
        theoretical switching time $t_j^{\mathrm{switch}}$ with the empirical
        transition time at which, in this direction, the denoising error of the $x_0$-prediction
        model becomes smaller than that of the $x_1$-prediction model (black dot in left figure).
        The theory agrees strongly with empirical observation. 
    }
    \label{fig:toy_spectral_switch_extremes_2}
\end{figure}

\Cref{fig:toy_spectral_switch_extremes_2} complements
\Cref{fig:toy_spectral_switch_extremes} by showing the same plots in the case of a higher global SNR. We see that the same conclusions hold.

It is worth noting that the $x_0$-prediction curves vary less across
directions than the $x_1$-prediction curves. This is expected: the source
distribution is isotropic, so $x_0$-prediction is trained to recover a target
with the same variance in every direction, whereas $x_1$-prediction directly
inherits the anisotropic spectrum of $\Sigma_1$. More generally, the dependence
of $x_0$-prediction on the source covariance mirrors that of $x_1$-prediction
on the data covariance.

\subsection{Residual comparison including velocity prediction}
\label{app:binary_regression_details}

For completeness, we extend the residual criterion of
Section~\ref{sec:binary_regression} to velocity prediction. Under the
Gaussian model with isotropic source,
\[
    \sigma_{t,j}^2=(1-t)^2\sigma_0^2+t^2\lambda_j,
\]
and the residual quantities are
\begin{align}
    \mathcal R_{t,j}^{x_1}
    &=
    \frac{t^2\lambda_j^2}{\sigma_{t,j}^2},
    &
    \mathcal R_{t,j}^{v}
    &=
    \frac{(1-t)^2
    \bigl((1-t)\sigma_0^2-t\lambda_j\bigr)^2}
    {\sigma_{t,j}^2},
    \\
    \mathcal R_{t,j}^{x_0}
    &=
    \frac{(1-t)^4\sigma_0^4}
    {t^2\sigma_{t,j}^2}.
\end{align}

Comparing these three expressions shows that velocity prediction is preferred
between the regimes dominated by $x_1$- and $x_0$-prediction. More precisely,
\begin{equation}
    \operatorname*{argmin}_{p\in\{x_1,v,x_0\}}
    \mathcal R_{t,j}^p
    =
    \begin{cases}
        x_1, & t<t_{-,j},\\
        v, & t_{-,j}\leq t\leq t_{+,j},\\
        x_0, & t>t_{+,j},
    \end{cases}
    \label{eq:ternary_residual_rule}
\end{equation}
where
\begin{equation}
    t_{-,j}
    =
    1-\sqrt{\frac{\lambda_j}{\sigma_0^2+\lambda_j}},
    \qquad
    t_{+,j}
    =
    \sqrt{\frac{\sigma_0^2}{\sigma_0^2+\lambda_j}}.
\end{equation}
Thus $x_1$-prediction is favored near the source, $x_0$-prediction near the
data, and velocity prediction provides an intermediate regime.

\subsection{Comparison of the binary and continuous parameterizations}

\begin{figure}
\begin{center}
\includegraphics[width=0.49\linewidth]{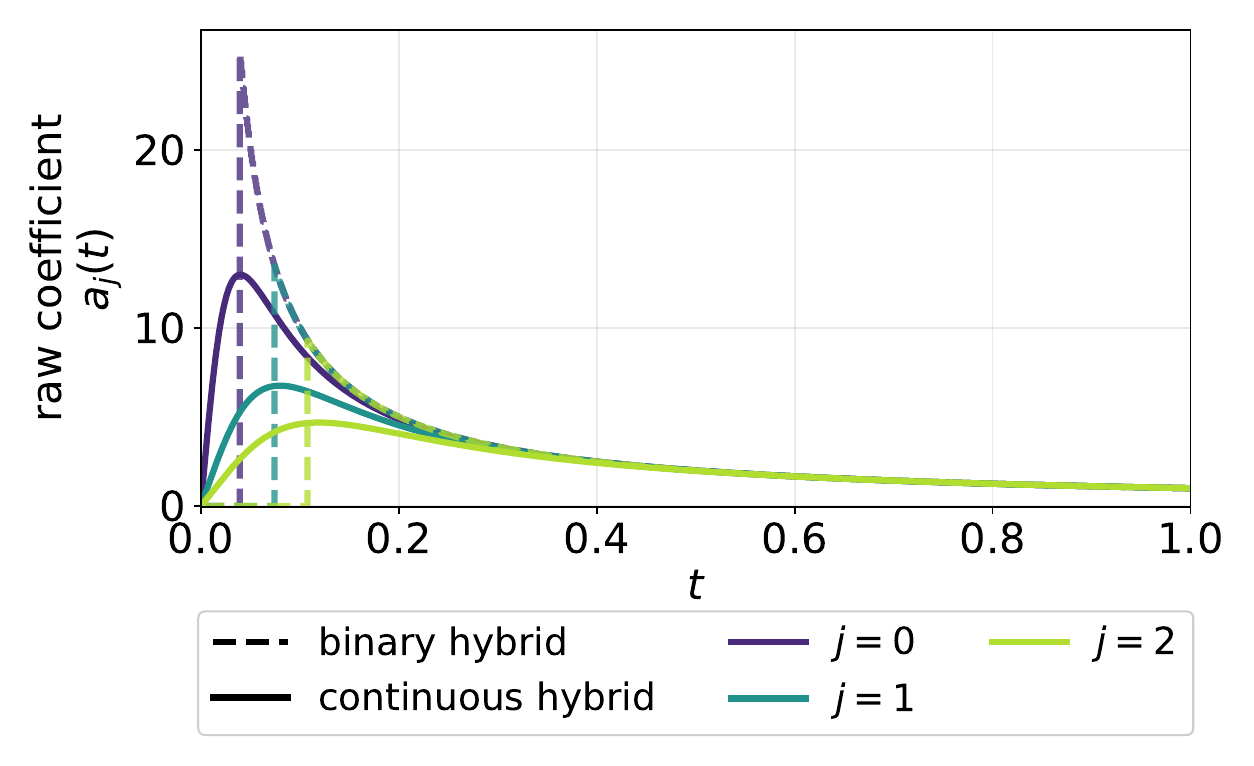}
\includegraphics[width=0.49\linewidth]{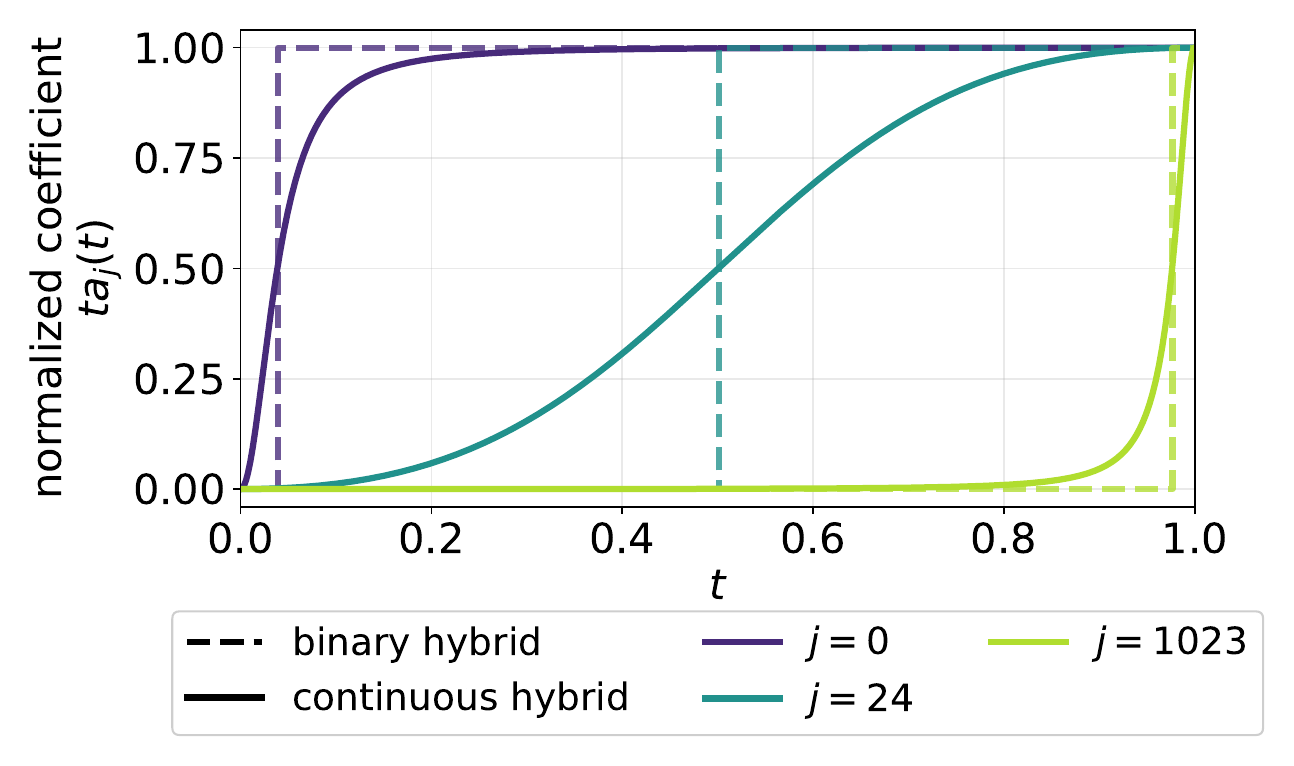}
\end{center}
\caption{Binary versus continuous spectral parameterizations for selected
directions. Left: analytical coefficient $a_j(t)$. Right: normalized
coefficient $\alpha_j(t)=t a_j(t)$, highlighting the hard binary switch and
its smooth continuous relaxation.}
\label{fig:binary_vs_continuous}
\end{figure}

The continuous model can be viewed as a smooth relaxation of the binary one.
As can be seen in \Cref{fig:binary_vs_continuous}, for each spectral direction, the binary model switches abruptly between
$x_1$- and $x_0$-prediction, while the continuous coefficient varies smoothly
between the same two regimes.

\subsection{Details on the MMSE estimator in the Gaussian case}
\label{app:details_mmse_gaussian_case}

We derive here the MMSE estimator used in the Gaussian analysis. 
We first consider the one-dimensional case. Let
$$
X_1\sim\mathcal N(m_j,\lambda_j),
\qquad
X_0\sim\mathcal N(0,\sigma_0^2),
\qquad
X_0\perp X_1,
$$
Conditionally on $X_1=x_1$,
$$
X_t\mid X_1=x_1
\sim
\mathcal N\!\left(
tx_1,(1-t)^2\sigma_0^2
\right).
$$
Let $s_t^2=(1-t)^2\sigma_0^2$. By Bayes' rule,
$
p(x_1\mid x_t)
\propto
p(x_t\mid x_1)p(x_1),
$
so that, up to terms independent of $x_1$,
\begin{align}
-\log p(x_1\mid x_t)
&=
\frac{(x_1-m_j)^2}{2\lambda_j}
+
\frac{(x_t-tx_1)^2}{2s_t^2}
+\mathrm{cst} \nonumber \\
&=
\frac12
\left(
\frac{1}{\lambda_j}
+
\frac{t^2}{s_t^2}
\right)x_1^2
-
\left(
\frac{m_j}{\lambda_j}
+
\frac{t}{s_t^2}x_t
\right)x_1
+\mathrm{cst}.
\end{align}
Completing the square therefore gives
\[
\mathbb E[X_1\mid X_t=x_t]
=
\frac{
\frac{m_j}{\lambda_j}
+
\frac{t}{s_t^2}x_t
}{
\frac{1}{\lambda_j}
+
\frac{t^2}{s_t^2}
}.
\]

After rearranging,
\begin{equation}
\mathbb E[X_1\mid X_t=x_t]
=
m_j+
\frac{t\lambda_j}{
(1-t)^2\sigma_0^2+t^2\lambda_j
}
(x_t-tm_j).
\label{eq:1d_gaussian_conditional_mean}
\end{equation}
Introducing the centered variables
$X_t=X_t-tm_j,
$
and defining
$
\sigma_{t,j}^2
=
(1-t)^2\sigma_0^2+t^2\lambda_j,
$
we obtain
\begin{equation}
D_j^\star(\bar x,t)
=\mathbb E[\bar X_1\mid \bar X_t=\bar x]
=
\frac{t\lambda_j}{\sigma_{t,j}^2}\bar x.
\label{eq:1d_gaussian_mmse}
\end{equation}

We now return to the $D$-dimensional setting. Let
$$
X_1\sim\mathcal N(m,\Sigma_1),
\qquad
X_0\sim\mathcal N(0,\sigma_0^2I),
$$
and diagonalize
$
\Sigma_1=U\Lambda U^\top$, $
\Lambda=\operatorname{diag}(\lambda_1,\ldots,\lambda_D).
$
In the centered eigenbasis,
$
\bar X_t=U^\top(X_t-tm),
$
the coordinates are independent and
$$
\bar X_{1,j}\sim\mathcal N(0,\lambda_j),
\qquad
\bar X_{t,j}=(1-t)\bar X_{0,j}+t\bar X_{1,j}.
$$

Applying \eqref{eq:1d_gaussian_mmse} coordinatewise yields
\begin{equation}
D_j^\star(\bar x_j,t)
=
\frac{t\lambda_j}{\sigma_{t,j}^2}\bar x_j
\end{equation}
Equivalently,
\begin{equation}
D^\star(\bar x,t)
=
t\Lambda
\left(
(1-t)^2\sigma_0^2I+t^2\Lambda
\right)^{-1}
\bar x.
\end{equation}
Transforming back to the original coordinates recovers
\begin{equation}
\mathbb E[X_1\mid X_t=x_t]
=
m+
t\Sigma_1
\left(
(1-t)^2\sigma_0^2I+t^2\Sigma_1
\right)^{-1}
(x_t-tm).
\end{equation}

\section{Additional details and results on architectural compression}
\label{app:architecture_details}

\subsection{CIFAR-10 compression and SNR sweep}
\label{app:compression_snr_details}

For the CIFAR-10 sweep in
\Cref{fig:cifar_fid_compression_snr}, we vary both the compression ratio
$\beta$ and the source variance $\sigma_0^2$, and hence the global
source--data SNR $\rho$. We consider patch bottleneck dimensions
$d_b \in \{48,43,34,24\}$, corresponding to
$\beta=48/d_b \approx \{1.0,1.1,1.4,2.0\}$, and
$\rho \in \{100,1,0.01\}$.
For each configuration, $x_1$- and $v$-prediction models use the same
architecture and optimization procedure. All models are trained for
$1000$ epochs using a ViT with $4\times4$ patches, a token dimension
of $768$, $12$ Transformer blocks, $6$ attention heads, and an MLP
expansion ratio of $4$. We use Adam with a batch size of $128$ and
a learning rate of $2\times10^{-4}$, preceded by a linear warm-up
over $5000$ optimization steps and held constant thereafter.
Gradients are clipped to a maximum norm of $1$.
We use exponential moving average (EMA) of
the model parameters with decay $0.9999$, starting after $1000$
optimization steps.

\begin{figure*}[t]
    \centering
    \begin{subfigure}[t]{0.49\textwidth}
        \centering
        \includegraphics[width=\linewidth]{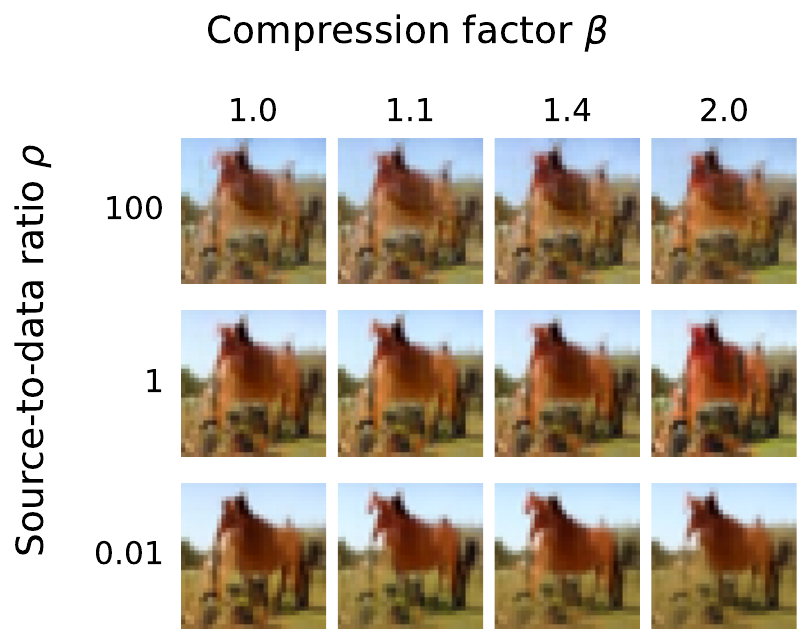
        }
        \subcaption{$x_1$-prediction}
    \end{subfigure}
    \hfill
    \begin{subfigure}[t]{0.49\textwidth}
        \centering
        \includegraphics[width=\linewidth]{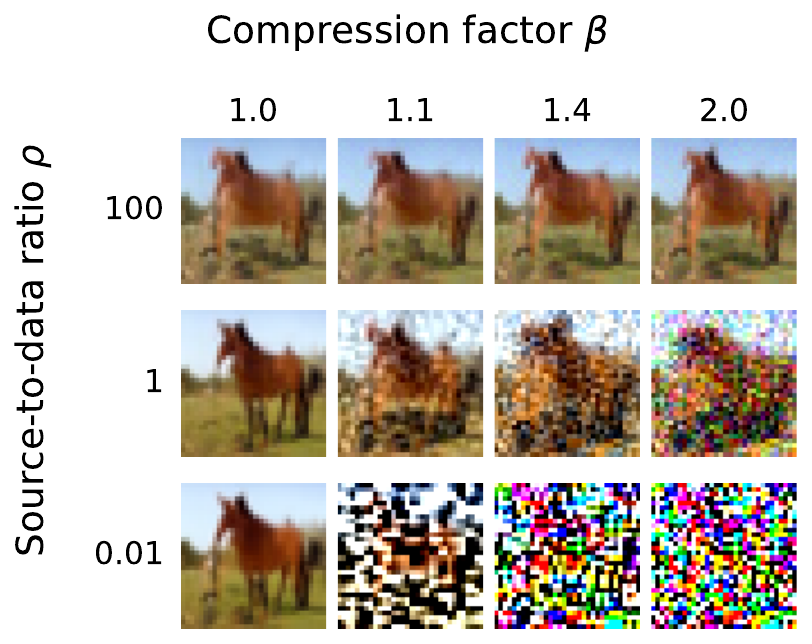
        }
        \subcaption{$v$-prediction}
    \end{subfigure}

    \caption{
        Representative CIFAR-10 samples generated by a ViT with
        a patch bottleneck, for different source-to-data ratios $\rho$ and
        compression factors $\beta$. Each panel shows the same generated sample
        index across the sweep, comparing $x_1$-prediction and $v$-prediction.
    }
    \label{fig:cifar10_images_compression}
\end{figure*}
\subsection{Learned and theoretical spectral allocation}
\label{app:spectral_allocation_details}

We provide additional details for the spectral-allocation analysis of
\Cref{sec:architecture}. We estimate the covariance of
$8\times8$ RGB CIFAR-10 patches ($D=192$) and construct the Gaussian
distribution
\begin{equation}
    X_1^{\rm G}\sim\mathcal N(0,\Sigma_1),
    \qquad
    \Sigma_1
    =
    U\operatorname{diag}(\lambda_1,\ldots,\lambda_D)U^\top,
\end{equation}
with isotropic source
$X_0\sim\mathcal N(0,\sigma_0^2I)$.

\paragraph{Learned bottlenecks at fixed time.}
For several fixed values of $t$, we independently train an MLP whose first layer
is a learned rank-$48$ bottleneck
$W_t^p\in\mathbb R^{48\times192}$. These separate bottlenecks are used only
as a diagnostic: they reveal which subspace each parameterization would
prefer to preserve at a given time.

This learned allocation can be compared directly with the residual criterion
$\mathcal R_{t,j}^p$ from~\eqref{eq:residual_complexity}. For a PCA-aligned
bottleneck retaining a set of directions $\{u_j, j \in S\}$ indexed by the set $S$, the residual that cannot be corrected by the
network is
\begin{equation}
    \sum_{j\notin S}\mathcal R_{t,j}^p.
    \label{eq:discarded_residual_appendix}
\end{equation}
Thus, under a rank constraint, the optimal bottleneck retains the directions
with largest $\mathcal R_{t,j}^p$. In particular,
$\mathcal R_{t,j}^{x_1}$ is increasing in $\lambda_j$, so $x_1$-prediction
favors the leading PCA directions, whereas the ordering for velocity
prediction changes with time.

\begin{figure*}[t]
    \centering
    \includegraphics[width=\textwidth]
    {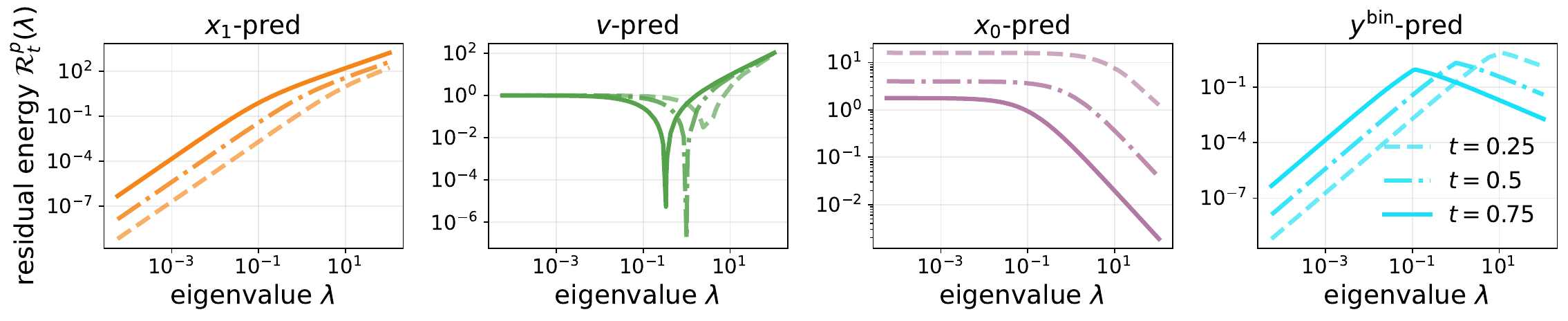}
    \\[1mm]
    \includegraphics[width=\textwidth]
    {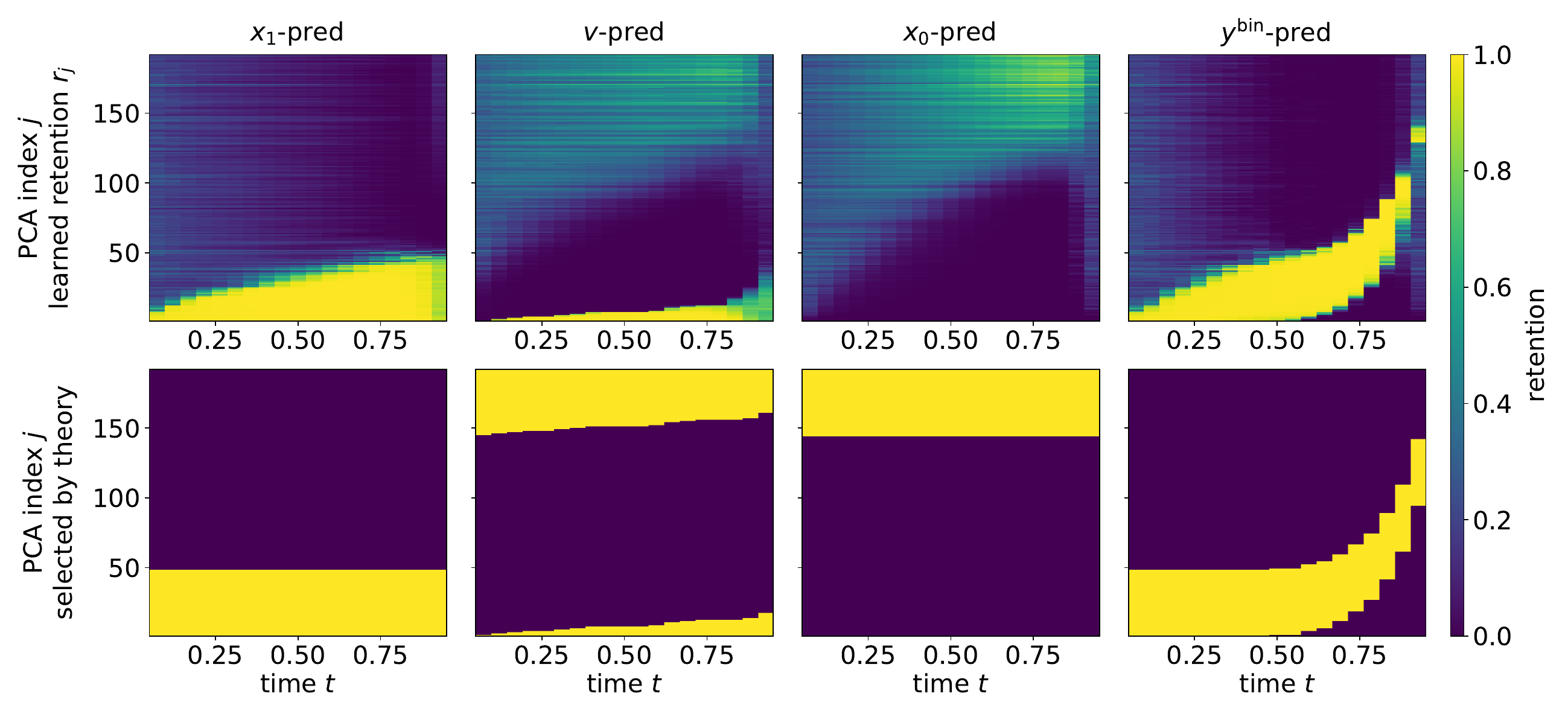}
    \caption{
    Target-dependent spectral allocation under a rank-$48$ bottleneck.
    \emph{Top:} directional residual
    $\mathcal R_{t,j}^{p}$ across the data spectrum.
    \emph{Bottom:} theoretical and learned retention of the corresponding
    PCA directions, measured by $r^p_j(t)$~\eqref{eq:rowspace_retention}. The learned bottlenecks closely follow the allocation predicted by minimizing the residual \eqref{eq:discarded_residual_appendix} : $x_1$ consistently favors the leading
    PCA directions, whereas velocity and the binary hybrid exhibit more
    time-dependent allocations.
    }
    \label{fig:bottleneck_theory}
\end{figure*}

To measure which PCA directions are retained by the learned bottleneck $W$, we
define
\begin{equation}
    r_j^p(t)
    :=
    \left\|P_{W_t^p}u_j\right\|^2,
    \label{eq:rowspace_retention}
\end{equation}
where $P_W$ is the orthogonal projector onto the input space preserved by $W$, i.e., the projection onto $\ker(W)^\perp$.
Values $r_j^p(t)\simeq1$ indicate that $u_j$ is preserved, whereas values
close to zero indicate that it is discarded. As shown in
\Cref{fig:bottleneck_theory}, the learned bottlenecks closely follow the
theoretical allocation predicted by the residual criterion. The agreement is
less sharp for some eigenvalues because the residual curves can exhibit
nearly flat maxima, making the ranking between neighboring directions
ambiguous.

\begin{figure*}[t]
    \centering
    \includegraphics[width=\textwidth]
    {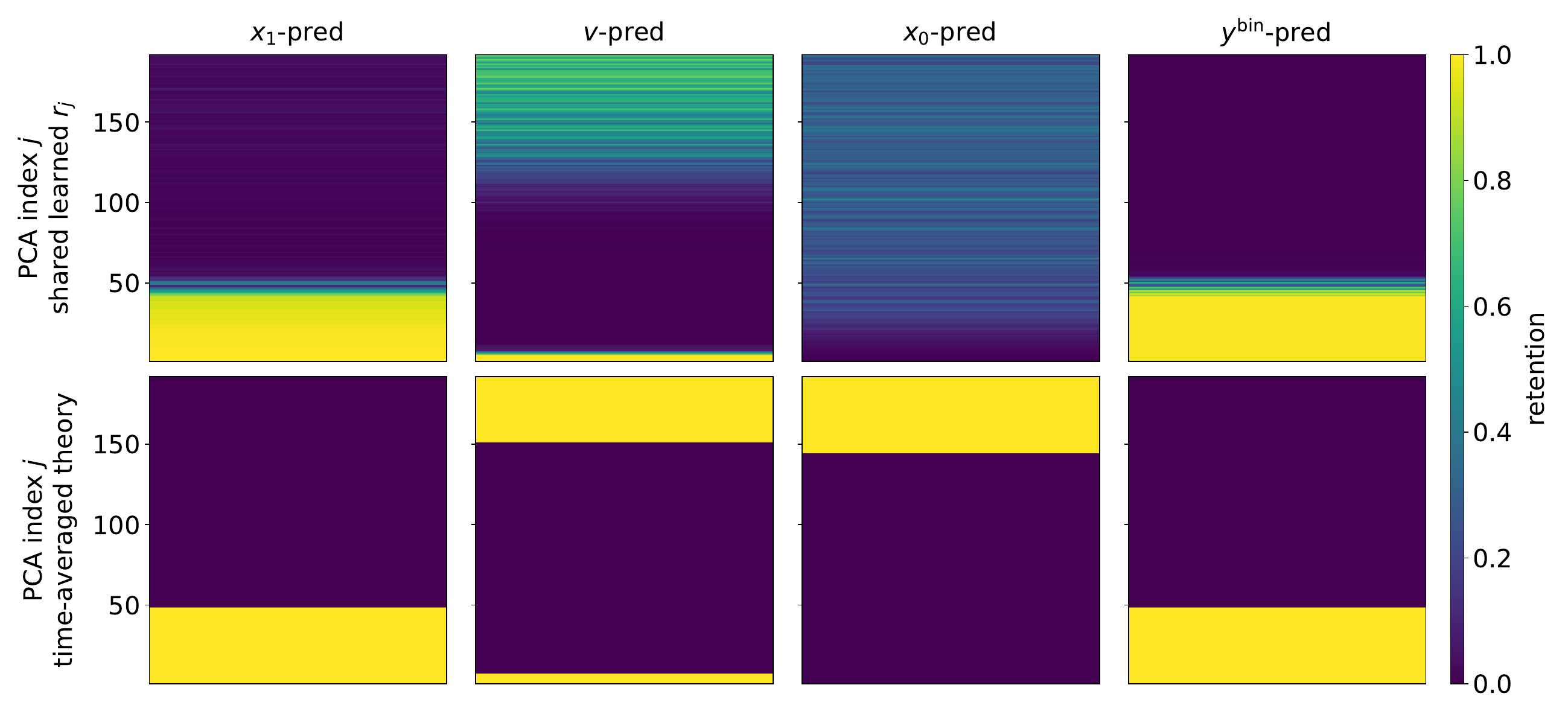}
    \caption{
    Spectral allocation of a rank-$48$ bottleneck shared across time.
    The time-averaged theoretical allocation is compared with the retention
    measured from the learned bottleneck.
    }
    \label{fig:shared_encoder_retention}
\end{figure*}

\paragraph{A bottleneck shared across time.}
A standard time-conditioned model uses a single encoder for all $t$. We
therefore repeat the experiment with one rank-$48$ bottleneck per
parameterization shared across the complete trajectory, and compare its
retention with the directions selected by the residual complexity averaged
over time. As shown in \Cref{fig:shared_encoder_retention},
$x_1$-prediction remains concentrated on the leading PCA directions, whereas
velocity prediction allocates more capacity deeper in the spectrum.

\paragraph{Validation on trained image models.}
Finally, we verify that this target-dependent allocation is also visible in
ViTs trained directly on CIFAR-10. We project their learned patch bottlenecks
onto the patch-PCA basis. Despite using the same bottleneck rank,
$x_1$- and velocity prediction retain different spectral directions:
$x_1$ emphasizes the leading PCA directions, whereas velocity allocates more
of its representation to later directions.

\begin{figure}[t]
    \centering
    \includegraphics[width=0.52\linewidth]
    {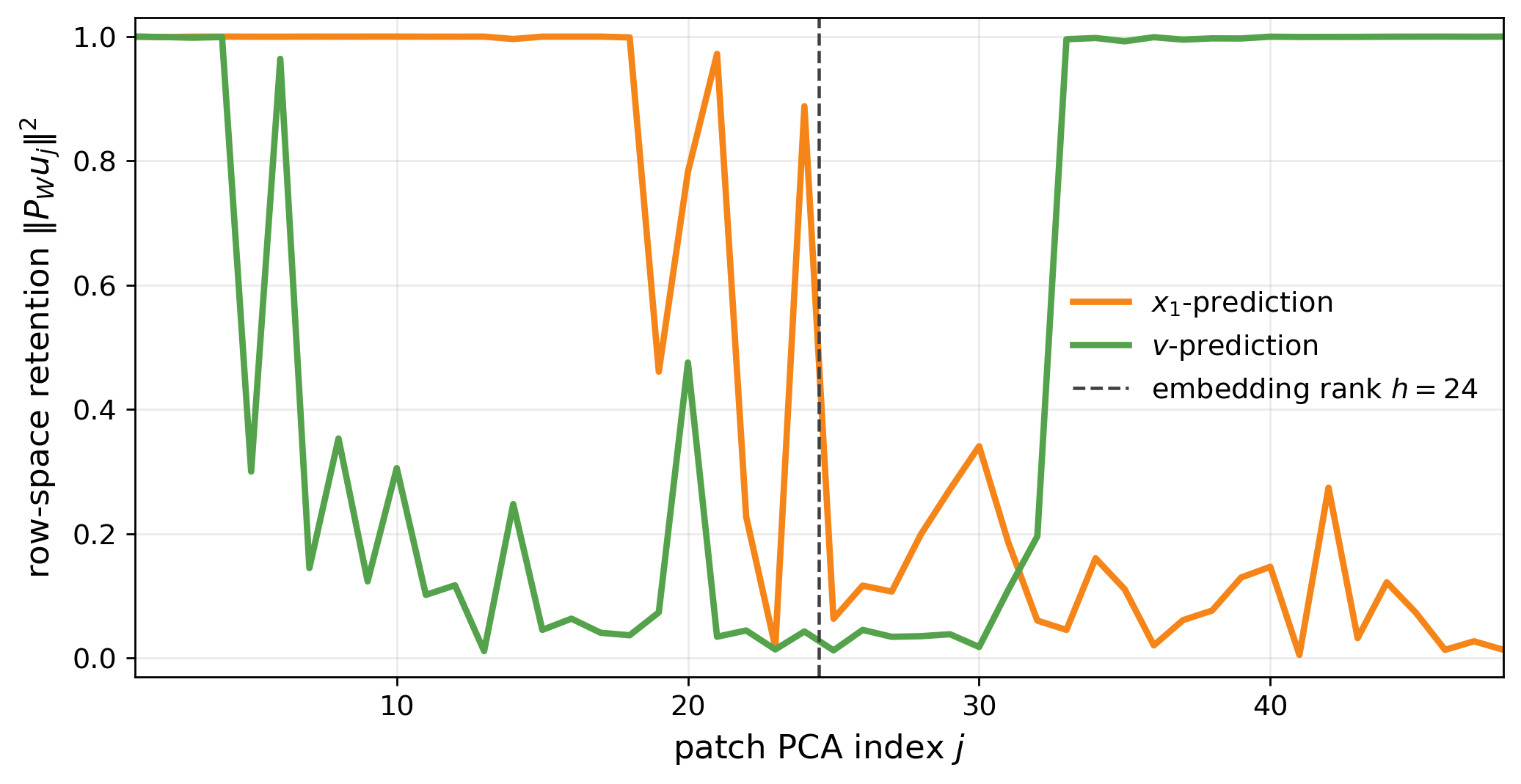}
    \caption{
    Patch-PCA retention of learned rank-$24$ bottlenecks in CIFAR-10 ViTs.
    The two prediction parameterizations learn different retained subspaces.
    }
    \label{fig:cifar_vit_retention}
\end{figure}

\subsection{Discarded directions}
\label{app:discarded_direction_details}

\paragraph{Gaussian analysis.}
We provide the derivation of the discarded-direction result in
\eqref{eq:missing_direction_solution}. For completeness, we consider here the
more general case of a potentially non-centered data distribution,
i.e., $X_1 \sim \mathcal{N}(m,\Sigma_1)$. Let $X^p(t)$ denote the inference-time
trajectory generated by the trained model under parameterization $p$, and
define its centered coordinate along $u_j$ by
\begin{equation}
    \bar X_j^p(t)
    :=
    u_j^\top\bigl(X^p(t)-tm\bigr).
\end{equation}
Consider a PCA direction $u_j$ that is completely unavailable to the learned
residual. Under the Gaussian model, the PCA coordinates are independent, so
the retained representation contains no information about the centered target
in direction $u_j$. The optimal squared-error network correction in this
direction is therefore zero.

In centered coordinates, the inference-time dynamics satisfy
\begin{equation}
    \frac{\mathrm d \bar X_j^p(t)}{\mathrm dt}
    =
    \frac{
        \bar D_{\theta,j}^p(X^p(t),t)
        -
        \bar X_j^p(t)
    }{1-t},
    \label{eq:generative_trajectory_appendix}
\end{equation}
where $\bar D_{\theta,j}^p$ denotes the centered denoiser coordinate.

For $x_1$-prediction, the learned output in the discarded direction is zero,
so $\bar D_{\theta,j}^{x_1}=0$. Hence
\begin{equation}
    \dot{\bar X}_j^{x_1}(t)
    =
    -\frac{\bar X_j^{x_1}(t)}{1-t},
    \qquad
    \bar X_j^{x_1}(t)
    =
    (1-t)X_{0,j},
\end{equation}
and therefore $\bar X_j^{x_1}(1)=0$.

For velocity prediction, a zero learned output gives
$\bar D_{\theta,j}^{v}=\bar X_j^{v}(t)$. Consequently,
$\dot{\bar X}_j^{v}(t)=0$ and
\begin{equation}
    \bar X_j^{v}(1)=X_{0,j}.
\end{equation}

For $x_0$-prediction, a zero learned output gives
$\bar D_{\theta,j}^{x_0}=\bar X_j^{x_0}(t)/t$, and therefore
\begin{equation}
    \dot{\bar X}_j^{x_0}(t)
    =
    \frac{\bar X_j^{x_0}(t)}{t}.
    \label{eq:x0_discarded_dynamics}
\end{equation}
Starting from any time $\tau>0$, these dynamics can be rewritten as
\[
    \frac{\mathrm d}{\mathrm dt}
    \left(
        \frac{\bar X_j^{x_0}(t)}{t}
    \right)
    =0.
\]
Hence $\bar X_j^{x_0}(t)/t$ is constant along the trajectory, and
\begin{equation}
    \bar X_j^{x_0}(t)
    =
    \frac{t}{\tau}\bar X_j^{x_0}(\tau),
    \qquad t\geq\tau>0.
    \label{eq:x0_discarded_solution}
\end{equation}
In particular,
$\bar X_j^{x_0}(1)=\bar X_j^{x_0}(\tau)/\tau$.
Pure $x_0$-prediction cannot be initialized directly at $t=0$ because these
dynamics are singular there.

The binary hybrid avoids this singularity by using $x_1$-prediction before
the directional switching time
$\tau_j=t_j^{\mathrm{switch}}$ and $x_0$-prediction afterwards.
The first phase gives
\[
    \bar X_j^{\mathrm{bin}}(\tau_j)
    =
    (1-\tau_j)X_{0,j}.
\]
Applying \eqref{eq:x0_discarded_solution} from $\tau_j$ to $1$ then yields
\begin{equation}
    \bar X_j^{\mathrm{bin}}(1)
    =
    \frac{1-\tau_j}{\tau_j}X_{0,j}
    =
    \frac{\sqrt{\lambda_j}}{\sigma_0}X_{0,j},
    \label{eq:binary_missing_direction_solution}
\end{equation}
where the second equality follows from
$\tau_j=\sigma_0/(\sigma_0+\sqrt{\lambda_j})$.

Since $X_{0,j}\sim\mathcal N(0,\sigma_0^2)$, we obtain
$\Var(\bar X_j^{x_1}(1))=0$,
$\Var(\bar X_j^{v}(1))=\sigma_0^2$, and
$\Var(\bar X_j^{\mathrm{bin}}(1))=\lambda_j$.

For the continuous hybrid, the learned residual is again zero in the
discarded direction. The remaining analytical term is
\begin{equation}
    \bar D_{\theta,j}^{\mathrm{cont}}
    =
    a_j^\star(t)\bar X_j^{\mathrm{cont}}(t),
    \qquad
    a_j^\star(t)
    =
    \frac{t\lambda_j}
    {(1-t)^2\sigma_0^2+t^2\lambda_j}.
\end{equation}
Writing
$\sigma_{t,j}^2=(1-t)^2\sigma_0^2+t^2\lambda_j$, the corresponding sampling
dynamics are
\begin{align}
    \dot{\bar X}_j^{\mathrm{cont}}(t)
    &=
    \frac{a_j^\star(t)-1}{1-t}
    \bar X_j^{\mathrm{cont}}(t)
    \\
    &=
    \frac{t\lambda_j-(1-t)\sigma_0^2}
    {\sigma_{t,j}^2}
    \bar X_j^{\mathrm{cont}}(t)
    \\
    &=
    \left(
        \frac{1}{2}\frac{\mathrm d}{\mathrm dt}
        \log \sigma_{t,j}^2
    \right)
    \bar X_j^{\mathrm{cont}}(t).
\end{align}
Integrating from $t=0$, where
$\bar X_j^{\mathrm{cont}}(0)=X_{0,j}$, gives
\begin{equation}
    \bar X_j^{\mathrm{cont}}(t)
    =
    \frac{\sigma_{t,j}}{\sigma_0}X_{0,j}.
    \label{eq:continuous_missing_direction_solution}
\end{equation}
Hence
\begin{equation}
    \bar X_j^{\mathrm{cont}}(1)
    =
    \frac{\sqrt{\lambda_j}}{\sigma_0}X_{0,j},
    \qquad
    \Var\!\left(\bar X_j^{\mathrm{cont}}(1)\right)
    =
    \lambda_j.
\end{equation}
Remarkably, the continuous hybrid recovers the correct Gaussian marginal
variance at every time $t$, not only at $t=1$, even though the corresponding
direction is unavailable to the learned residual.

\paragraph{Why the switch must be direction dependent.}
The same calculation also shows why a uniform switching time cannot in
general recover the covariance of discarded directions. If all directions
switch at a common time $\tau$, then
\begin{equation}
    \Var\!\left(\bar X_j^{\mathrm{bin}}(1)\right)
    =
    \left(\frac{1-\tau}{\tau}\right)^2\sigma_0^2,
\end{equation}
which is identical for every discarded direction $j$. It therefore cannot
match an anisotropic target spectrum $\{\lambda_j\}_j$. In contrast, the
direction-dependent choice
$\tau_j=\sigma_0/(\sigma_0+\sqrt{\lambda_j})$ yields
$\Var(\bar X_j^{\mathrm{bin}}(1))=\lambda_j$
independently in every direction.

\paragraph{Non-Gaussian CIFAR-10 patch experiment.}
For \Cref{fig:patch_bottleneck_generation}, we use the original
CIFAR-10 patch distribution rather than its Gaussian approximation. After
centering and globally rescaling the patches such that
$D^{-1}\Tr(\Sigma_1)=1$, we impose the same fixed rank-$48$ top-PCA
bottleneck
\begin{equation}
    W_{\mathrm{PCA}}=U_{1:48}^\top
\end{equation}
for all prediction parameterizations. Thus, all compressed models receive
exactly the same retained coordinates, so differences in the discarded
directions arise from the prediction parameterization rather than from the
bottleneck itself.

We repeat the experiment with a bottleneck chosen to favor $v$-prediction.
Specifically, we average the velocity residual criterion over time,
\begin{equation}
    \overline{\mathcal R}_j^v
    :=
    \int_0^1 \mathcal R_{t,j}^v\,\mathrm dt,
\end{equation}
and retain the $48$ directions with largest
$\overline{\mathcal R}_j^v$. Despite this deliberately favorable choice,
$v$-prediction still fails to reproduce the target variance in directions
that are unavailable to the network, substantially overestimating a large
part of the spectrum. Conversely, $x_1$-prediction strongly underestimates
much of the discarded spectrum. In contrast, both spectral hybrids almost
exactly recover the target variance across PCA directions. This confirms that
their robustness to discarded information is not specific to the top-PCA
bottleneck used in \Cref{fig:patch_bottleneck_generation}.

\begin{figure*}[t]
    \centering
    \begin{minipage}[t]{0.46\textwidth}
        \centering
        \includegraphics[width=\linewidth]
        {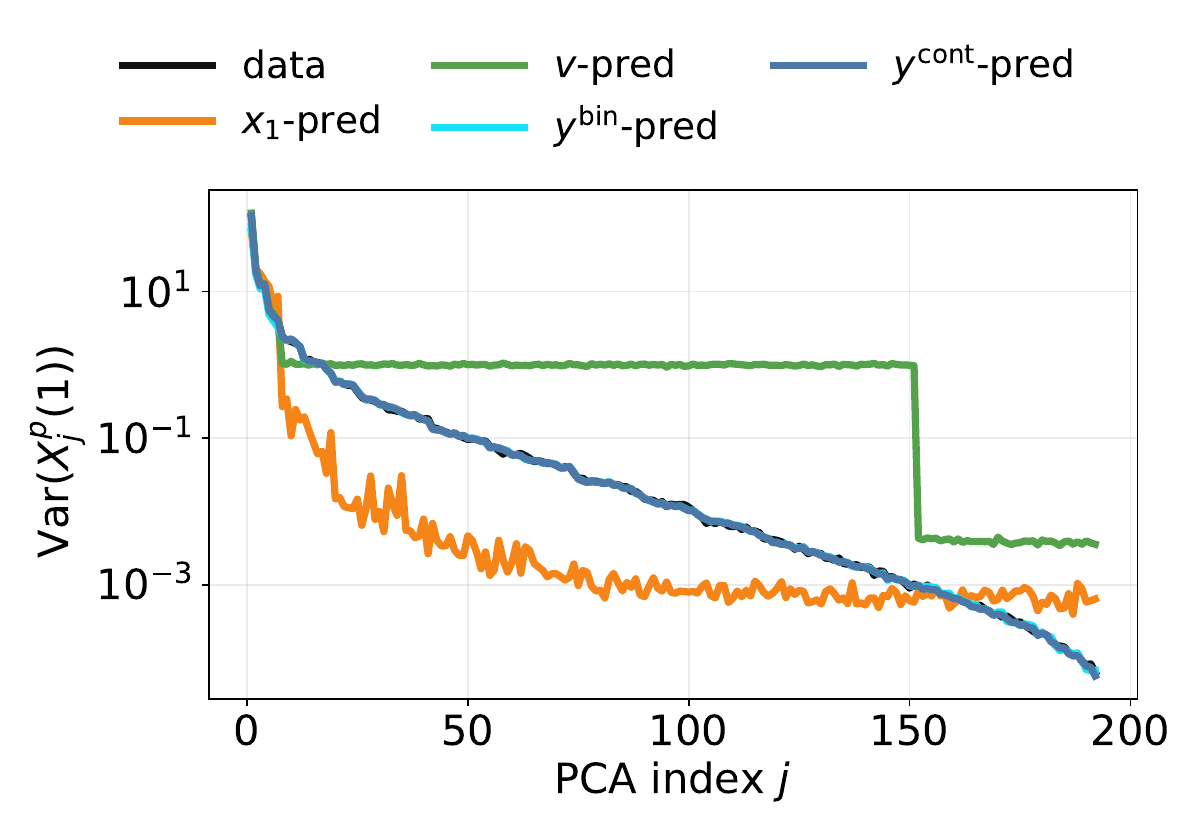}
    \end{minipage}
    \caption{
    Generation with a fixed rank-$48$ time-averaged velocity-optimal
    bottleneck (i.e., the one in \Cref{fig:shared_encoder_retention}) on non-Gaussian CIFAR-10 patches. The spectral hybrids again
    track the target covariance and reduce marginal error across discarded
    directions, showing that the effect is not specific to a top-PCA
    bottleneck.
    }
    \label{fig:v_optimal_bottleneck_generation}
\end{figure*}

\section{Full experimental results}

We report here the complete evaluation for both U-Net and compressed ViT
architectures. These results complement the ViT results shown in the main
paper and confirm that the proposed spectral parameterizations behave
consistently across architectures.

\begin{figure}[t]
    \centering
    \caption{
    FID on Bedroom-$64$ throughout training, for a U-Net (left) and a ViT with
patch size $4$ and compression factor $2$ (right). The two spectral hybrid
parameterizations, $Y^{\mathrm{bin}}$ and $Y^{\mathrm{cont}}$, converge
substantially faster toward their final FID than the fixed prediction targets.
    }
    \begin{subfigure}[t]{0.49\textwidth}
        \centering
    \includegraphics[width=1.0\linewidth]{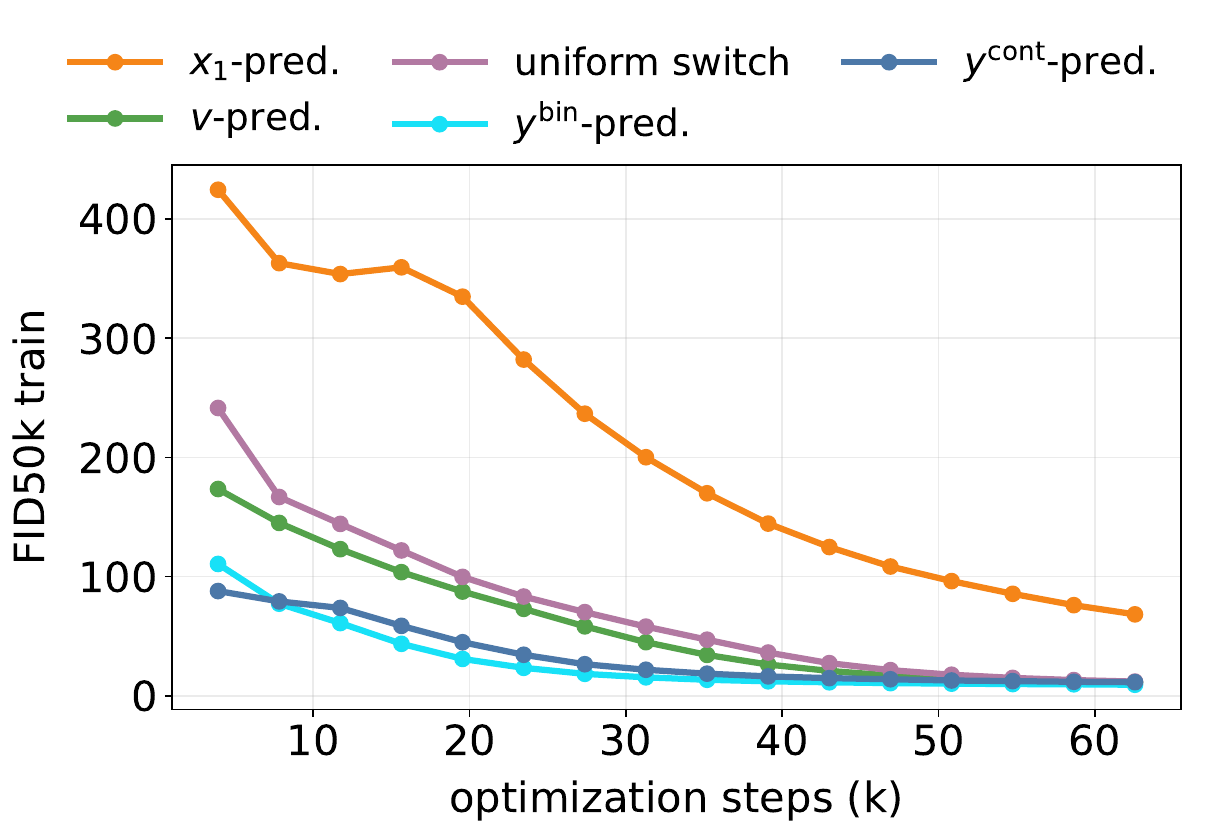}
    \caption{With UNet}
    \end{subfigure}
    \begin{subfigure}[t]{0.49\textwidth}
        \centering
    \includegraphics[width=1.0\linewidth]{figures/lsun/lsun_vit_epoch_fids.pdf}
    \caption{With ViT-4}
    \end{subfigure}
    \label{fig:bedroom_convergence_full}
\end{figure}

On Bedroom-$64$, \Cref{fig:bedroom_convergence_full} shows that the U-Net
reaches a similar final FID with $v$-prediction and with the two spectral
parameterizations. However, the spectral models start from substantially
better generative performance and converge much faster, reaching their final
FID after far fewer optimization steps.

\begin{figure*}[t]
    \centering
    \scriptsize
    \begin{tabular}{@{}>{\centering\arraybackslash}m{0.09\textwidth}@{\hspace{1mm}}ccccc@{}}
        & Epoch 100 & Epoch 200 & Epoch 400 & Epoch 600 & Epoch 800 \\

        \raisebox{0.07\textwidth}[0pt][0pt]{$v$-pred} &
        \includegraphics[width=0.15\textwidth]{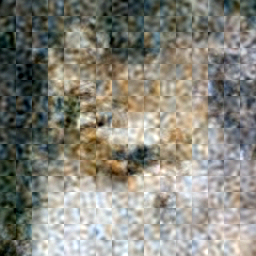} &
        \includegraphics[width=0.15\textwidth]{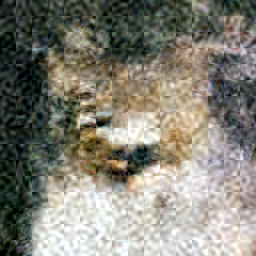} &
        \includegraphics[width=0.15\textwidth]{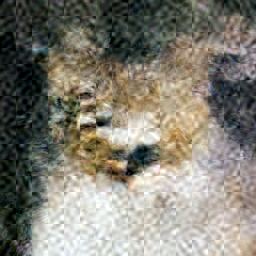} &
        \includegraphics[width=0.15\textwidth]{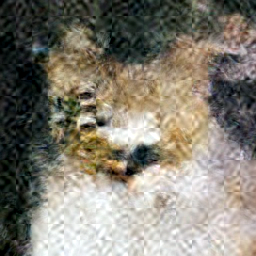} &
        \includegraphics[width=0.15\textwidth]{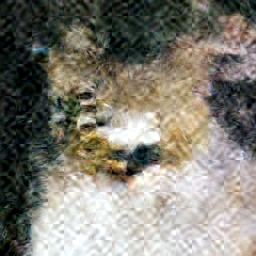} \\
        
        \raisebox{0.07\textwidth}[0pt][0pt]{$x_1$-pred} &
        \includegraphics[width=0.15\textwidth]{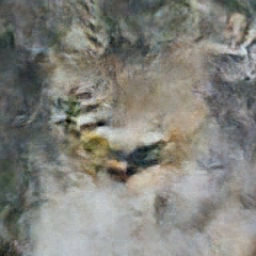} &
        \includegraphics[width=0.15\textwidth]{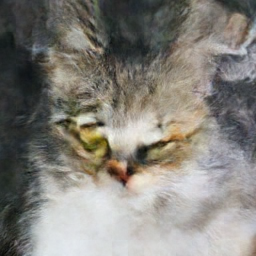} &
        \includegraphics[width=0.15\textwidth]{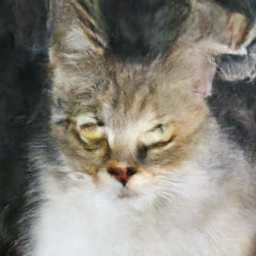} &
        \includegraphics[width=0.15\textwidth]{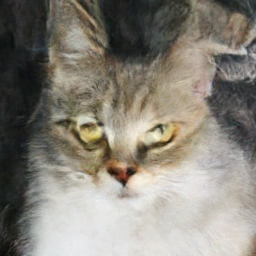} &
        \includegraphics[width=0.15\textwidth]{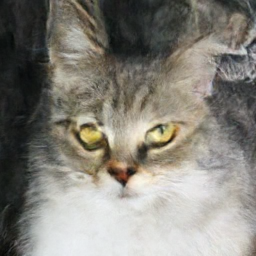} \\

        \raisebox{0.065\textwidth}[0pt][0pt]{\tikz[remember picture,baseline=(afhqoursfive.center)]
            \node[inner sep=0pt,align=center] (afhqoursfive)
            {$y^\text{cont}$-pred\\[-0.3mm]\textcolor{red}{\textbf{(ours)}}};} &
        \tikz[remember picture,baseline=(afhqoursfivestart.base)]
            \node[inner sep=0pt] (afhqoursfivestart)
            {\includegraphics[width=0.15\textwidth]{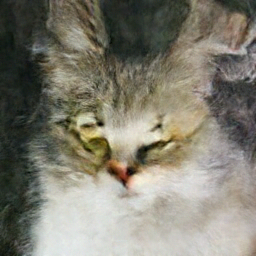}}; &
        \includegraphics[width=0.15\textwidth]{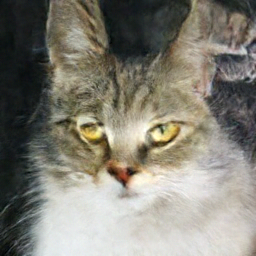} &
        \includegraphics[width=0.15\textwidth]{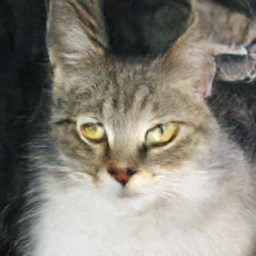} &
        \includegraphics[width=0.15\textwidth]{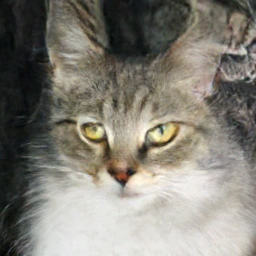} &
        \tikz[remember picture,baseline=(afhqoursfiveend.base)]
            \node[inner sep=0pt] (afhqoursfiveend)
            {\includegraphics[width=0.15\textwidth]{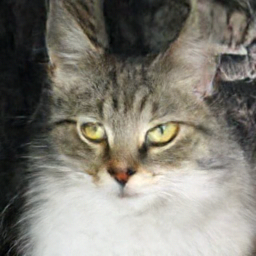}};
    \end{tabular}

    \begin{tikzpicture}[remember picture,overlay]
        \draw[red,line width=1.2pt,rounded corners=1pt]
            ($(afhqoursfive.west |- afhqoursfivestart.north)+(-1mm,0mm)$) rectangle
            ($(afhqoursfiveend.south east)+(1mm,0mm)$);
    \end{tikzpicture}

    \caption{
   AFHQ-$256$ samples generated from two fixed source samples throughout training using a ViT with patch size $16$, token dimension $768$, and no bottleneck. The same source samples are used at every checkpoint, allowing the evolution of individual generated images to be compared directly.
    }
    \label{fig:afhq_epochs_more}
\end{figure*}
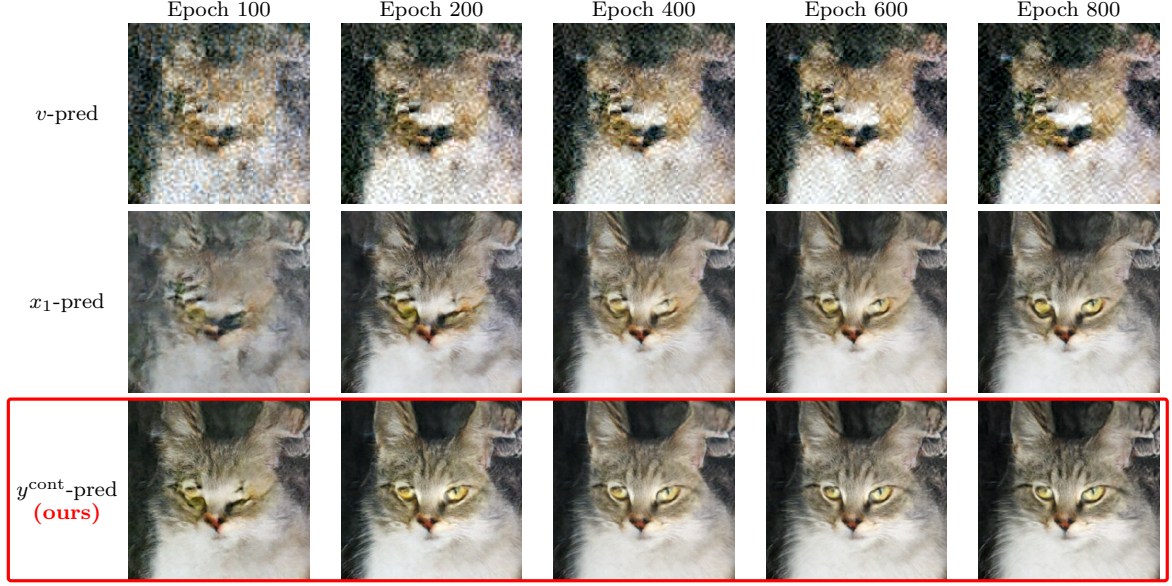

Finally, in \Cref{fig:afhq_epochs_more}, we show additional AFHQ samples generated throughout training from a fixed noise realization, for $x_1$-, $v$-, and $y^{\mathrm{cont}}$-prediction.


\section{Evaluating the hybrid parametrizations on the spiral dataset}

\begin{figure*}[t]
    \centering

    \begin{minipage}[c]{0.115\textwidth}
        \centering
        \textbf{Fixed SNR}\\[1mm]
        $\rho=1$
    \end{minipage}
    \hspace{1mm}
    \raisebox{-0.5\height}{\tcbox[
            colback=gray!8,
            colframe=gray!60,
            boxrule=0.8pt,
            arc=3mm,
            boxsep=0pt,
            left=2mm,
            right=0mm,
            top=0.5mm,
            bottom=0mm
        ]{%
        \begin{tabular}{@{}c@{\hspace{-1mm}}c@{\hspace{0mm}}c@{\hspace{0mm}}c@{}}
            & $D=64$ & $D=128$ & $D=256$ \\[-1mm]
            \spirallabel{$y^{\mathrm{bin}}$-pred} &
            \spiralplot{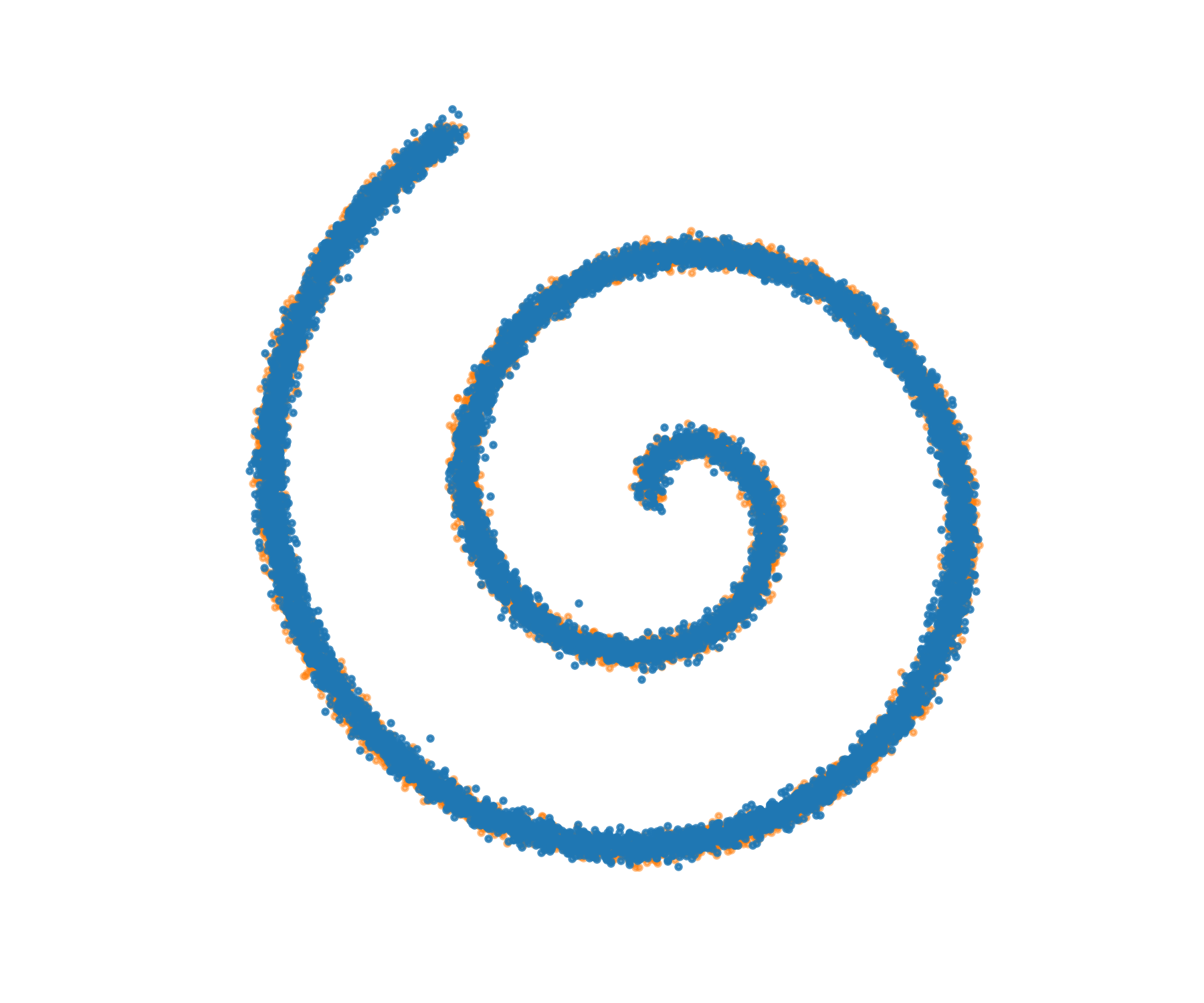} &
            \spiralplot{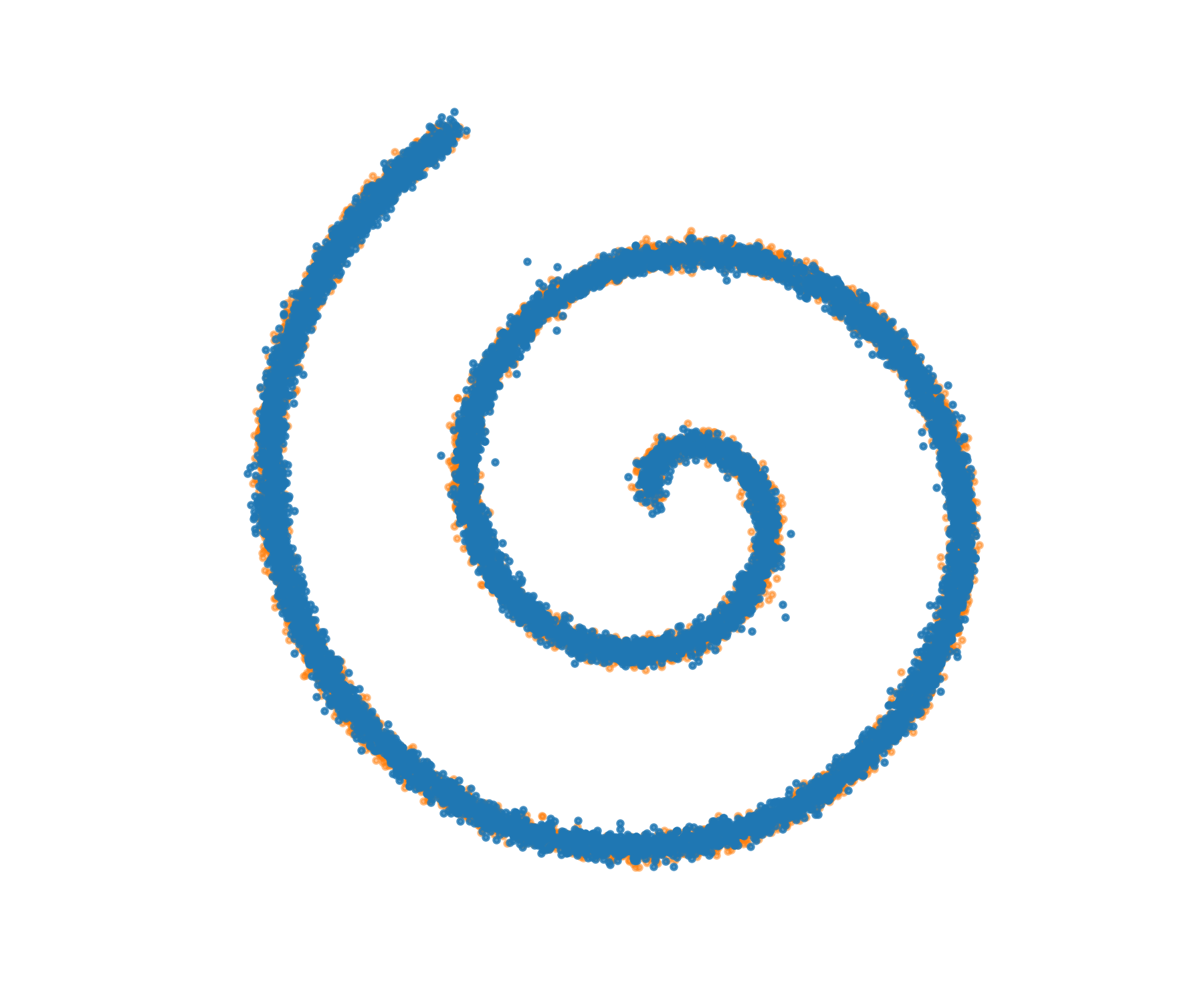} &
            \spiralplot{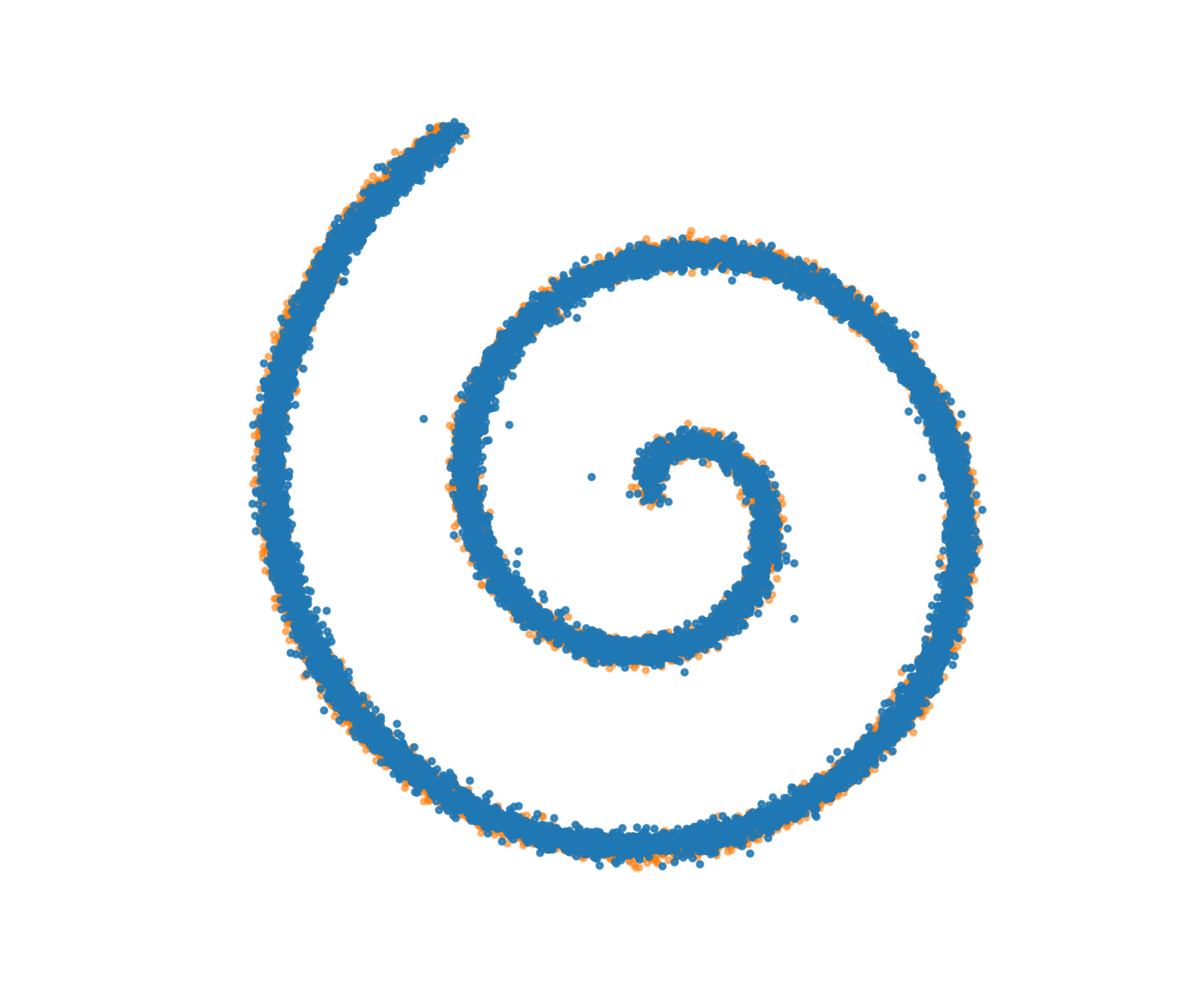} \\[-2mm]
            \spirallabel{$y^{\mathrm{cont}}$-pred} &
            \spiralplot{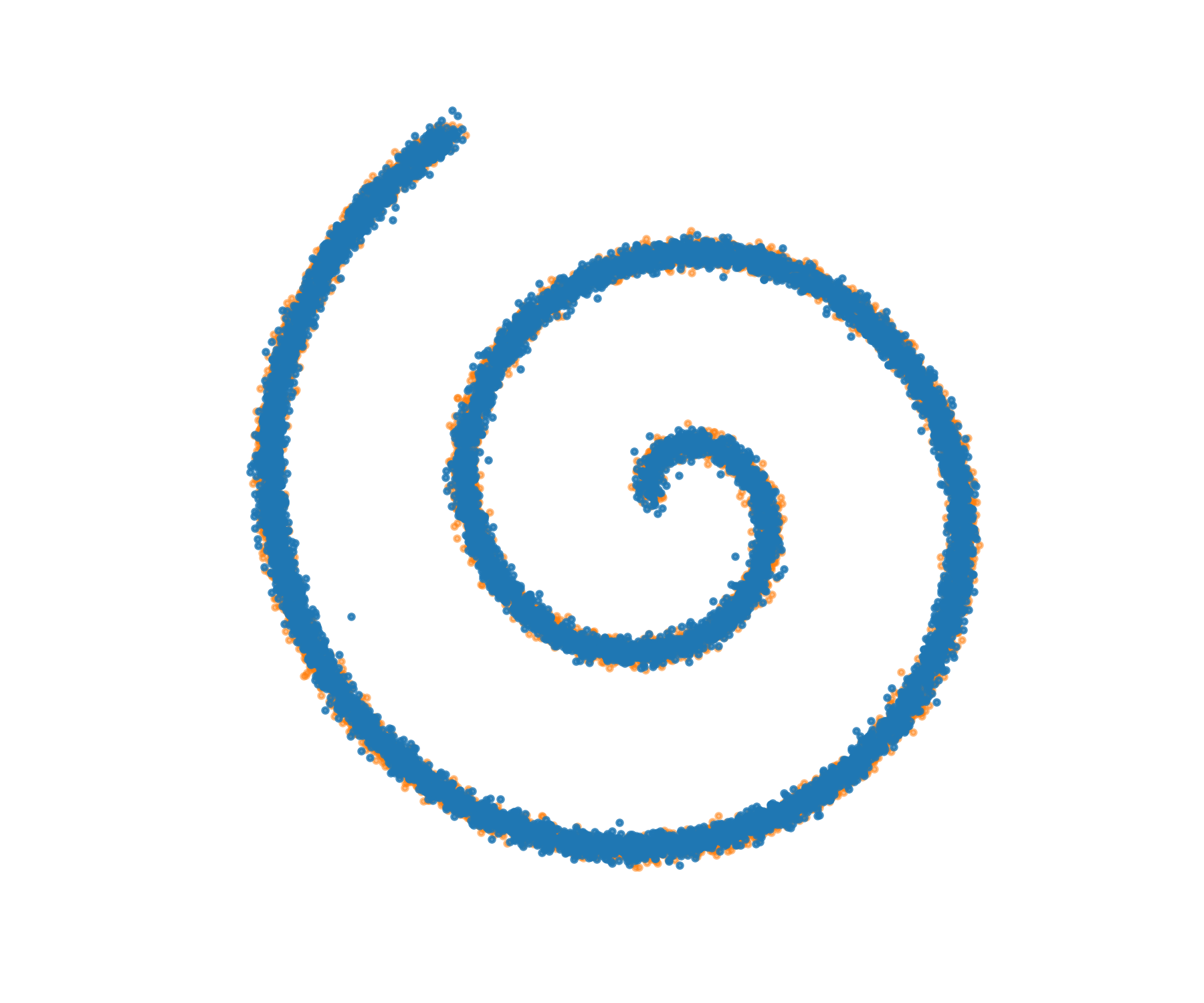} &
            \spiralplot{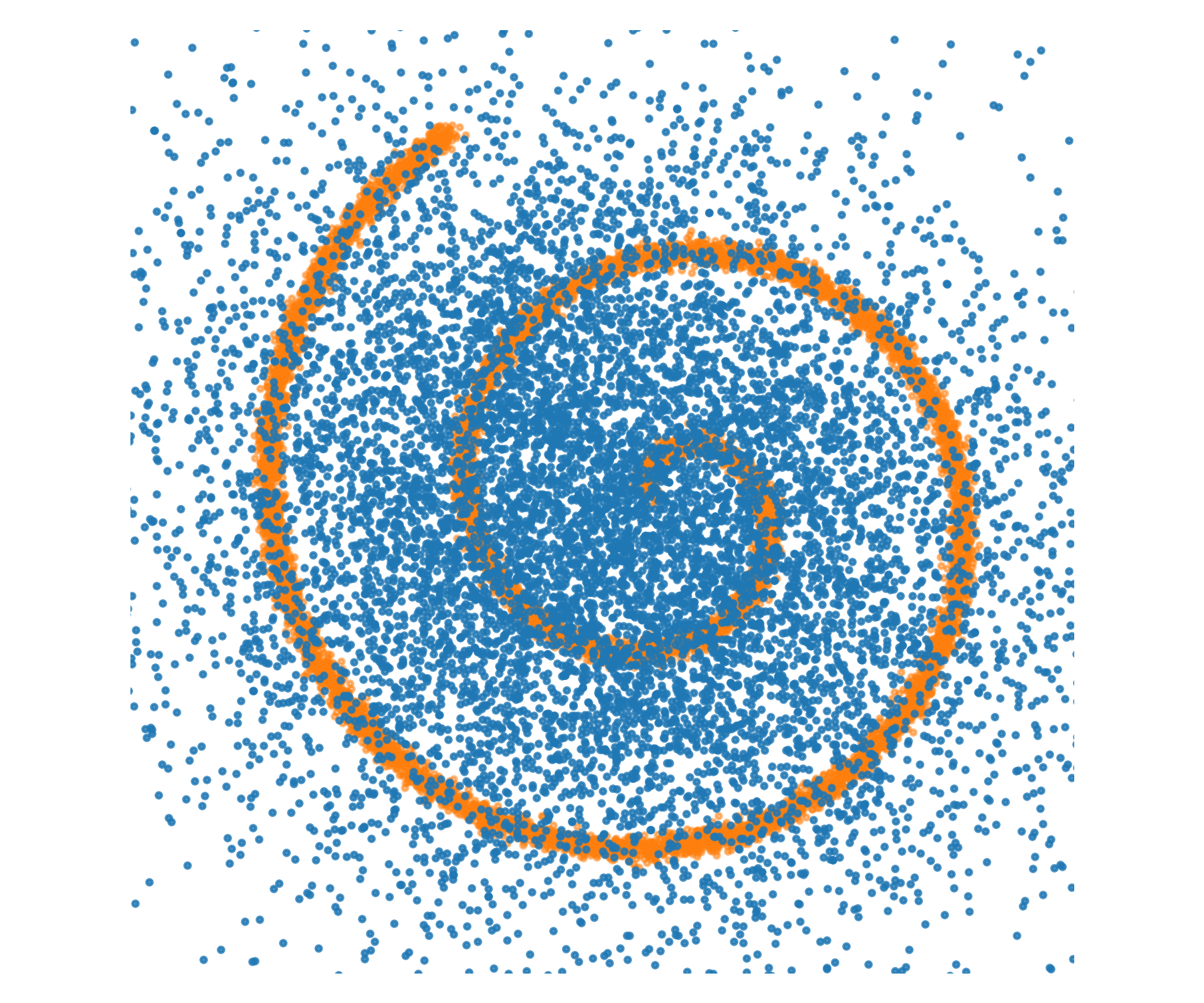} &
            \spiralplot{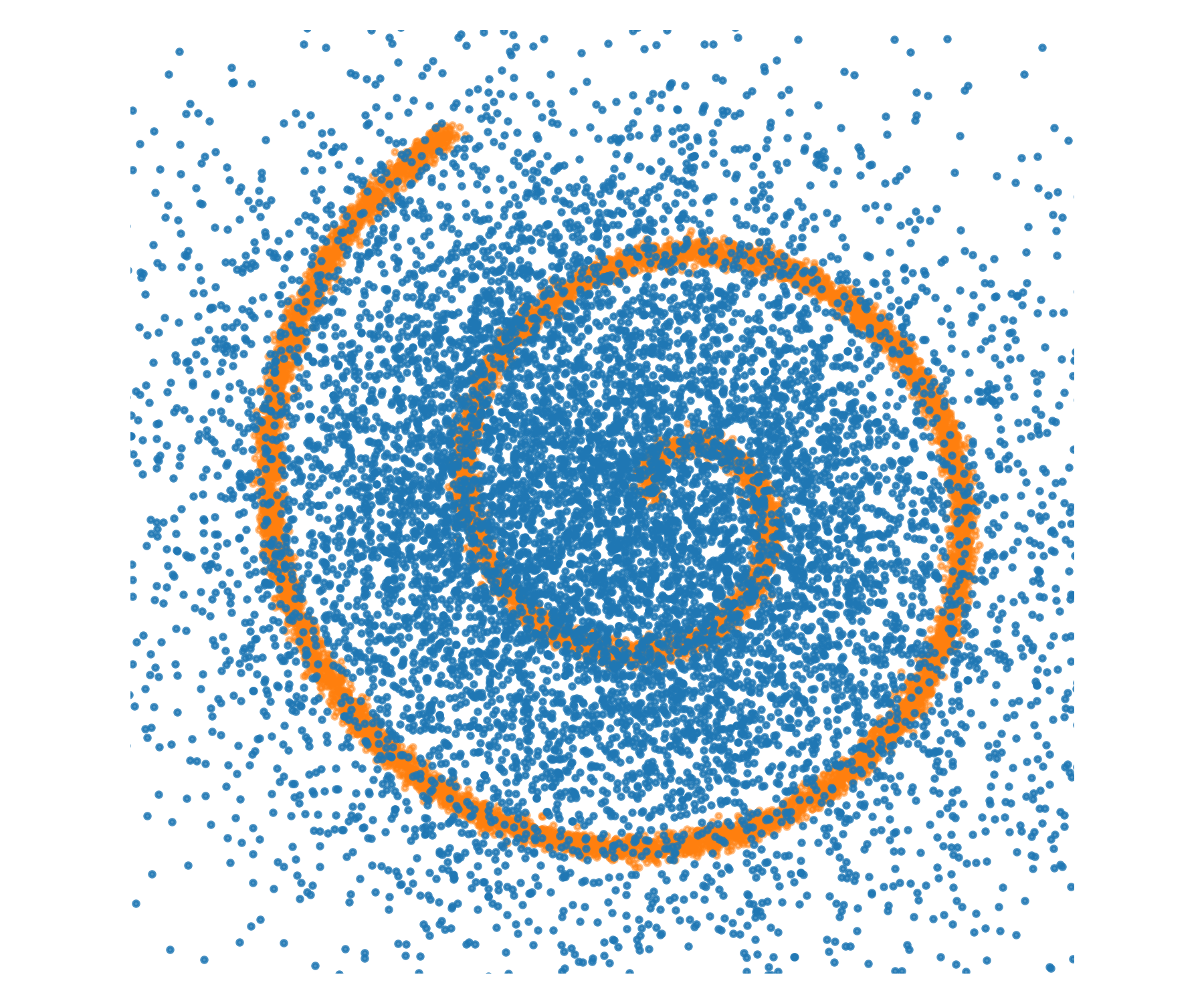}
        \end{tabular}
    }}

    \vspace{2mm}

    \begin{minipage}[c]{0.115\textwidth}
        \centering
        \textbf{Fixed ambient}\\
        \textbf{dimension}\\[1mm]
        $D=256$
    \end{minipage}
    \hspace{1mm}
    \raisebox{-0.5\height}{\tcbox[
            colback=gray!8,
            colframe=gray!60,
            boxrule=0.8pt,
            arc=3mm,
            boxsep=0pt,
            left=2mm,
            right=0mm,
            top=0.5mm,
            bottom=0mm
        ]{%
        \begin{tabular}{@{}c@{\hspace{-1mm}}c@{\hspace{0mm}}c@{\hspace{0mm}}c@{}}
            & $\rho=0.01$ & $\rho=1$ & $\rho=100$ \\[-1mm]
            \spirallabel{$y^{\mathrm{bin}}$-pred} &
            \spiralplot{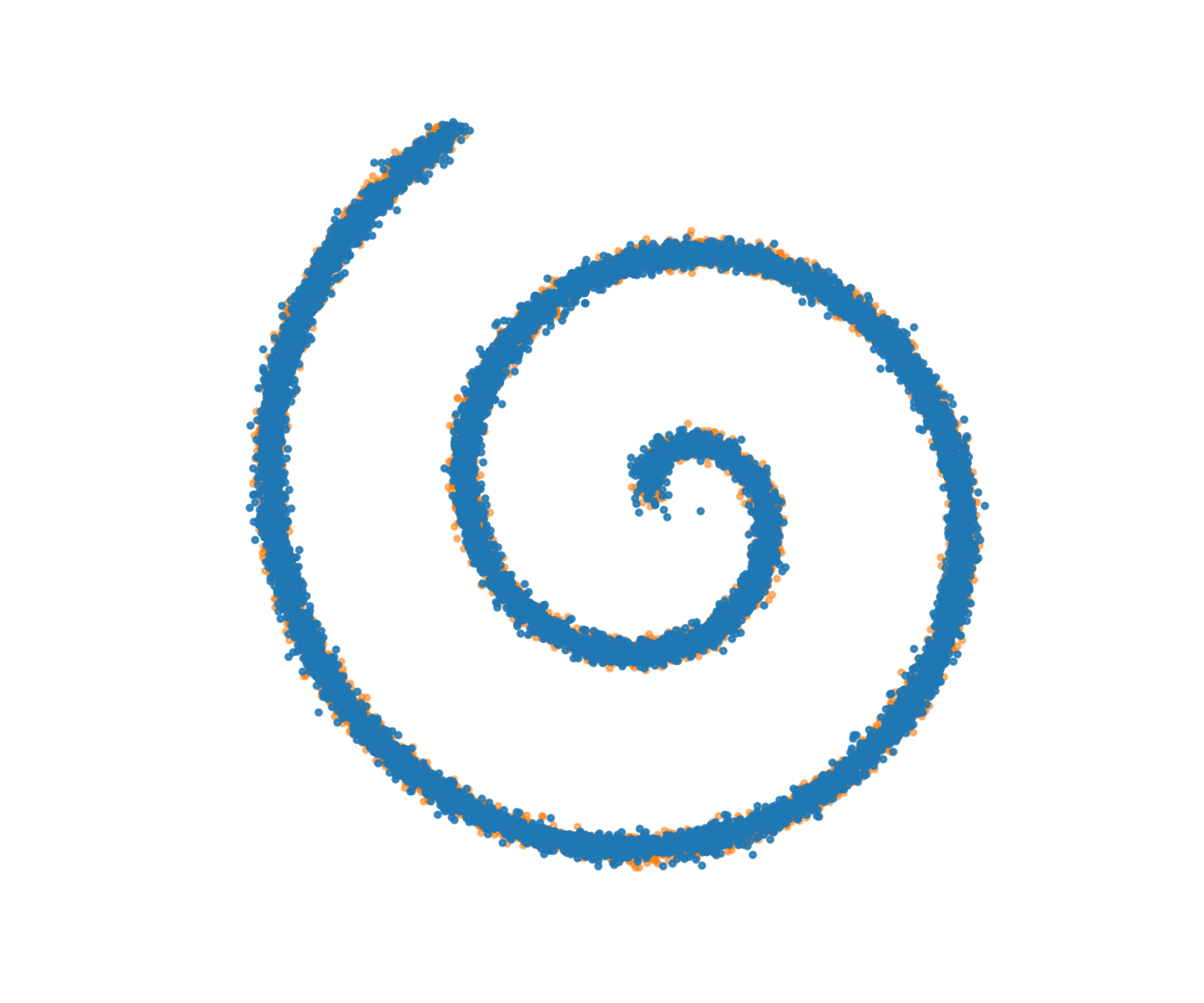} &
            \spiralplot{thick_spiral_D256_snr1.0_sigma0.0601644010322685_spectral_switch_pred.png} &
            \spiralplot{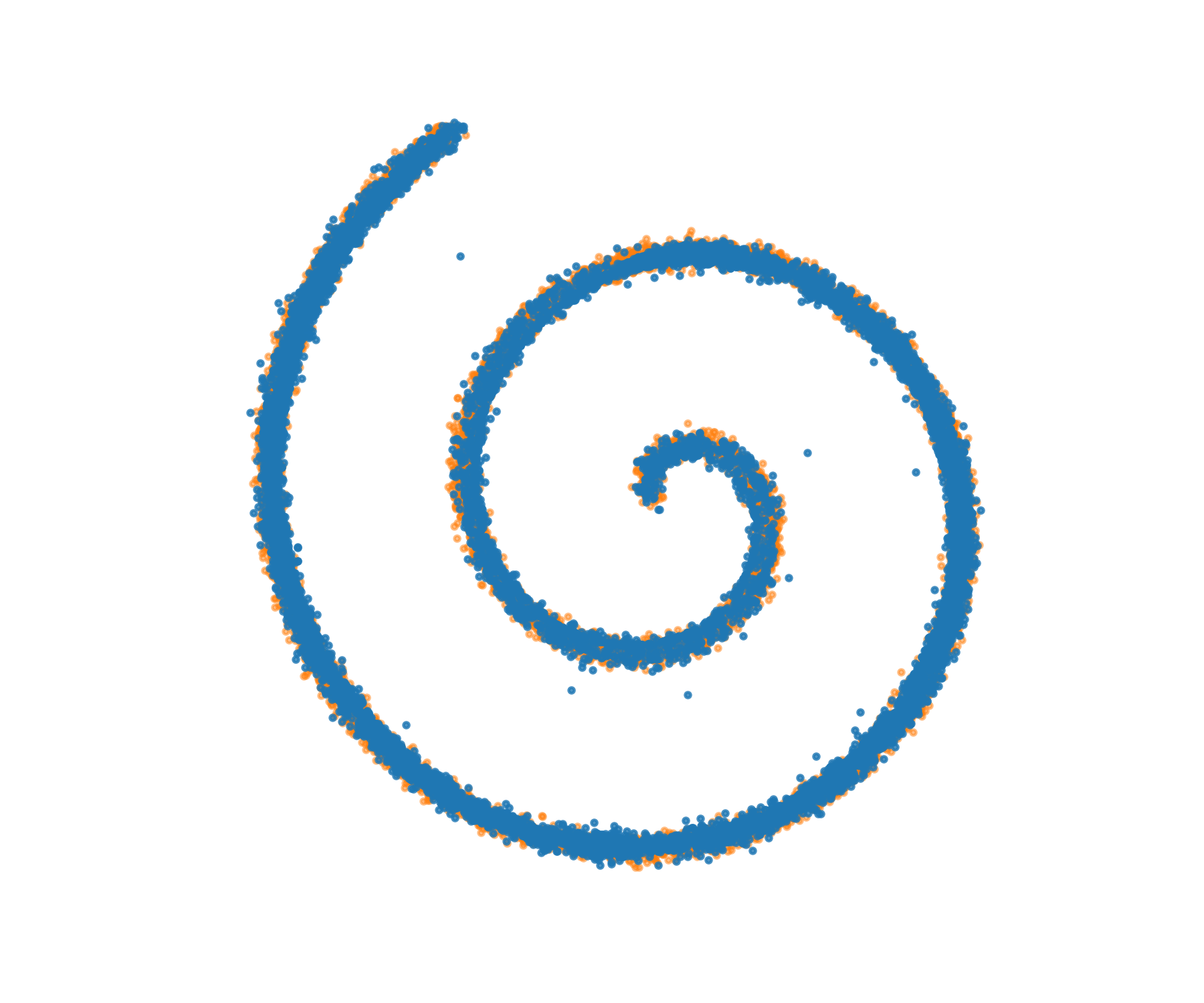} \\[-2mm]
            \spirallabel{$y^{\mathrm{cont}}$-pred} &
            \spiralplot{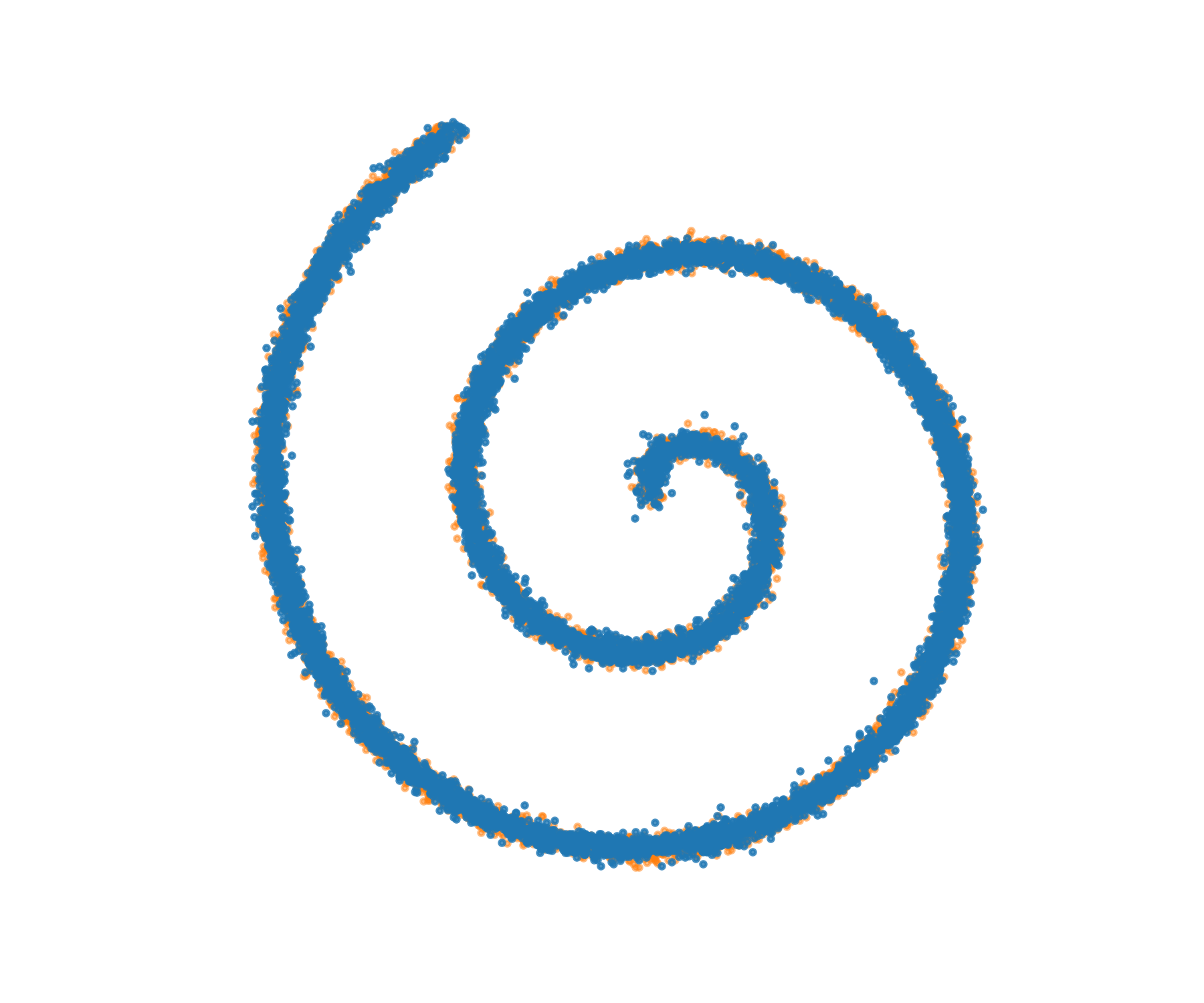} &
            \spiralplot{thick_spiral_D256_snr1.0_sigma0.0601644010322685_continuous_hybrid_pred.png} &
            \spiralplot{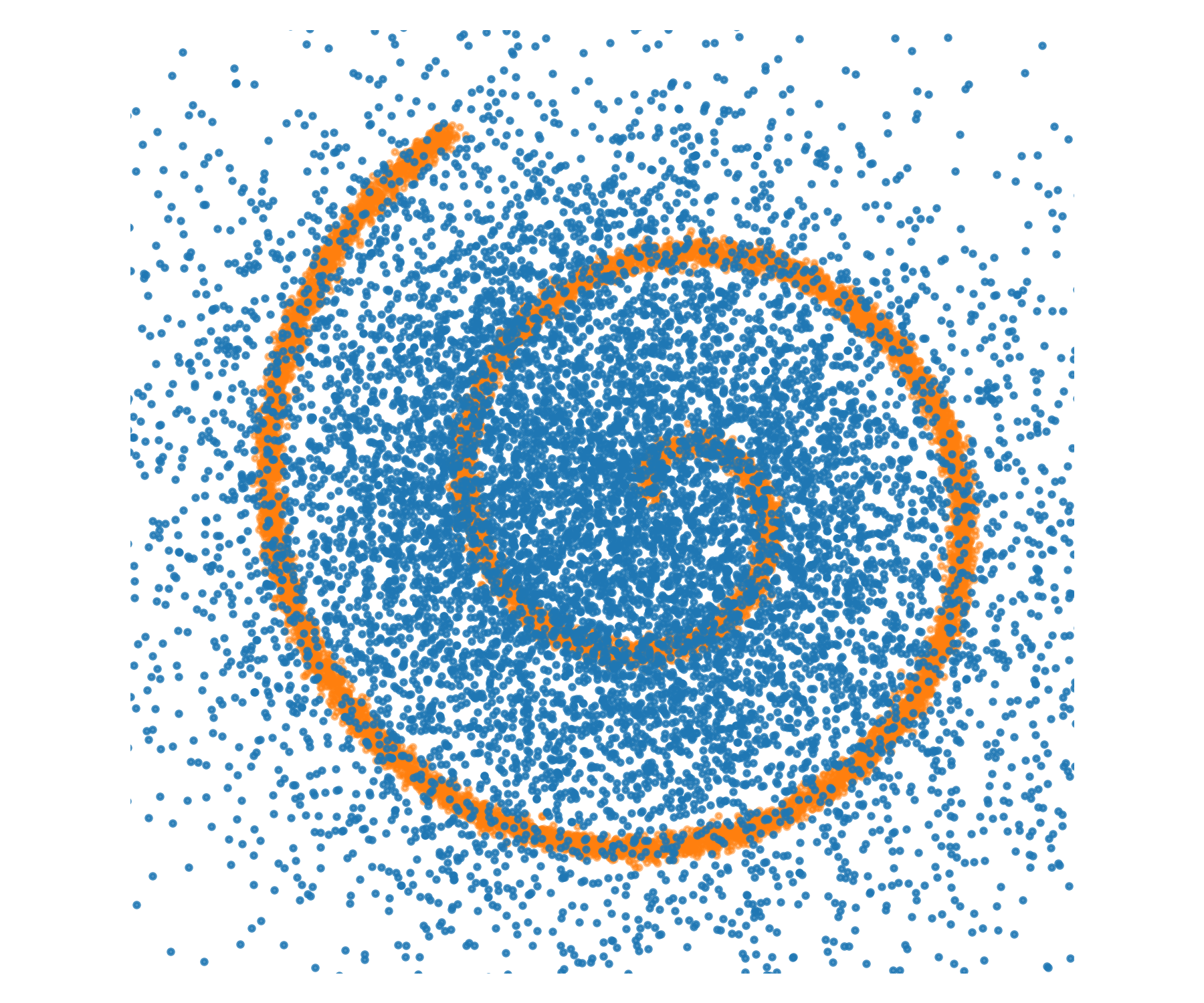}
        \end{tabular}
    }}

    \caption{
        True data are shown in
        \textcolor{taborange}{orange} and generated samples in
        \textcolor{tabblue}{blue}. \emph{Top:} the SNR is fixed to $\rho=1$
        while the ambient dimension varies. \emph{Bottom:} the ambient
        dimension is fixed to $D=256$ while the source--data SNR varies;
        $\rho\in\{0.01,1,100\}$ corresponds respectively to
        $\sigma_0\in\{0.6,0.06,0.006\}$.
        We observe that the $y^{\mathrm{bin}}$ parametrization achieves good performance in every scenario, whereas the $y^{\mathrm{cont}}$ parametrization fails most of the time. 
    }
    \label{fig:spiral_dimension_snr_hybrids}
\end{figure*}

\Cref{fig:spiral_dimension_snr_hybrids} reveals a marked difference between
the two spectral parameterizations. The binary hybrid remains robust across
all considered dimensions and SNRs, whereas the continuous hybrid often
generates an approximately Gaussian cloud instead of recovering the spiral.
We attribute this difference to their distinct dependence on the Gaussian
model. The binary parameterization uses the covariance only to select, in each
direction and at each time, between the two endpoint parameterizations. In contrast, the continuous parameterization explicitly incorporates the Gaussian conditional mean through the linear term $A^\star_tX_t$. For the strongly non-Gaussian spiral distribution, the learned residual must therefore correct this very bad linear approximation.

This interpretation is also consistent with the dependence on dimension and
SNR. In direction $j$, the Gaussian contribution is
$a^\star_j(t)=t\lambda_j/((1-t)^2\sigma_0^2+t^2\lambda_j)$ and is therefore weaker when $\sigma_0$ is larger ($\rho$ is small). This occurs both at smaller SNR and, at fixed global SNR, at smaller ambient dimension since
$\sigma_0^2=\Tr(\Sigma_1)/(D\rho)$. 

\section{Proposed parameterizations for non-centered data}\label{app:non_centered}

We now extend the proposed parameterizations to the case
\begin{equation}
X_1 \sim \mathcal N(m,\Sigma_1),
\qquad
X_0 \sim \mathcal N(0,\sigma_0^2 I),
\end{equation}
with $X_0$ and $X_1$ independent. Introducing the centered variables
$\bar X_1=X_1-m$ and $\bar X_t=X_t-tm$, we have
\begin{equation}
\bar X_t=(1-t)X_0+t\bar X_1.
\end{equation}
Hence, all covariance-based derivations from the centered setting apply
unchanged to $(X_0,\bar X_1,\bar X_t)$. In particular, the switching matrix
$M_t$ and the linear coefficient $A^\star(t)$ depend only on the source and
data covariances and are therefore unaffected by the data mean.

\paragraph{Binary parameterization.}
The binary model is trained to predict $X_1$ directly in the directions
selected by $M_t$ and $X_0$ in the complementary directions, i.e.,
\begin{equation}
N_\theta^{\mathrm{bin}}(X_t,t)
\approx
M_tX_1+(I-M_t)X_0.
\end{equation}
Since
\begin{equation}
X_1=\frac{X_t-(1-t)X_0}{t},
\end{equation}
the conversion to a common $X_1$-denoising representation is identical to
the centered case:
\begin{equation}
D_\theta^{\mathrm{bin}}(x,t)
=
M_tN_\theta^{\mathrm{bin}}(x,t)
+(I-M_t)
\frac{x-(1-t)N_\theta^{\mathrm{bin}}(x,t)}{t}.
\end{equation}
Thus, no explicit mean correction is required for the binary
parameterization when the network predicts the original, non-centered
variable $X_1$.

\paragraph{Continuous parameterization.}
For the continuous model, the linear component derived in the centered
setting is applied to $\bar X_t=x-tm$. The denoiser therefore becomes
\begin{equation}
D_\theta^{\mathrm{cont}}(x,t)
=
m + A^\star(t)(x-tm)
+(1-t)N_\theta^{\mathrm{cont}}(x,t).
\end{equation}
Equivalently, the network is trained to predict the residual
\begin{align}
Y^{\mathrm{cont}}_t
&=
\frac{I-tA^\star(t)}{1-t}(X_1-m)
-
A^\star(t)X_0
.
\end{align}

\section{Implementation details}

The hybrid targets are implemented in a spectral basis. For CelebA-$64$, we use PCA directions computed once from the training set: we collect up to $50,000$ training images, flatten them, subtract the empirical mean, and compute the eigendecomposition of the empirical covariance matrix. These statistics are cached and reused across runs sharing the same dataset and source-noise configuration. For AFHQ-Cat, we instead use an orthonormal two-dimensional DCT applied independently to each RGB channel.

For the binary hybrid $Y^{\mathrm{bin}}$, each spectral direction switches from the $x_1$ target to the $x_0$ target at its analytical switching time $t_j^{\mathrm{switch}}$. In all image experiments, we lower-bound the switching times as
$
t_j^{\mathrm{switch}}
\leftarrow
\max\!\left(t_j^{\mathrm{switch}},0.1\right),
$
to avoid the instability of $x_0$-prediction near $t=0$. For the continuous hybrid $Y^{\mathrm{cont}}$, we use the analytical direction-wise coefficient $a_j(t)$ and apply only an upper cap, $
a_j(t)
\leftarrow
\min\!\left(a_j(t),a_{\max}\right),
$
with $a_{\max}=10$ for CelebA-$64$ and $a_{\max}=100$ for AFHQ-Cat.

For the spiral experiments, we use a MLP with $5$ hidden layers, hidden width
$512$, ReLU activations, and sinusoidal time embeddings of dimension $128$.
Models are trained for $3000$ epochs with batch size $1024$, Adam learning rate
$2\cdot 10^{-4}$, default Adam betas $(0.9,0.999)$, and EMA decay $0.9999$. 

All other models on image datasets are trained with Adam using a learning rate of $2\times10^{-4}$, default Adam coefficients $(\beta_1,\beta_2)=(0.9,0.999)$, and weight decay $10^{-4}$. The learning rate is linearly warmed up over the first $5000$ optimization steps and then kept constant. We maintain exponential moving average (EMA) weights with decay $0.9999$, starting after $1000$ optimization steps.

For the CelebA-$64$ and Bedroom-$64$ ViT experiments, the ViT
models are trained with patch size $4$, token dimension $384$, patch bottleneck
dimension $24$, depth $12$, MLP ratio $4$, and $12$ attention heads. Models are trained for $400$ epochs with batch size $128$. For AFHQ-Cat, we use $256\times256$ images and a ViT with patch size $16$. The corresponding uncompressed patch dimension is $3\times16^2=768$, and the reported AFHQ-Cat runs use token dimension of $768$. The AFHQ-Cat transformer has depth $12$, MLP ratio $4$, and $6$ attention heads. Models are trained for $1000$ epochs with batch size $32$ and source noise $\sigma_0=1$.

\end{document}